\documentclass[10pt,twocolumn,letterpaper]{article}

\usepackage[pagenumbers]{cvpr} %

\newcommand{\myparagraph}[1]{ \vspace{3pt} \noindent {\bf #1}\,\,\,}

\usepackage{graphicx,calc,array}
\usepackage{caption}

\newcommand\lft{\mathopen{}\left}
\newcommand\rgt{\aftergroup\mathclose\aftergroup{\aftergroup}\right}

\definecolor{cvprblue}{rgb}{0.21,0.49,0.74}
\usepackage[pagebackref,breaklinks,colorlinks,allcolors=cvprblue]{hyperref}
\usepackage{graphicx}
\usepackage{adjustbox}

\def\paperID{****} %
\def\confName{CVPR}
\def\confYear{2026}

\title{GR3EN: Generative Relighting for 3D Environments}

\author{\hspace{7mm} Xiaoyan Xing~$^{1, 3}$ ~~~ Philipp Henzler~$^{2}$~~~  Junhwa Hur~$^{1}$~~~ Runze Li~$^{1}$~~~ \\ Jonathan T. Barron~$^{1}$~~~ Pratul P. Srinivasan~$^{1}$~~~  Dor Verbin~$^{1}$ \hspace{7mm}\\\vspace{-8pt}\\
~~~$^{1}$Google DeepMind~~~ $^{2}$Google Research~~~ $^{3}$University of Amsterdam  \\\vspace{-10pt}\\
\normalfont{\url{https://GR3EN-relight.github.io}}
}

\begin{document}
\twocolumn[{%
\maketitle
\thispagestyle{empty}
\begin{center}
\centering
\captionsetup{type=figure}
\vspace{-1em}%
\includegraphics[width=\textwidth]{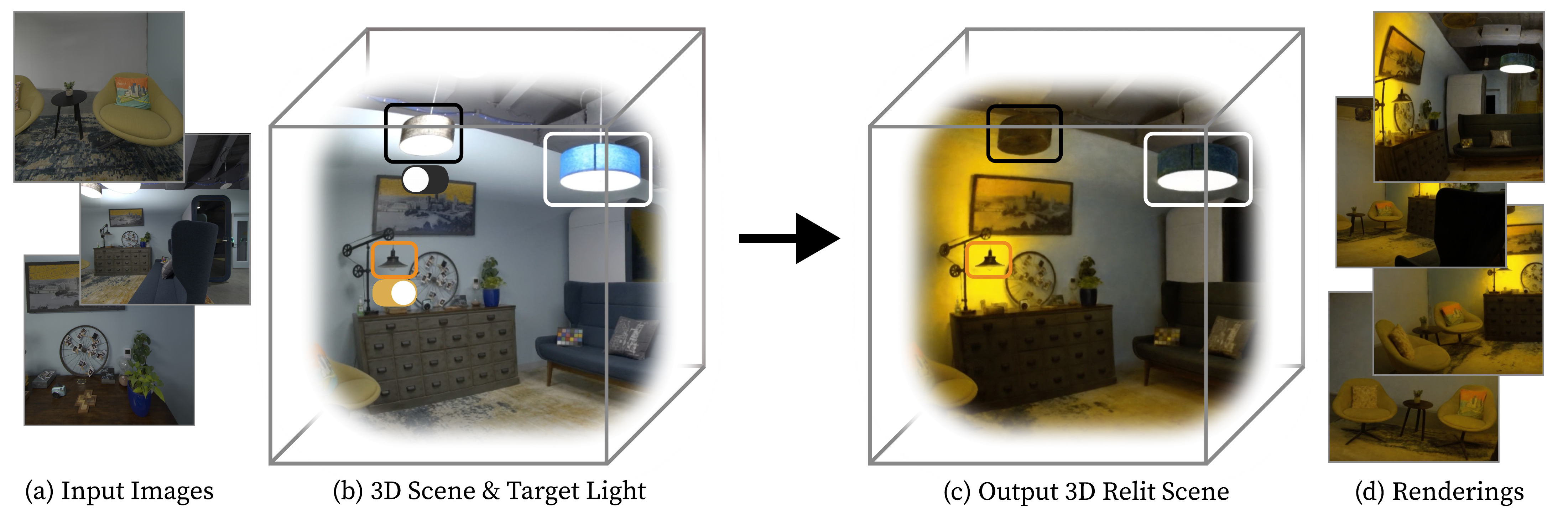}
\captionof{figure}{
Our method provides fine-grained control over the lighting of a 3D scene. Given (a) multi-view input images, we first (b) reconstruct a 3D representation. Given target lighting conditions (here turning off the left overhead light and turning on the lamp on the dresser), our method outputs a (c) relit 3D scene, which can be used to render (d) novel views of the scene under the target lighting.
}%
\label{fig:teaser-fig}%
\end{center}%
\vspace{1em}%
}]

\begin{abstract}

We present a method for relighting 3D reconstructions of large room-scale environments.
Existing solutions for 3D scene relighting often require solving under-determined or ill-conditioned inverse rendering problems, and are as such unable to produce high-quality results on complex real-world scenes.
Though recent progress in using generative image and video diffusion models for relighting has been promising, these techniques are either limited to 2D image and video relighting or 3D relighting of individual objects.
Our approach enables controllable 3D relighting of room-scale scenes by distilling the outputs of a video-to-video relighting diffusion model into a 3D reconstruction.
This side-steps the need to solve a difficult inverse rendering problem, and results in a flexible system that can relight 3D reconstructions of complex real-world scenes.
We validate our approach on both synthetic and real-world datasets to show that it can faithfully render novel views of scenes under new lighting conditions. 

\end{abstract}

\section{Introduction}
\label{sec:intro}

Relighting 3D scene representations is a long-standing challenge that is crucial to enable the use of captured real-world scenes in computer graphics applications. 
Traditionally, this problem has been addressed by first using inverse rendering techniques to decompose a scene's appearance into materials, geometry, and illumination, and then re-rendering that scene under modified illumination conditions.

Since inverse rendering is inherently ill-posed, solutions often rely on strong priors or simplifying assumptions, such as Lambertian reflectance, global directional lighting, or convex geometry. Consequently, these simplifying assumptions compromise photorealism and restrict generalization to complex environments.
Recent methods leverage large generative image and video models for relighting, effectively bypassing the complexities of inverse rendering.
However, existing approaches are limited to 2D relighting of images, or 3D relighting of isolated 3D objects illuminated by an environment map (see a summary of these methods in Table~\ref{tab:related_work}) --- there exist no effective solutions for relighting 3D scene reconstructions, a task which is particularly difficult due to the complex lighting occlusion and interreflection effects that are present in even the simplest room-scale captures.

We introduce a relighting approach that leverages generative video models to relight 3D reconstructions of real-world scenes.
First, we present a method for fine-tuning a pre-trained video diffusion model to perform consistent video-to-video relighting given a user-specified target lighting condition and source video. We then use this video-to-video model to relight video sequences rendered from a reconstructed 3D scene representation.
Finally, we introduce a technique for distilling these relit 2D videos into a high-fidelity 3D scene representation that looks like the input scene, but relit under new lighting conditions. %

Our system is flexible and enables users to perform intuitive 3D scene lighting edits such as turning existing lights on and off, changing their color, or removing outdoor lighting entering through a window. We quantitatively validate our system on synthetic scenes to demonstrate that it can render high-fidelity novel views with different illumination conditions, and we qualitatively demonstrate that it enables rendering photorealistic novel views of real-world 3D scene reconstructions under novel user-controlled lighting.

\vspace{4mm}

\begin{figure*}[t]
  \centering
\includegraphics[width=\textwidth]{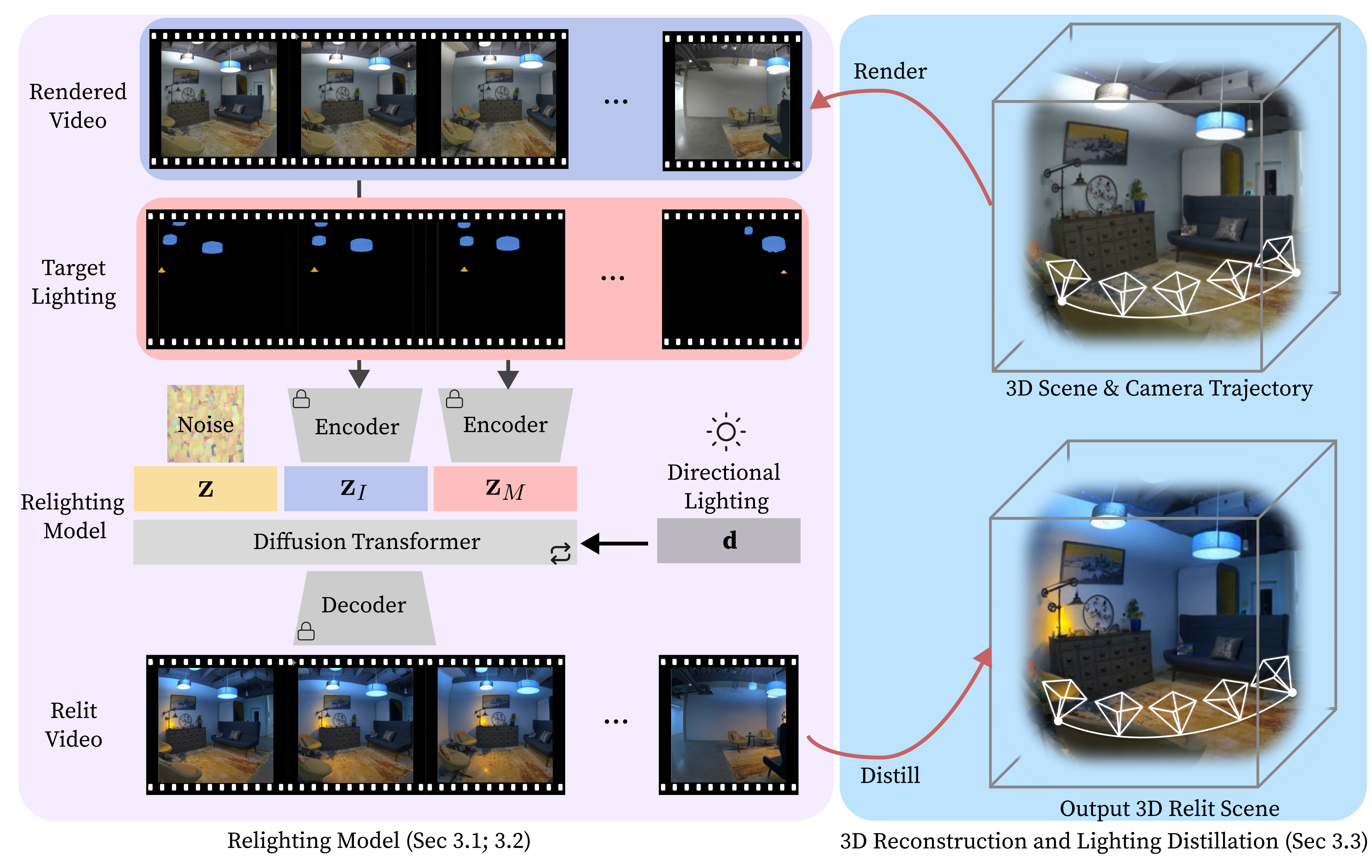}
  \caption{\textbf{Our pipeline for 3D scene relighting.} Starting with a 3D reconstruction (top right), we first render a video of the scene along a chosen camera path (top left). We then use our relighting diffusion model to relight the input video given a target lighting, which is provided as a video whose values specify the color and intensity of the target light sources. Finally, the output relit video is distilled back into the 3D representation (bottom right), resulting in a relit 3D scene that can be rendered from novel viewpoints.}
  \label{fig:pipeline}
\end{figure*}

\vspace{-3ex}
\section{Related Work}
\label{sec:rw}
\begin{table}[t!]
    \centering
        \caption{\textbf{Summary of prior relighting methods.} We introduce the first generative relighting framework for room-scale 3D environments. Our method enables novel view synthesis (NVS) of the relit scene, is conditioned only on the target lighting, and does not require access to any additional G-buffer data such as depth, surface normals, or material properties.}
    \vspace{-0.1in}
    \resizebox{\linewidth}{!}{
    \begin{tabular}{l | c | c |c }
    \toprule
         \textbf{Method} & \textbf{Conditioning} & \textbf{Relit Modality} &\textbf{NVS} \\
        \midrule
        LightLab \citep{lightlab} & depth map, light source image & image & $\times$ \\
        Light-A-Video \citep{Zhou_2025_ICCV} & text & video & $\times$  \\
        DRenderer \citep{DiffusionRenderer2025}& G-buffer, environment map& video& $\times$ \\
        UniRelight \citep{he2025unirelight} & albedo, environment map & video & $\times$\\
        ROGR \citep{tang2025rogr}  & environment map & 3D object & \checkmark\\
        Ours & light source video &3D representation& \checkmark\\
        \bottomrule
    \end{tabular}
    }
    \label{tab:related_work}
\end{table}

\myparagraph{Inverse rendering.}
Early solutions for scene relighting were built upon inverse rendering systems, which decompose a scene's appearance into its constituent factors of materials, illumination, and geometry (or some subset of these, assuming the others are known)~\cite{barrow1978recovering,ramamoorthi2001signal,yu1999,sato1997object}. These methods either directly invert a simple image formation model~\cite{horn1974determining}, or iteratively optimize these factors~\cite{yu1999, barron2014shape,careagaRelighting2025,indoor_3d_1,indoor_3d_4,indoor_3d_3,indoor_env_icra}, typically using simplified models for reflectance and lighting. 
Early attempts at inverse rendering parameterize scene geometry using meshes or depth maps, while the rise of radiance fields~\cite{mildenhall2020nerf, kerbl20233d} enabled effective approaches for inverse rendering that parameterize a scene using the weights of a neural network~\cite{zhang2021nerfactor,nerv2021,attal2024flash,Jin2023TensoIR,mai2023neural,indoor_3d_5} or a set of 3D Gaussians~\cite{gao2023relightable, relit_GS2}. While this has substantially improved inverse rendering results by enabling gradient-based optimization through a differentiable image formation process, this line of work is still limited by the central challenge of inverse rendering: that it is highly ill-posed.
As such, these methods rely on simplifying assumptions of the scene or brittle priors over scene content, which severely limits photorealism, or rely on an abundance of input data (\eg, images of both novel views \emph{and} novel lighting conditions), which limits the settings in which these models can be used.

\myparagraph{2D image relighting.}
Many recent approaches have attempted to circumvent the challenging inverse rendering decomposition problem by directly learning to relight images in a fully data-driven manner.
These approaches have trained feed-forward neural networks~\cite{pandey2021total,sun2019single,griffiths2022outcast,Latent-Intrinsic,scribble_relit}, or generative image models such as Generative Adversarial Networks (GANs) \cite{tewari2020stylerig,pan2021shading,tan2022volux,bhattad2024stylitgan} and more recently diffusion models \cite{lightlab,IC-Light,Neural-Gaffer,xing2025luminet,xing2024retinex,zeng2024dilightnet,luminabrush2024,he2024diffrelight,kocsis2024lightit,choi2024scribblelight} to generate plausible relit images, conditioned on a desired target lighting.
While these methods have been effective at producing plausible 2D results, they do not produce 3D outputs that can be used for view synthesis or other graphics applications.  A concurrent work, LuxRemix \cite{liang2026luxremix}, proposes a relighting framework based on image diffusion models that first decomposes a scene into its constituent light sources, followed by a trained multi-view harmonization model to propagate lighting across different views for scene relighting. In contrast, our method directly achieves consistent room-scale relighting without requiring explicit light source decomposition or view-wise harmonization.

\myparagraph{Video relighting.} 
Building on recent advancements in video diffusion models, a number of works have addressed the challenge of 2D video relighting by using diffusion models \cite{Zhou_2025_ICCV, fang2025relightvidtemporalconsistentdiffusionmodel, Cai_2024_CVPR,he2025unirelight,DiffusionRenderer2025,mei2025lux}. However, these approaches provide limited control over lighting, only allowing editing coarse control such as modifying directional lights or using text prompts to describe the target light. 

\myparagraph{Generative 3D relighting.}
Relighting objects and scenes is a core task in many 3D content creation workflows, and several recent works have focused on relighting reconstructed 3D representations. Neural Gaffer~\cite{Neural-Gaffer} and IllumiNeRF~\cite{IllumiNeRF} trained 2D diffusion models to relight a set of images, conditioned on a target environment lighting, and used these relit images to reconstruct NeRFs of individual objects. Extensions of this approach to multiview diffusion models improves the consistency of the results by jointly relighting multiple images of an object captured under varying illumination conditions \citep{alzayer2024generativemvr, tang2025rogr}. Other approaches use generative models to first estimate material properties and then relight the objects~\cite{litman2025lightswitch}.
While these techniques produce impressive relighting results for isolated objects, they all rely on environment lighting representations that cannot be applied to larger scenes.

\section{Method}
Given a set of posed multiview images and user-provided target lighting, our method produces a 3D representation of the relit scene. Our pipeline addresses this task in three stages. First, we reconstruct a 3D scene representation from the input multiview images. Second, we render a video sequence from this 3D representation and pass it along with the target lighting through our video-to-video relighting diffusion model, to generate a relit video from the same camera path. Finally, we ``distill'' the relit appearance from the relit video back into the 3D representation, to get a final 3D representation that enables rendering the relit scene from any camera pose. This pipeline (not including the initial 3D reconstruction) is illustrated in Figure~\ref{fig:pipeline}.

In Sections~\ref{sec:relighting_model} and~\ref{sec:relighting_training} we describe the relighting model and its training process, and in Section~\ref{sec:distill} we describe the 3D reconstruction and distillation pipeline.

\subsection{Relighting Model}
\label{sec:relighting_model}

The main component of our method is a video-to-video relighting diffusion model. We start with the Wan 2.2 TI2V-5B base model~\cite{wan2025}, and we fine-tune it for video relighting. Our model takes an input video and a specification of the target lighting. The input video is rendered from our 3D reconstruction, and it provides the majority of the content of the output video, such as geometry, materials, and camera path. The goal of the relighting model is to maintain the content of the input video while only changing the lighting.

Our lighting conditioning must model all light sources in indoor environments, which consists of directly observed light sources (\eg, ceiling lights or standing lamps), and from unobserved sources such as the sun shining through a window.

For observed indoor light sources, we provide the relighting model with a per-pixel annotation of the target lighting, containing its intensity and its hue. For unobserved light sources outside the scene, we assume that external illumination is dominated by the sun, so we provide a single scalar for the sun's intensity.

\myparagraph{Visible lighting conditioning.}
Given an RGB video $\textbf{I}\in \mathbb{R}^{T\times H\times W\times 3}$ consisting of $T$ frames and spatial dimensions $H\times W$, we define the visible lighting conditioning as a video $\textbf{M}\in\mathbb{R}^{T\times H\times W\times 3}$. 
The frames of $\textbf{M}$ correspond to the same cameras and content as $\textbf{I}$, but their values control the light sources. A pixel with RGB value $\mathbf{c} \in \mathbb{R}^3$ indicates that the light source observed at that pixel should have intensity $\mathbf{c}$, whereas a pixel with value $(-1, -1, -1)$ indicates that the light source should not be edited. Having a special value of $-1$ for not editing is important because otherwise we would have to set $\mathbf{c}$ to exactly the right value at every pixel. We provide examples of our lighting conditioning videos $\textbf{M}$ in Figure~\ref{fig:data_pipeline} and the supplement.

Prior work on relighting using generative models such as LightLab~\cite{lightlab} concatenates the input image and the lighting condition along the image's channel dimension. This uses the fact that the content in the input and output images should be aligned. However, in our experiments this approach did not yield high quality video relighting results. Other approaches like UniRelight~\cite{he2025unirelight} instead concatenate the output and conditioning videos along the temporal dimension, along with positional embedding that is identical for all videos (\ie, the target lighting videos, input video, and output video all use the same positional embedding). Instead, we found another method to work better than these existing approaches: we also concatenate the videos along the temporal dimension, but we use rotary positional encodings~\cite{heo2024rotary}. This design allows full self-attention and provides different positional encodings to different videos. Empirically we show that this approach results in superior performance compared to other conditioning strategies.

\myparagraph{Directional lighting conditioning.}
Our representation of unobserved light is extremely simple, consisting of a single scalar for the sun's intensity. Similar to LightLab~\cite{lightlab}, the intensity gets lifted to a high-dimensional space using Fourier features, passed through a shallow MLP to get the directional lighting embedding $\mathbf{d}$ and fed into the model's cross-attention layers.

\subsection{Relighting Model Training}
\label{sec:relighting_training}

Each batch for training our diffusion model consists of an input video $\textbf{I}$, the lighting condition video $\textbf{M}$, and the directional lighting embedding $\mathbf{d}$. A pretrained, frozen autoencoder~\cite{wan2025} is used to map the video sequences to spatial-temporal latents, one for the input video $\mathbf{z}_I$, and one for the lighting condition video $\mathbf{z}_M$. Those two latents are concatenated along the temporal dimension with the output video latent (also compressed by the same     shared autoencoder) which is being denoised $\mathbf{z}$. A flow-based noise scheduler $f_\text{noise}(\mathbf{z},t)$ introduces noise to $\mathbf{z}$ at a given timestep $t$. The parameters $\theta$ of the relighting model $f_\theta$ are optimized by minimizing the objective function at time step $t$:
\begin{equation}
    \mathcal{L}(\theta) = ||\hat{\mathbf{z}}_t-f_\text{noise}(\mathbf{z},t)||^2_2,
\end{equation}
where the noise added to the latent is computed as:
\begin{equation}
    \hat{\mathbf{z}}_{t}= f_\theta(f_\text{noise}(\mathbf{z}_t,t),\mathbf{z}_I,\mathbf{z}_M,\mathbf{d},\mathbf{e};t).
\end{equation}
After denoising, we use the pretrained autoencoder's decoder to obtain the relit output video.

During training, we sample the timestep $t$ with a strong bias towards higher noise magnitudes, by drawing it from the following distribution:
\begin{equation}
    p(t) = 
    \begin{cases} 
        \frac{\rho}{\tau} & \text{if } 0 \le t \le \tau, \\
        \frac{1-\rho}{1-\tau} & \text{if } \tau < t \le 1,
    \end{cases}
\end{equation}
where we set $\tau=0.4$ and $\rho=0.85$, meaning that 85\% of training steps are done on 40\% of the noisiest timesteps.

We find that this bias promotes much faster convergence and improved relighting fidelity. This is because plausible relighting requires global reasoning to ensure illumination consistency across different parts of the scene, and correctness with respect to the target lighting. This low-frequency structure is primarily established during high noise levels.
Low noise levels are responsible for refining high-frequency details, which is a relatively trivial task in our setting, since the scene's content and texture are already provided in the input video.

\subsection{Reconstruction and 3D Distillation}
\label{sec:distill}

The first step in our pipeline is obtaining a 3D reconstruction of a scene. We can reconstruct a 3D scene from a collection of multi-view images using Neural Radiance Fields~\cite{mildenhall2020nerf} or 3D Gaussian Splatting~\cite{kerbl20233d}. However, because our method only relies on videos rendered from the 3D model (and not the multi-view images used to reconstruct it), our method can also be applied to existing 3D reconstructions.

Once a 3D reconstruction is available, we render it into a video to be passed to our relighting video-to-video model. We render an elliptical camera path using a fisheye camera with a wide field of view. This is because the video diffusion model can only process a limited number of frames at once, so increasing the field of view of the camera improves the coverage of the scene.

After passing this video through our relighting model, we distill the relit video by fine-tuning the 3D model to match the relit video. The relit 3D representation then enables rendering images and videos from arbitrary new viewpoints. 

Throughout the main paper, our results are obtained by using Zip-NeRF for reconstruction and 3D distillation due to its high quality~\cite{barron2023zipnerf}.
However, nothing about our approach is specific to any particular radiance field technique, and one could instead use a real-time (but slightly lower quality) approach such as 3D Gaussian Splatting~\cite{kerbl20233d}.

\section{Training Data}
\label{sec:training_data}
\begin{figure}[t]
  \centering
  \includegraphics[width=\linewidth]{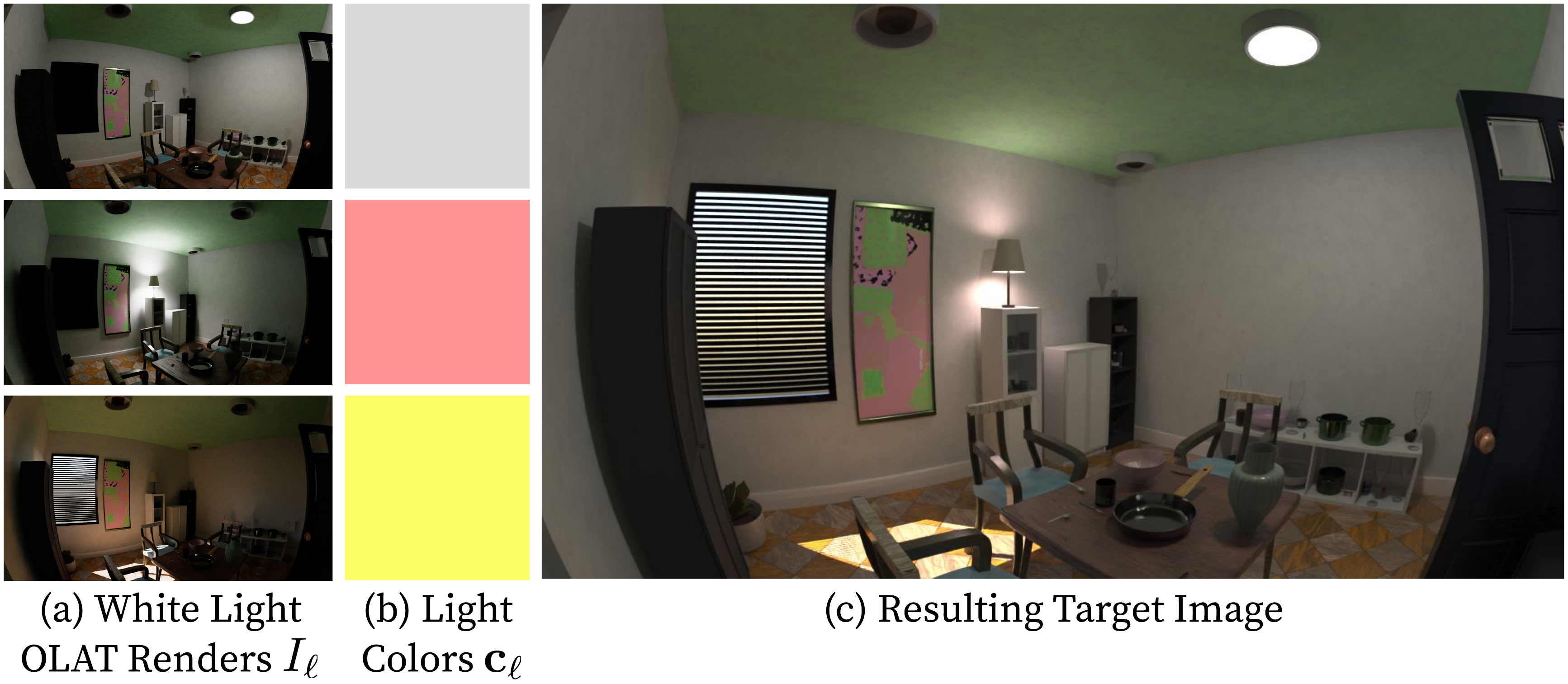}
  \caption{\textbf{A visualization of our ``one-light-at-a-time'' data synthesis pipeline.} We (a) separately render the scene illuminated by each of the light sources, (b) multiply each one by its target light color, and (c) combine all light sources according to Equation~\ref{eq:olat_combo} to get the target illumination. }
  \label{fig:data_pipeline}
\end{figure}
Training a generative model for relighting requires a large-scale dataset of various scenes that are rendered under diverse, controllable lighting conditions. To this end, we use Infinigen~\cite{raistrick2023infinite,raistrick2024infinigen} to construct a large-scale synthetic dataset of photorealistic room-scale environments. Our dataset comprises over 300 indoor scenes spanning a variety of room types, including dining rooms, kitchens, and bedrooms. Each scene is rendered under 30 distinct lighting conditions, resulting in more than 9,000 unique scene-lighting combinations.
In total, the dataset consists of 108,000 unique videos of 81 frames each. We render all images using Blender's Cycles engine at a resolution of $720\times1280$ pixels. We include samples from our dataset in the supplemental material.

Given a training scene, for each training camera $\pi$ we render a set of $L+1$ ``one-light-at-a-time'' (OLAT) images $\mathcal{I}^\pi = \{I^\pi_\ell\}_{\ell=0}^L$, where $L$ denotes the number of artificial light sources, and the additional light corresponds to the unobserved outside light sources. The $\ell$th image is rendered by turning off all lights, and only turning on the $\ell$th light source.
We render and store these images in high dynamic range linear color space, enabling the synthesis of arbitrary target lighting conditions during training via linear combinations of the basis images:
\begin{equation}
\label{eq:olat_combo}
    I^\pi_\text{target} = \gamma\lft(e\cdot\sum_{\ell=0}^L \mathbf{c}_\ell I^\pi_\ell\rgt)\,.
\end{equation}
Here, $\mathbf{c}_\ell \in [0, 1]^3$ represents the intensity of the $\ell$th light source, $e$ is a selected exposure value, and $\gamma$ is a simple sRGB tone mapping function~\cite{anderson1996proposal}. The OLAT data synthesis pipeline is illustrated in Figure~\ref{fig:data_pipeline}.

During training, the intensities $\{\mathbf{c}_\ell\}_{\ell=0}^L$ are selected randomly. Every light source is independently set to be on or off ($\mathbf{c}_\ell$=0) with equal probabilities, with at least one light being on (\ie, we reject samples where all lights are off). For lights that are on, $\mathbf{c}_\ell$ is chosen to reflect typical indoor scene illuminants: for 80\% of iterations the color values are set to the color of blackbody radiation with temperature uniformly chosen between 2500K and 10500K, and for the remaining 20\% of iterations it is assigned a uniformly random color in HSV space. This sampling strategy encourages the model to learn a robust representation of lighting without overfitting to specific lighting patterns.

\myparagraph{Camera settings.} Our training camera paths are designed to capture large portions of the scene. We use equisolid fisheye cameras with a $180^\circ$ field of view. Each elliptical rendering path consists of $90$ evenly-spaced cameras.%

\section{Experiments}
\label{sec:results}

We evaluate our method on two data sources to thoroughly assess its performance, robustness, and generalization capabilities. Our first set of experiments shows qualitative results on real-world indoor scenes from the Eyeful Tower dataset~\cite{xu2023vr}. We show that our method is successful at handling the domain gap between synthetic and real data: even though our relighting model is trained solely on synthetic data, it is able to relight real scenes. Additionally, while our training data is diverse, it contains a limited number of lighting fixtures. However, we demonstrate that our approach is able to generalize to unseen light shapes and render plausible relit videos which can be distilled into 3D representations. We conduct a large-scale user study to evaluate the performance of our method against LightLab~\cite{lightlab}, and the improved Cosmos version of DiffusionRenderer~\cite{DiffusionRenderer2025}.

 Our second dataset consists of held-out synthetic scenes rendered via Infinigen, that follows the same distribution as our training data. We use this dataset for qualitatively and quantitatively testing our method, and we perform an ablation study on this data to validate our design of the diffusion model and its training.

\begin{figure*}[t]
\vspace{2ex}
  \centering
\includegraphics[width=\linewidth]{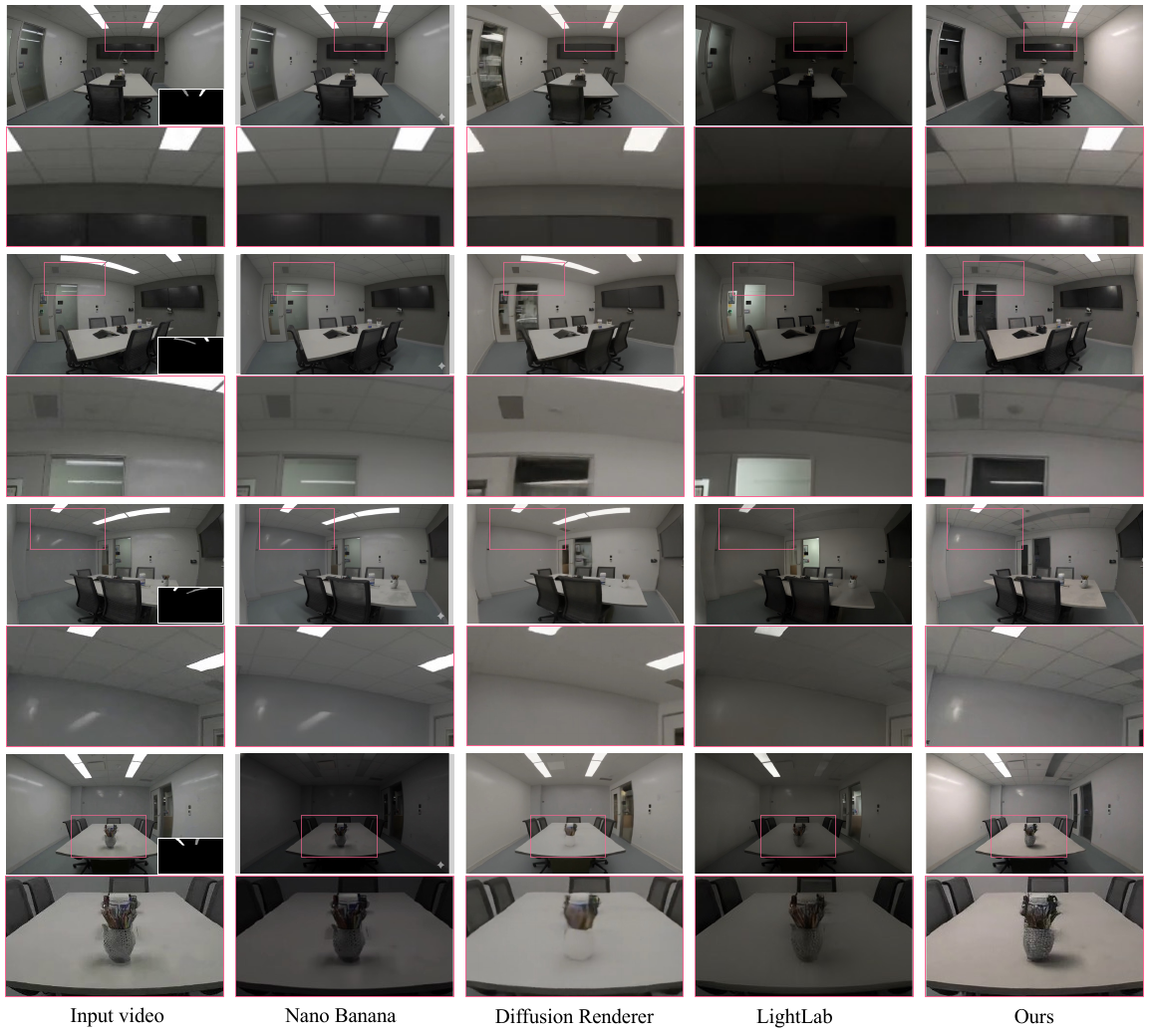}
\caption{\noindent
\textbf{Real-world relighting results on the Eyeful Tower dataset.} Each pair of rows presents a video frame (with the target lighting condition indicated in the bottom-right corner of the input frame) and the corresponding relighting results generated by Nano Banana, DiffusionRenderer (Cosmos version), LightLab, and our method. The top row displays the full scene, while the bottom row shows a zoomed-in region (highlighted with a colored frame). In this setup, both the indoor light near the window and the external light source have been turned off, so no illumination should enter through the window.
In the \textit{first} frame, our method successfully removes the TV reflection caused by the now-disabled light source, whereas the baselines fail to do so. In the \textit{second} frame, our model accurately preserves the remaining valid reflection while removing the one associated with the turned-off light, and also eliminates external illumination. In contrast, Nano Banana and LightLab produce overly dark reflections and fail to suppress the external light. In the \textit{third} frame, only our method removes the wall reflection while preserving surface textures, whereas DiffusionRenderer struggles to control lighting conditions effectively, as its environment map lacks the necessary detail for precise light source conditioning. In the \textit{fourth} frame, our method generates realistic edits with plausible global illumination effects, such as the shadow cast by the object on the table, which other methods fail to reproduce. Additionally, DiffusionRenderer's rendered results tend to be overly
diffuse due to the inherent difficulty in precise material property estimation. See our project webpage for interactive comparisons.
}\label{fig:real_world_results}
\vspace{2ex}
\end{figure*}

\begin{figure}
    \centering
    \includegraphics[width=1\linewidth]{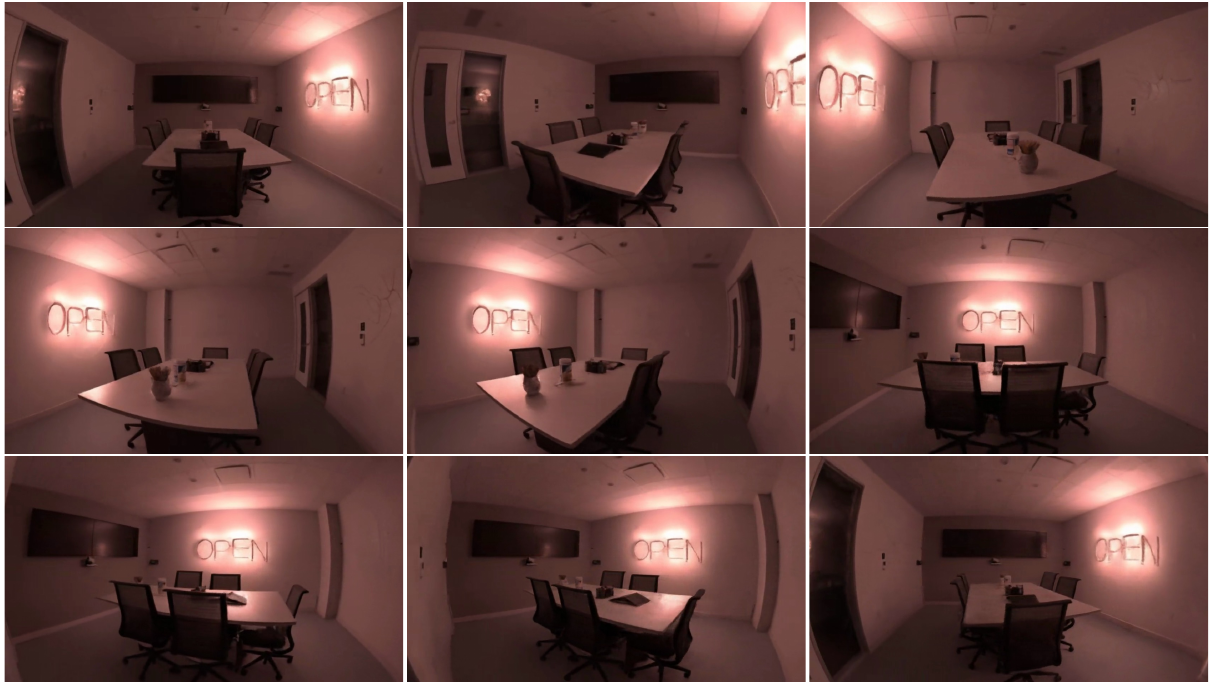}
    \vspace{-.25in}
    \caption{\textbf{3D light insertion.} We add a new out-of-distribution light source to a real reconstructed 3D scene. Despite the training data not containing this type of light, and despite the relighting model being trained only using synthetic data, our method still obtains realistic results. Note in particular the consistent global illumination effects observed for example in the shadows of the objects on the table.}
    \label{fig:neon}
\end{figure}
\subsection{Implementation Details}
\label{sec:implementation}

\myparagraph{Relighting model.} Our video diffusion model, based on Wan 2.2 \cite{wan2025}, is finetuned from the standard weights for 11,200 iterations on our relighting dataset. We use a batch size of 64, processing video clips of 81 frames at a resolution of $352 \times 640$. Training takes $\sim$36 hours on 64 NVIDIA H100 GPUs.

\myparagraph{3D distillation.} To distill the relit video into the final relit Zip-NeRF model, we optimize it using the standard training pipeline for $2000$ iterations using the Adam optimizer with a learning rate of $10^{-3}$. This takes $\sim$10 minutes on 8 NVIDIA H100 GPUs.

We plan to release our code and trained relighting model weights upon acceptance of our paper.

\subsection{Baselines}

To the best of our knowledge, our method is the first to relight full 3D scenes with fine-grained, per-source control over intensity and color. Prior 2D image editors such as LightLab~\cite{lightlab} provide similar controls but operate on one image at a time and thus do not maintain multi-view or temporal consistency. Conversely, scene editors like DiffusionRenderer ensure temporal consistency but offer limited control and cannot target specific light sources. Text-driven video relighting such as Light-A-Video~\cite{Zhou_2025_ICCV} can modify lighting using text prompts, but only for certain video types and without explicit per-light control. Because no existing approach offers controllable 3D scene relighting, we introduce a new benchmark for this task and compare against the closest available baselines from these adjacent settings.

\myparagraph{Conditioning for LightLab.} %
LightLab only supports editing a single light source. Therefore in order to modify multiple light sources we apply their model multiple times, each only adjusting the color or intensity of one light source.

\myparagraph{Conditioning for DiffusionRenderer.} DiffusionRenderer relights scenes based on an environment map. Therefore for each target illumination we use our relit NeRF to render a target environment map. Since DiffusionRenderer supports at most 57 frames, we evaluate it on the first 57 frames of our camera path and report its results on this subset.

\myparagraph{Conditioning for Light-A-Video.} Light-A-Video is conditioned on text. For each sequence, we use prompts that first describe the unobserved illumination, and then specify the color of each observed light and whether it is on or off. We employ its WAN 2.1-based variation, as it handles longer videos and higher resolution. 

\begin{figure}
    \centering
    \includegraphics[width=\linewidth]{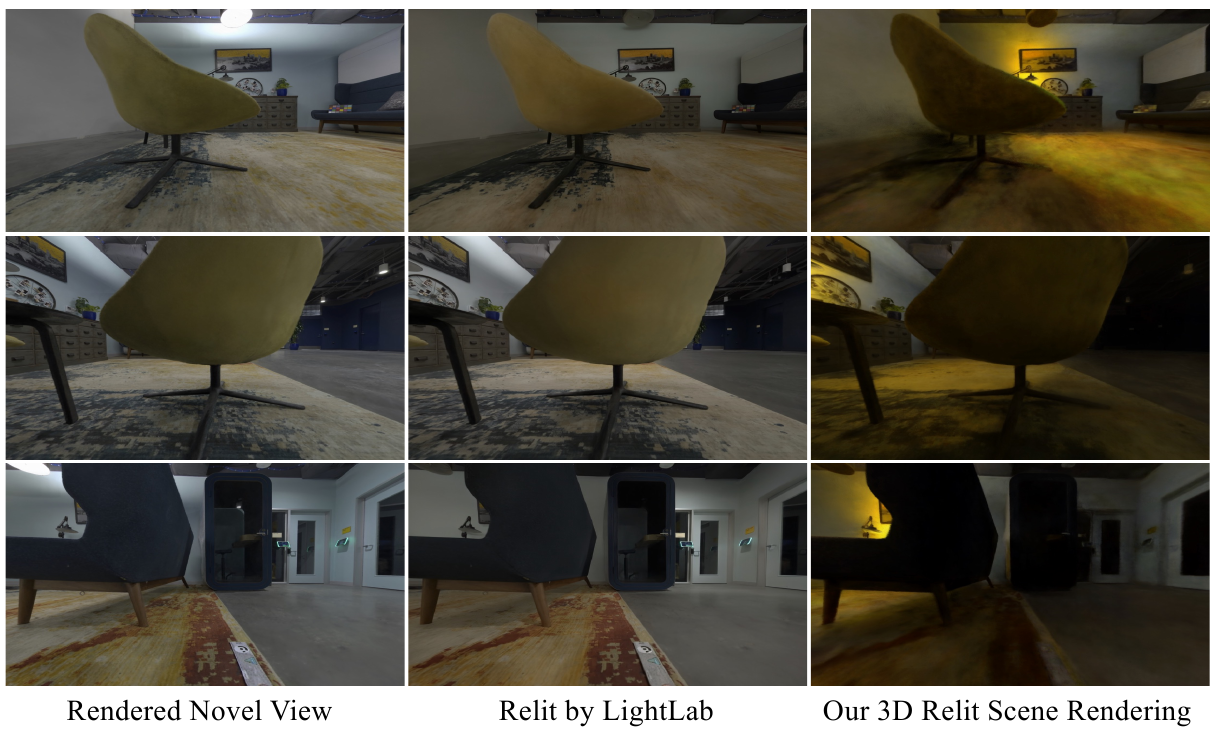}
    \vspace{-.25in}
    \caption{\textbf{Comparison of our relit novel views with those obtained by LightLab.} We show novel view renderings of the original scene (left), compared with the same image relit by LightLab (center), and the corresponding viewpoint rendered from our 3D relit scene (right). 
     Note that LightLab fails to relight the scene whether the target light is visible (top row), as well as when it is not directly visible (bottom row). 
     }
    \label{fig:realresults_nvs}
\end{figure}

\subsection{Real-World Scene Relighting Results}
\label{sec:real_world_results}

We evaluate our method on captured scenes from the Eyeful Tower dataset~\cite{xu2023vr}. This datasets consists of high-quality captures of indoor environments. We show qualitative results of our entire pipeline (reconstruction, relighting, and 3D distillation) in Figures~\ref{fig:teaser-fig},~\ref{fig:pipeline}, ~\ref{fig:real_world_results}, and ~\ref{fig:neon}. Perhaps surprisingly, our model was never trained on light fixture shapes present in this data, such as the large rectangular ceiling light sources in the top half of Figure~\ref{fig:real_world_results}, but it successfully generalizes to out-of-distribution lighting conditions.

Figure~\ref{fig:realresults_nvs} shows our relit 3D model renders under different lighting conditions, demonstrating the ability of our full pipeline to render the scene under different illumination conditions,
compared with LightLab~\cite{lightlab}, a single-image relighting method, which fails relighting this image, and therefore cannot be used for 3D relighting. 

\begin{figure*}[h]
  \centering
\includegraphics[width=\textwidth]{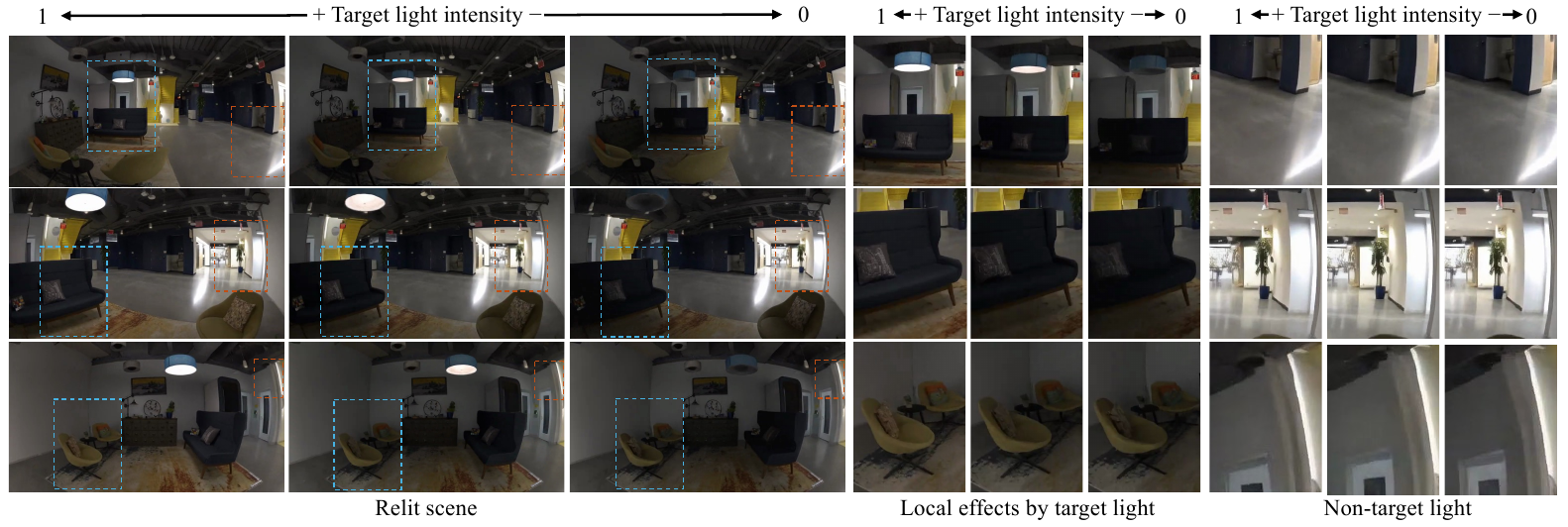}
  \caption{\textbf{Control of light intensity.} Our method enables fine control of the target intensity. The leftmost three columns display the relit scenes under the same target lights but different target light intensities. The regions primarily lit by the modified target light are marked in blue boxes and those dominated by light that is not modified are in orange. The middle three columns and the rightmost three columns are insets corresponding to the blue and orange boxed regions, respectively. Note the dramatic changes in local appearance within the target-lit regions, contrasted with the nearly constant illumination in regions primarily lit by unmodified light sources. 
  }\label{fig:intensity condition}
\end{figure*}

Figures~\ref{fig:intensity condition} and~\ref{fig:kitchen} show the results of our relighting model, demonstrating that our diffusion model provides fine-grained control over the illumination of the scene, rendering realistic global illumination effects like cast shadows and reflections.

Finally, our project page contains many more video results and comparisons with other methods.

\myparagraph{User study.} We conducted a user study comparing our method with LightLab~\cite{lightlab}, DiffusionRenderer~\cite{DiffusionRenderer2025}, and the original input video (which has high quality but does not respect the target lighting). Users were shown pairs of videos with identical camera paths and a text description of the target lighting condition, then asked to evaluate frame-to-frame consistency, object identity preservation, and lighting similarity.

Table \ref{tab:user_study} presents the results. Our method achieves an overall win rate of 79.0\%, with strong performance across all criteria: 88.0\% for lighting similarity, 75.9\% for temporal consistency, and 73.1\% for identity preservation. We outperform both LightLab and DiffusionRenderer with 84.3\% win        rates, and achieve 68.5\% against the artifact-free input, demonstrating that users prefer our improved lighting fidelity. Details are provided in the supplementary material.

\begin{table}[t!]
    \small
    \centering
    \caption{\textbf{User Study on Real-World Scene Video Relighting}. We conduct a human preference study evaluating three critical aspects: frame-to-frame consistency (F2F-C), object identity preservation (O-Pres), and lighting similarity to the provided lighting description (L-Sim). We report the win rates (\%) of our method against each competing approach.} 
    \vspace{-0.1in}
    \resizebox{\linewidth}{!}{
    \begin{tabular}{l |c c c}
    \toprule
         \textbf{Opponent Method}
   &\textbf{ F2F-C. $\uparrow$}& \textbf{{O-Pres.}$\uparrow$} & \textbf{{L-Sim.}$\uparrow$}   \\
        \midrule
        Input & 61.1 & 55.6 & 88.9 \\
        LightLab~\cite{lightlab} & 86.1 & 80.6 & 86.1 \\
        DiffusionRenderer~\cite{DiffusionRenderer2025} & 80.6 & 83.3 & 88.9 \\
        \midrule
        Ours Overall & 75.9 & 73.1 & 88.0 \\
        \bottomrule
    \end{tabular}
    }
    \label{tab:user_study}
\end{table}

\subsection{Synthetic Scene Relighting Results}
\label{sec:synthetic_results}

\begin{table}[t!]
\centering
\caption{\small \textbf{Quantitative evaluation of per-frame relighting. }Our method outperforms DiffusionRenderer, which is a method capable of relighting 3D scenes, but does not allow as much control as our model. The ``input video'' entry denotes the metrics computed between the input video and the relit video, for reference.}
\setlength{\tabcolsep}{4pt}
\footnotesize
\vspace{-0.1in}
\resizebox{\linewidth}{!}{
\begin{tabular}{l|ccc}
\toprule
\textbf{Method} & \textbf{PSNR~$\uparrow$} & \textbf{SSIM~$\uparrow$} & \textbf{LPIPS~$\downarrow$} \\
\midrule
Input video                               & 13.11    & 0.679    & 0.252     \\
DiffusionRenderer~\cite{DiffusionRenderer2025}               & 15.17 & 0.553 & 0.668  \\
LightLab~\cite{lightlab}                           & 11.95    & 0.699    & 0.281      \\
Light-A-Video~\cite{Zhou_2025_ICCV}                    &  10.88   & 0.525    & 0.663      \\
\textbf{Ours}                                 & \textbf{24.77} & \textbf{0.741}  & \textbf{0.130}    \\
\bottomrule
\end{tabular}
}

\label{tab:relighting}
\end{table}

\begin{figure*}[!h]
    \centering
    \resizebox{\textwidth}{!}{
    \includegraphics[width=\linewidth]{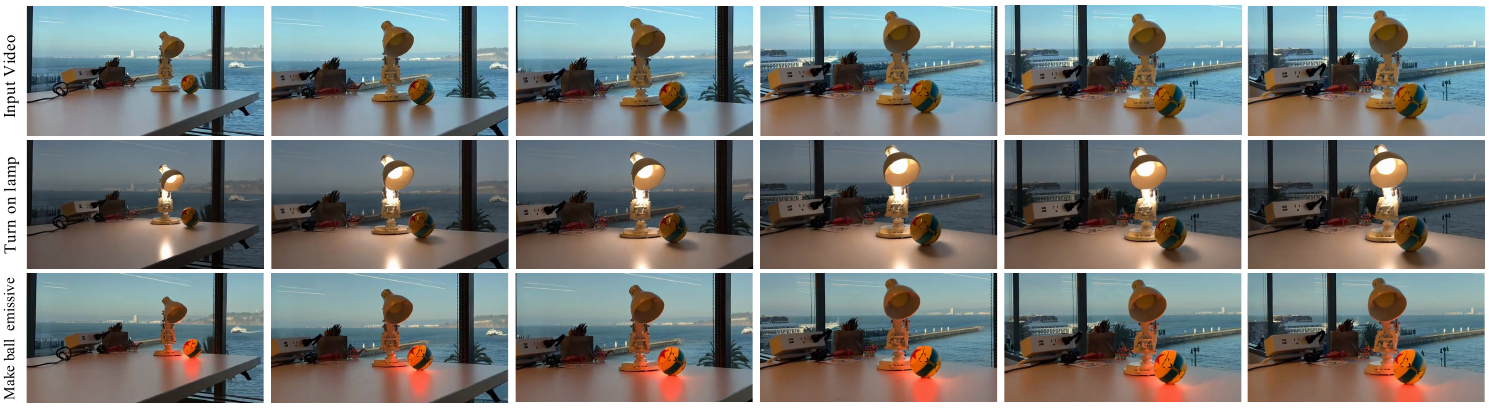}}
    \vspace{-0.2in}
    \caption{\textbf{Real-world video relighting.} Our model is capable of relighting real-world videos (top row). In the first example (middle row), the lamp is turned on and its color set to yellow; note the cast shadow of the ball and the reflectance on the surface of the ball. In the second example (bottom row), our model generalizes and successfully transforms the ball into a light source, thereby producing plausible global illumination effects.}
    \label{fig:real_video_relit}
\end{figure*}
In Table~\ref{tab:relighting} we report our relighting results compared with DiffusionRenderer (Cosmos version)~\cite{DiffusionRenderer2025}, LightLab~\cite{lightlab} and Light-A-Video (WAN 2.1 version)~\cite{Zhou_2025_ICCV}. Although DiffusionRenderer can theoretically relight 3D scenes, its control over light sources is extremely limited. As shown in Figure~\ref{fig:infinigen_show}, DiffusionRenderer fails to accurately modify visible light sources as well as exterior light sources that are not directly visible.
While LightLab can edit the color of the lights, its control is limited, and it fails to turn on exterior light sources.

\subsection{Ablation Studies}
\label{sec:ablation}

We perform an ablation study to validate two of our major design choices in our relighting model's architecture and training process. The results are reported in Table~\ref{tab:ablation}. First, we replace our conditioning approach described in Section~\ref{sec:relighting_model} with concatenation along the channel dimension, as done in LightLab~\cite{lightlab}, which results in a significant degradation in relighting quality. Second, we replace the sampling strategy described in Section~\ref{sec:relighting_training} with standard uniform samples $t \sim \mathrm{U}([0, 1])$, which also leads to a degradation in the accuracy of the relit results.

\subsection{Additional Applications}

\myparagraph{3D light insertion.} Our method can also be used to insert new light sources with arbitrary shapes into a 3D scene. Figure~\ref{fig:neon} shows an example where we insert into the scene a new ``neon'' light. This is done by projecting the target light shape onto the planar wall, and using it as the conditioning lights. Despite being trained on a limited set of lighting conditions, which does not include this type of light, our method produces plausible-looking results.

\myparagraph{Video relighting.}
Although our full method is designed for relighting full 3D scenes, the video relighting component can be used on its own, enabling video relighting without requiring 3D reconstruction. We show results of our video relighting model in Figure~\ref{fig:real_video_relit}. Our relighting model is capable of removing exterior lights and their associated lighting effects such as reflections and cast shadows.

\begin{table}[t]
    \small
    \centering
        \caption{\textbf{Ablation study of model design.} We observe that replacing our conditioning approach with simple channel-wise concatenation, or removing the high-noise training bias, significantly degrades relighting performance.}
        \vspace{-0.1in}
    \begin{tabular}{l | c c c c}
    \toprule
         & \textbf{PSNR}~$\uparrow$ & \textbf{SSIM}~$\uparrow$ & \textbf{LPIPS}~$\downarrow$  \\
        \midrule
        Channel concatenation & 17.06 & 0.648 & 0.172 \\
        
        w.o.\ high noise bias  & 23.82 & 0.637 & 0.192 \\
        \textbf{Ours} & \textbf{24.77} & \textbf{0.741} &\textbf{0.130} \\
        \bottomrule
        
    \end{tabular}
    \label{tab:ablation}
\end{table}

\section{Conclusion}
\label{sec:conclusion}

We have presented a system to relight 3D reconstructions of entire scenes. Our technique renders a video from the 3D scene representation, relights it using a video-to-video diffusion model conditioned on user-specified target lighting, and distills this relit video back into the 3D reconstruction to generate a relit 3D scene representation.

Our experiments have demonstrated that our approach produces photorealistic relit 3D representations of real-world scenes, faithfully representing global illumination effects such as cast shadows and specular highlights, which prior methods often struggle with. This opens up opportunities for editing and using 3D scene reconstructions as controllable graphics assets.

\myparagraph{Limitations.} Although our relighting model demonstrates good generalization across various types of light sources, its performance may be limited by the diversity of light source assets used during training, for example in Figure~\ref{fig:neon}, where the light source isn't entirely emissive. This limitation could be addressed by incorporating a broader range of light source assets in the training process.

Additionally, our relighting model relies on the input video having full coverage of the scene, and may exhibit artifacts in other cases. This can be seen for example in the back side of the emissive ball at the bottom of Figure~\ref{fig:real_video_relit}.

\section*{Acknowledgment}
We would like to thank Gene Chou, Rundi Wu, Ruiqi Gao, Songyou Peng and Alexander Raistrick for their valuable help and suggestion with implementation. We also thank Amir Hertz and Eric Tabellion for providing LightLab assets.
We appreciate the insightful discussions with Connie He, Aleksander Holynski, Georgios Kopanas, Janne Kontkanen, Michael Okunev, Ben Poole, Rick Szeliski, and Dimitris Tzionas. 
Finally, we thank all participants who took part in our user study.

{
    \small
    \bibliographystyle{ACM-Reference-Format}
    \bibliography{main}
}

\newlength{\cellW}
\newlength{\cellH}
\newlength{\rowLabelW}

\setlength{\cellW}{0.16\textwidth}

\setlength{\cellH}{0.5625\cellW}

\setlength{\rowLabelW}{0.03\textwidth}

\begin{figure*}[p]
  \centering
  \setlength{\tabcolsep}{2pt}          %
  \renewcommand{\arraystretch}{0.8}    %
    \resizebox{\textwidth}{!}{\begin{tabular}{@{}c*{5}{c}@{}}

    \includegraphics[width=\cellW,height=\cellH,keepaspectratio]{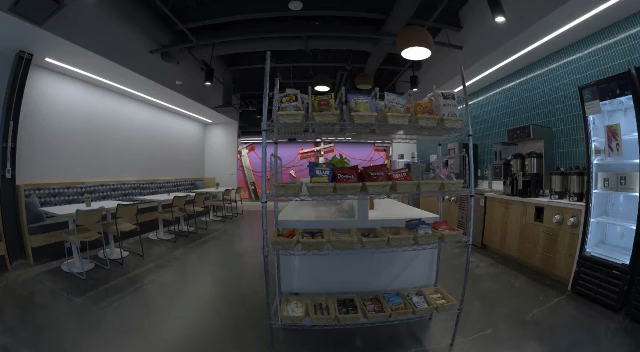} &
    \includegraphics[width=\cellW,height=\cellH,keepaspectratio]{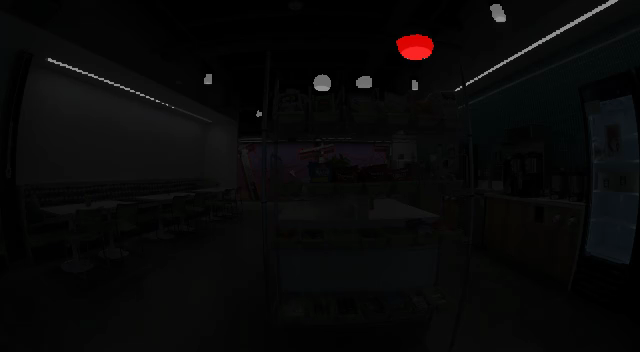} &
    \includegraphics[width=\cellW,height=\cellH,keepaspectratio]{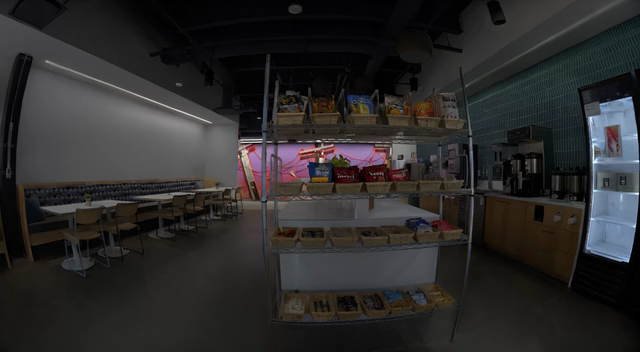} &
    \includegraphics[width=\cellW,height=\cellH,keepaspectratio]{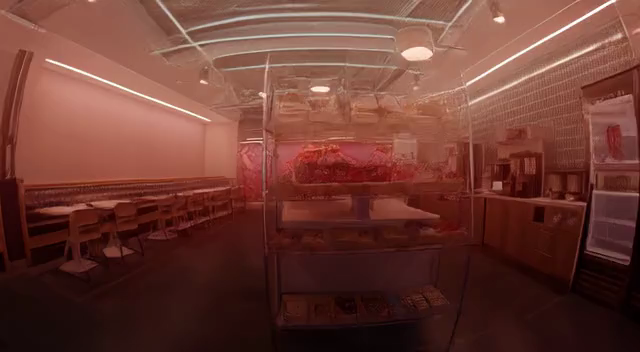} &
    \includegraphics[width=\cellW,height=\cellH,keepaspectratio]{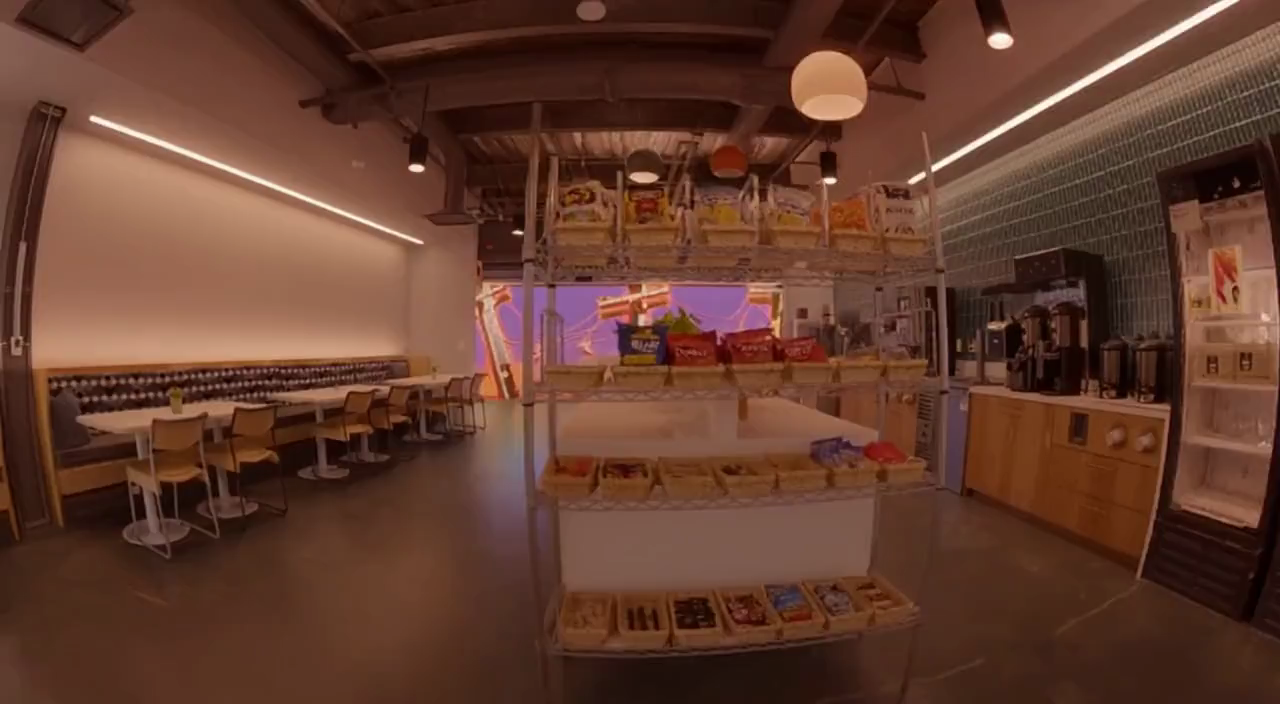} &
    \includegraphics[width=\cellW,height=\cellH,keepaspectratio]{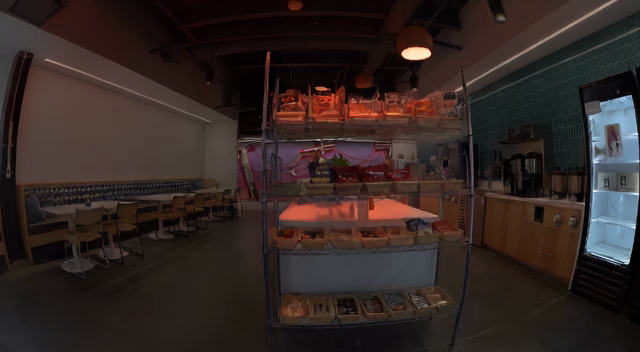}  \\[-2pt]
    \includegraphics[width=\cellW,height=\cellH,keepaspectratio]{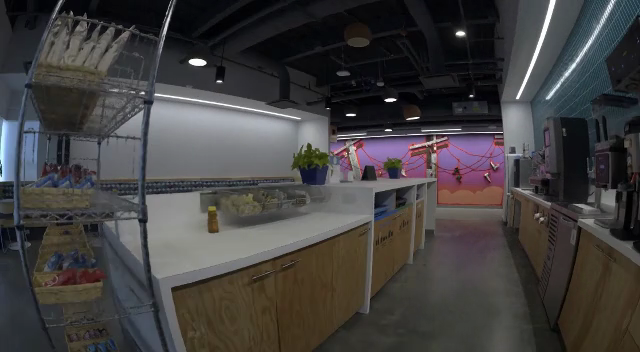} &
    \includegraphics[width=\cellW,height=\cellH,keepaspectratio]{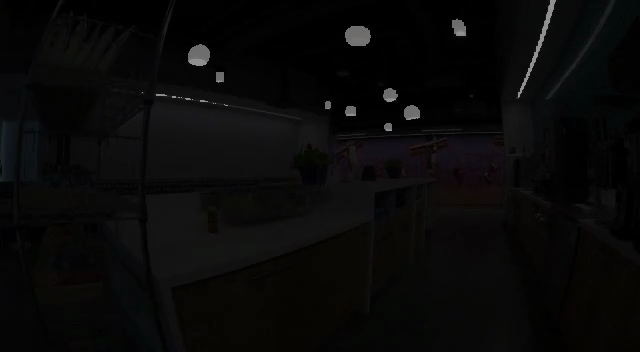} &
    \includegraphics[width=\cellW,height=\cellH,keepaspectratio]{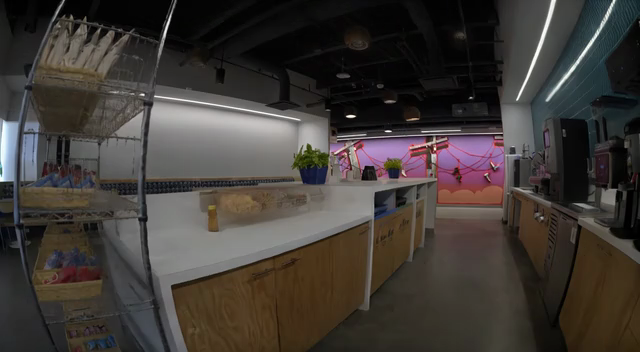} &
    \includegraphics[width=\cellW,height=\cellH,keepaspectratio]{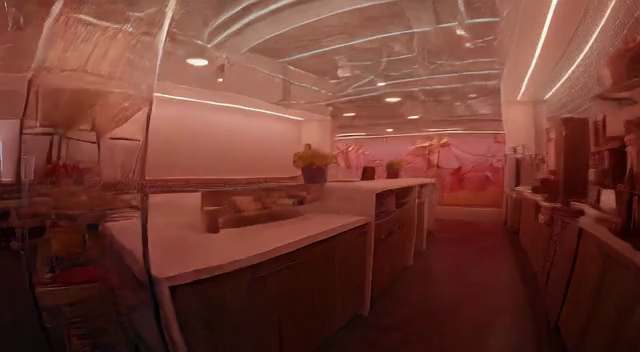} &
    \includegraphics[width=\cellW,height=\cellH,keepaspectratio]{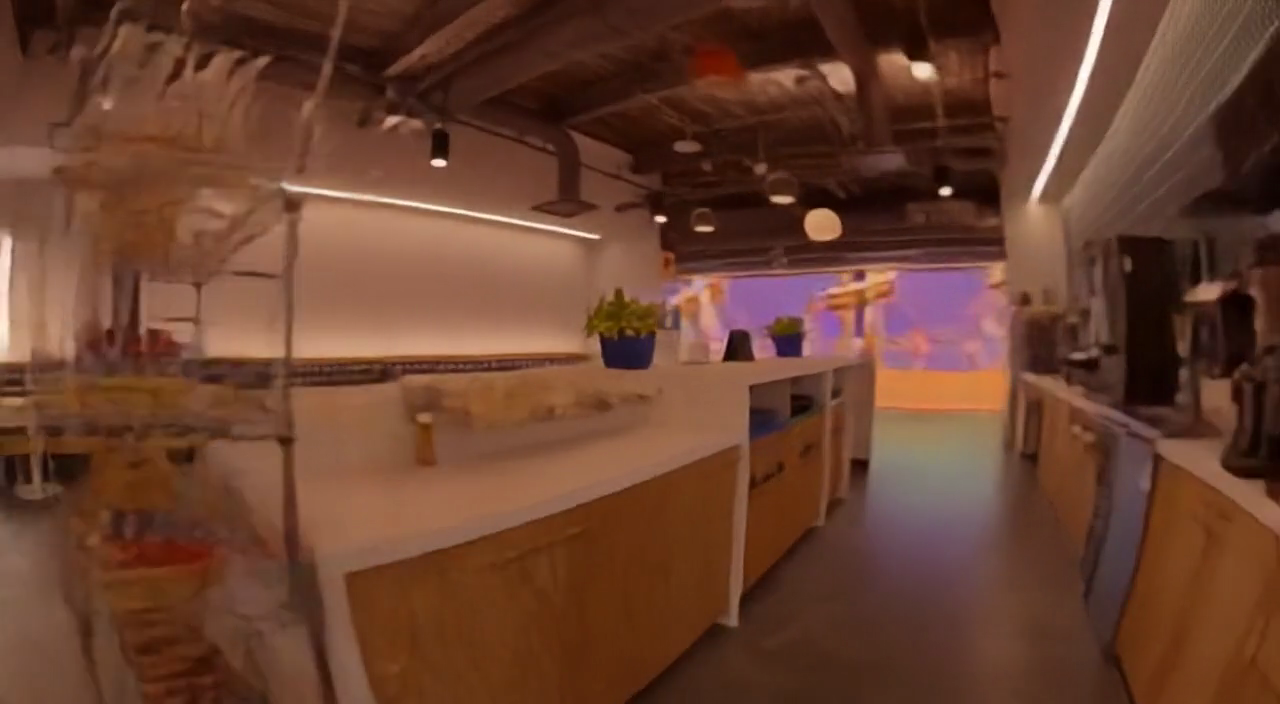} &
    \includegraphics[width=\cellW,height=\cellH,keepaspectratio]{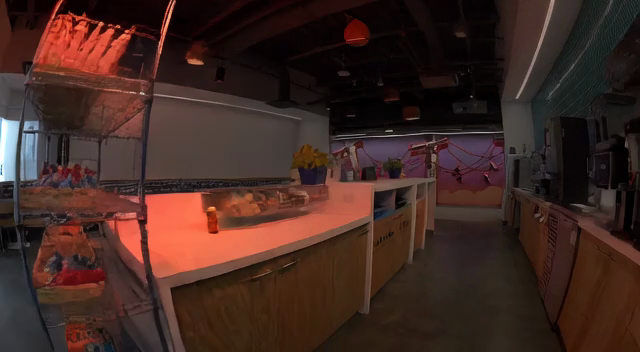} \\[-2pt]
    \includegraphics[width=\cellW,height=\cellH,keepaspectratio]{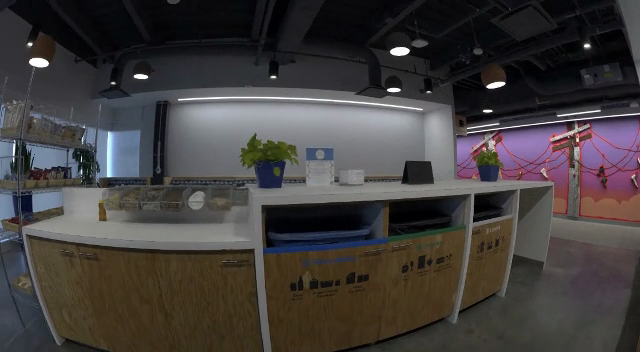} &
    \includegraphics[width=\cellW,height=\cellH,keepaspectratio]{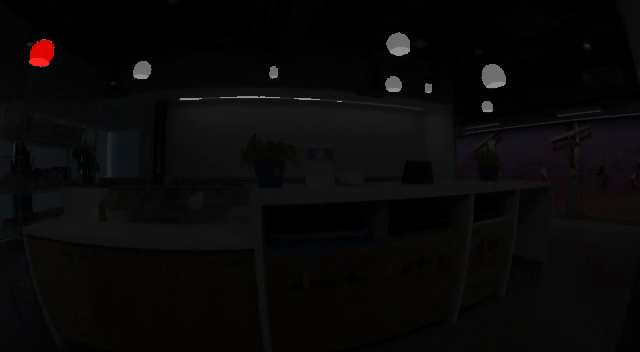} &
    \includegraphics[width=\cellW,height=\cellH,keepaspectratio]{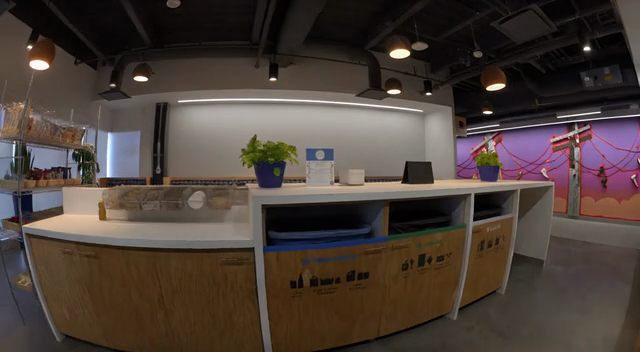} &
    \includegraphics[width=\cellW,height=\cellH,keepaspectratio]{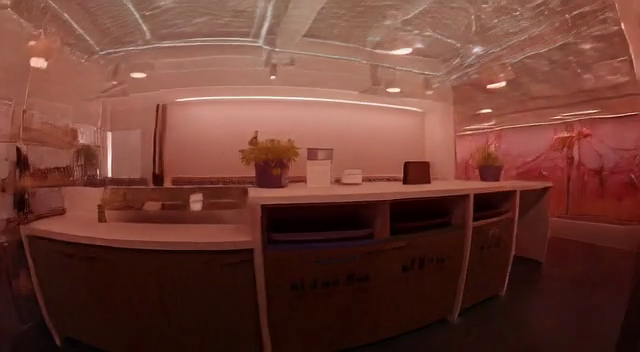} &
    \includegraphics[width=\cellW,height=\cellH,keepaspectratio]{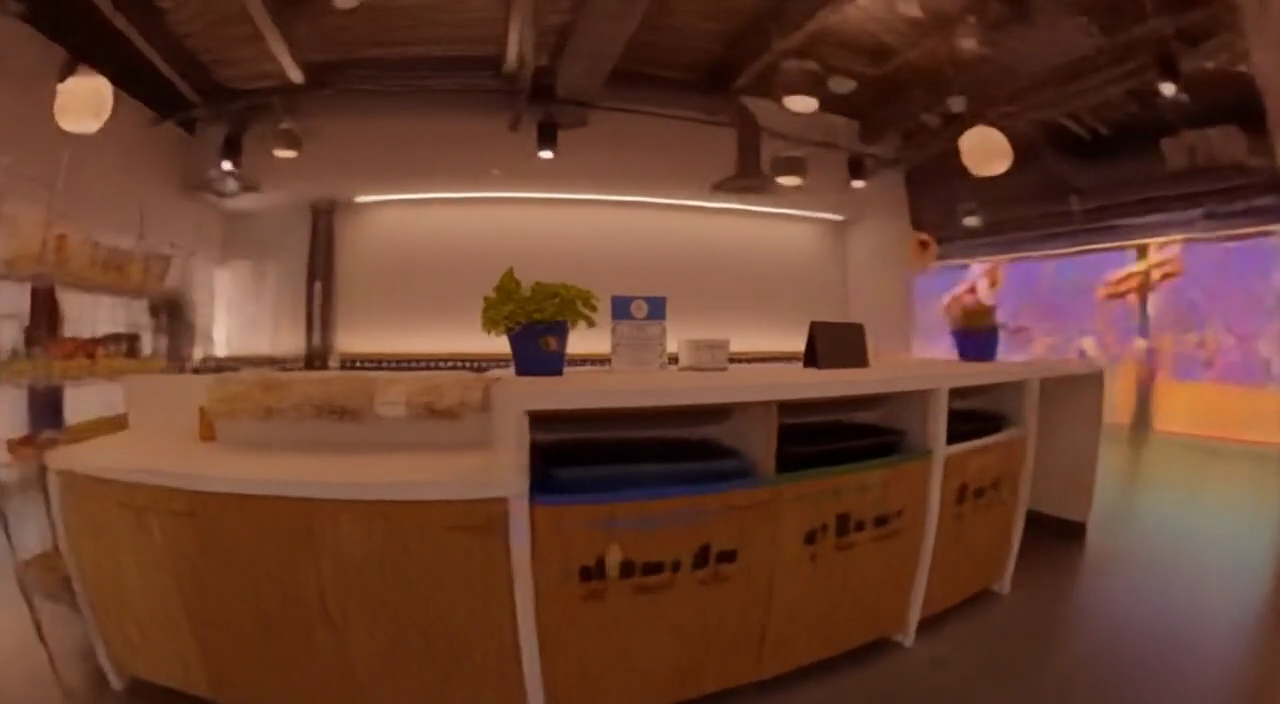} &
    \includegraphics[width=\cellW,height=\cellH,keepaspectratio]{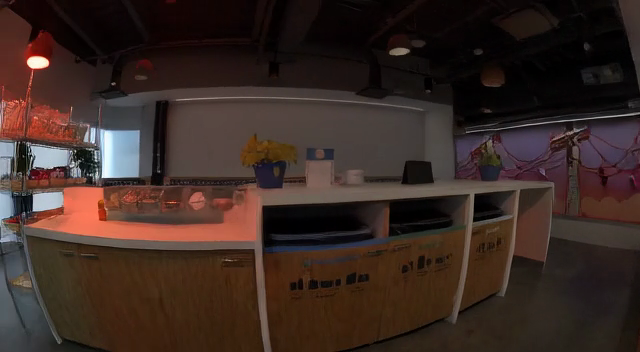} \\[-2pt]
    \includegraphics[width=\cellW,height=\cellH,keepaspectratio]{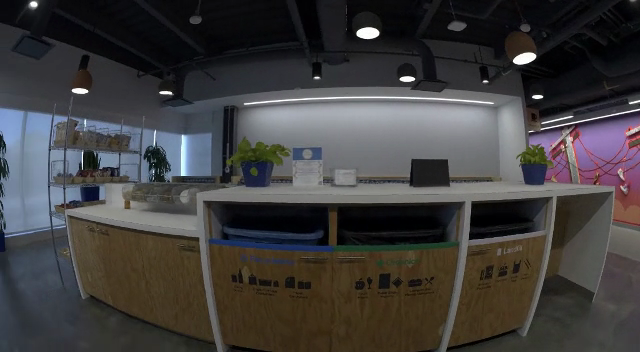} &
    \includegraphics[width=\cellW,height=\cellH,keepaspectratio]{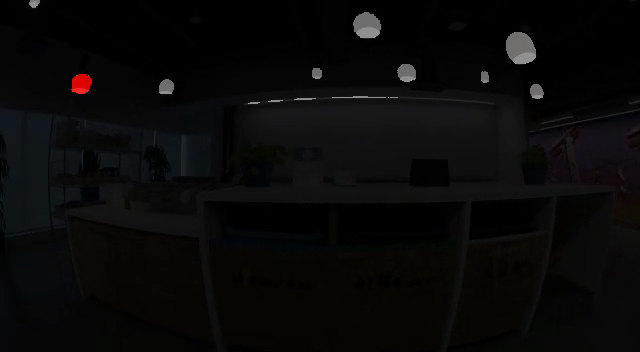} &
    \includegraphics[width=\cellW,height=\cellH,keepaspectratio]{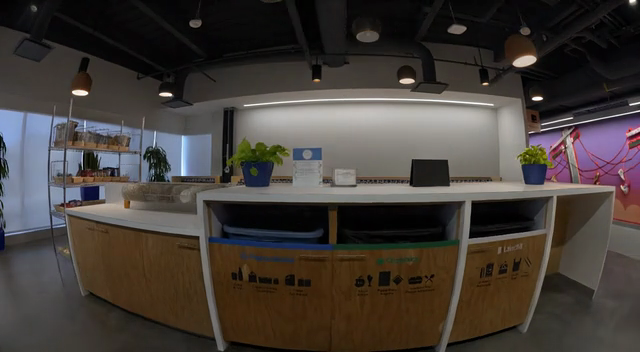} &
    \includegraphics[width=\cellW,height=\cellH,keepaspectratio]{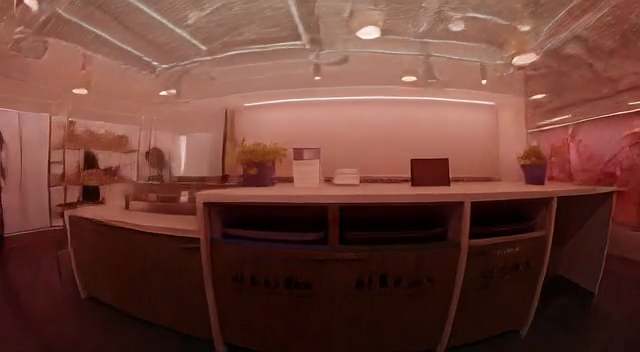} &
    \includegraphics[width=\cellW,height=\cellH,keepaspectratio]{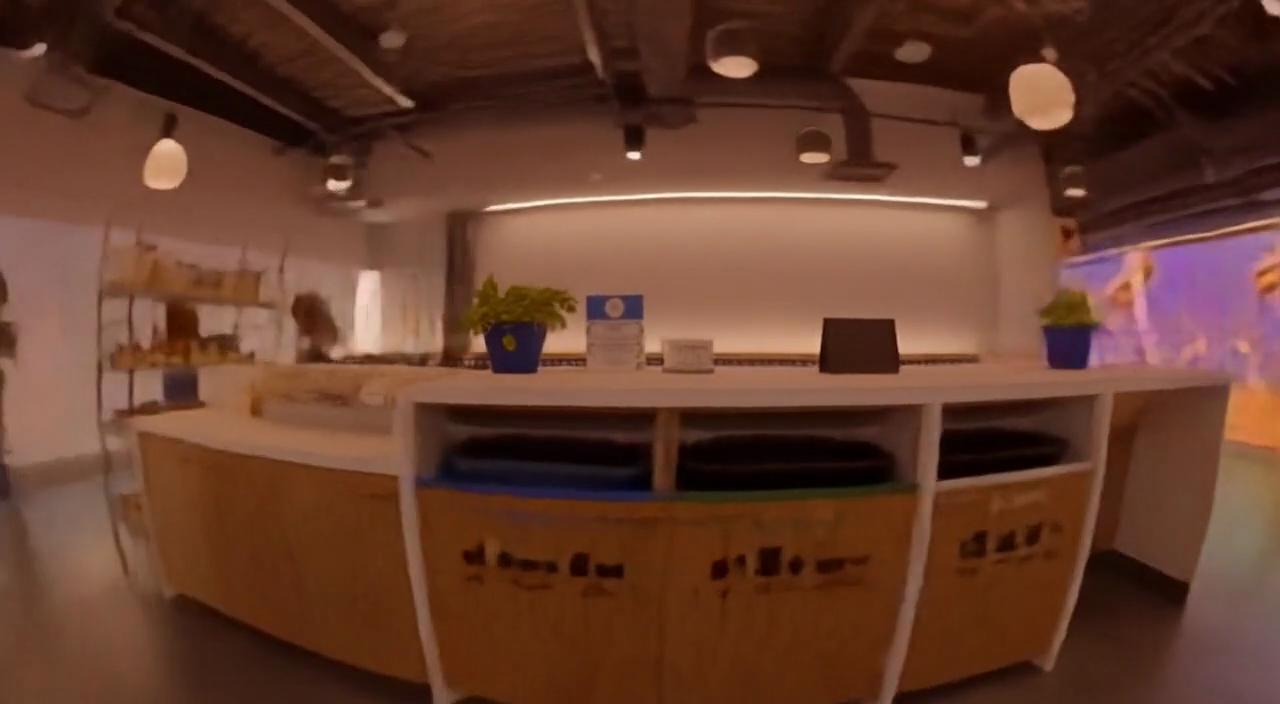} &
    \includegraphics[width=\cellW,height=\cellH,keepaspectratio]{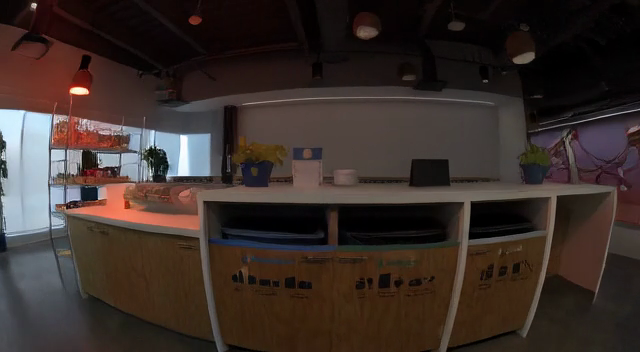} 
    \\[-2pt]
    \includegraphics[width=\cellW,height=\cellH,keepaspectratio]{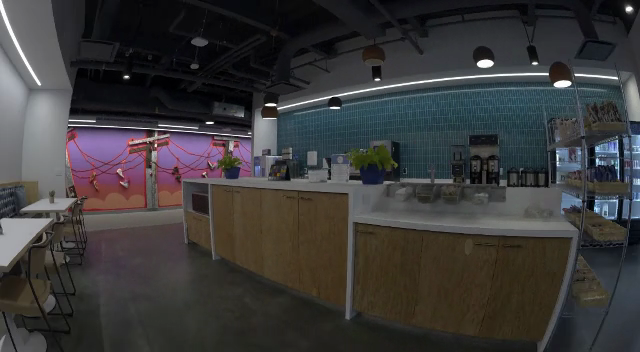} &
    \includegraphics[width=\cellW,height=\cellH,keepaspectratio]{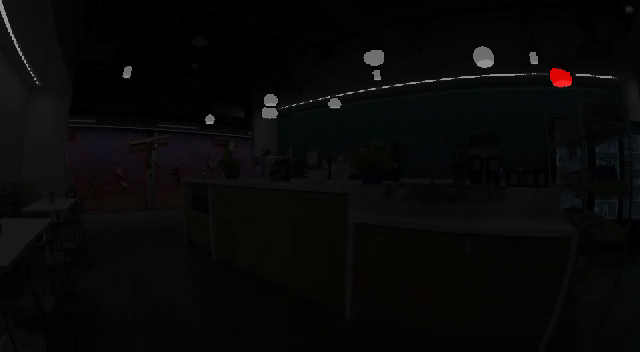} &
    \includegraphics[width=\cellW,height=\cellH,keepaspectratio]{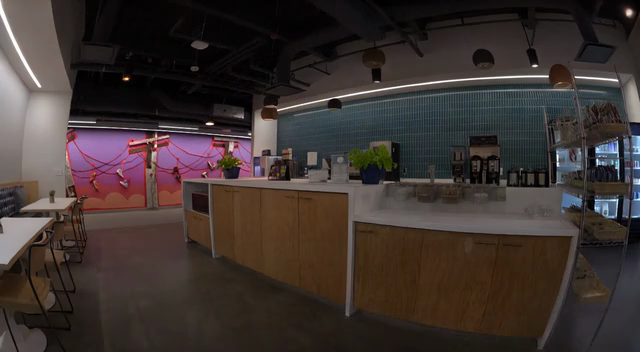} &
    \includegraphics[width=\cellW,height=\cellH,keepaspectratio]{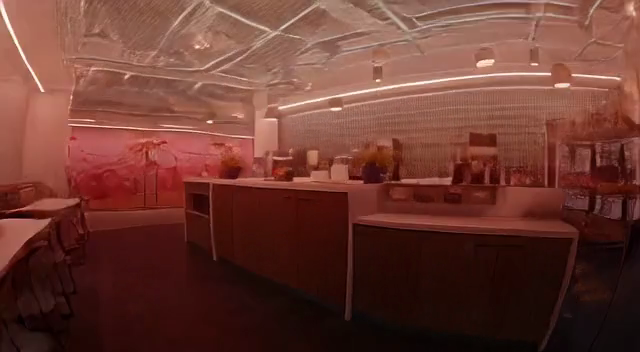} &
    \includegraphics[width=\cellW,height=\cellH,keepaspectratio]{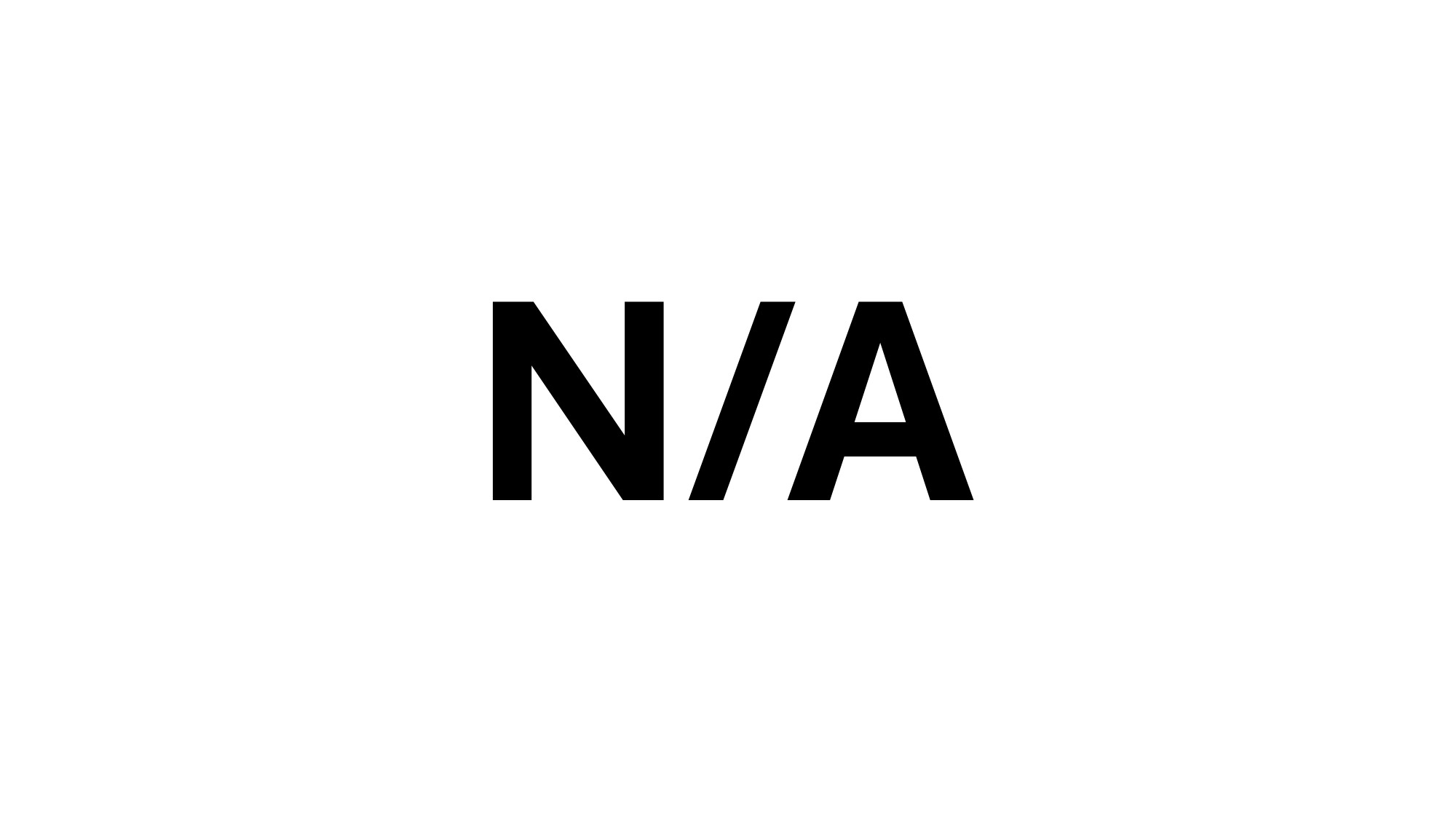} &
    \includegraphics[width=\cellW,height=\cellH,keepaspectratio]{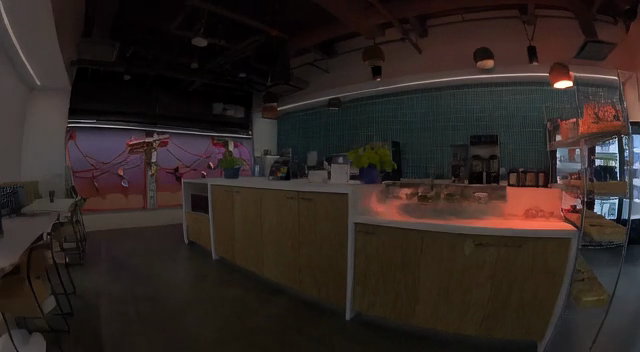} \\
     \parbox[c][\baselineskip][c]{\cellW}{\centering\small Input} &
    \parbox[c][\baselineskip][c]{\cellW}{\centering\small Condition} &
    \parbox[c][\baselineskip][c]{\cellW}{\centering\small LightLab} &
    \parbox[c][\baselineskip][c]{\cellW}{\centering\small Light-A-Video} &
    \parbox[c][\baselineskip][c]{\cellW}{\centering\small DiffusionRenderer} &
    \parbox[c][\baselineskip][c]{\cellW}{\centering\small Ours} 
  \end{tabular}}
  \caption{\textbf{Additional real-world results.} Here we present qualitative comparisons of our method with LightLab, Light-A-Video, and DiffusionRenderer on the \texttt{kitchen} scene from the Eyeful Tower dataset. LightLab, which relights each video frame independently, exhibits noticeable temporal flickering artifacts. Light-A-Video offers limited lighting control and produces low-quality outputs. DiffusionRenderer yields temporally consistent results but suffers from blurriness and poor visual quality. In contrast, our method achieves realistic relighting that adheres closely to the target lighting configuration. Notably, while other methods struggle with scenes involving multiple light sources, our approach effectively controls the lighting solely based on the provided target lights.}
  \label{fig:kitchen}
\end{figure*}

\setlength{\cellW}{0.16\textwidth}

\setlength{\cellH}{0.5625\cellW}

\setlength{\rowLabelW}{0.03\textwidth}

\begin{figure*}[t]
  \centering
  \setlength{\tabcolsep}{2pt}          %
  \renewcommand{\arraystretch}{0.8}    %
  \resizebox{\textwidth}{!}{\begin{tabular}{@{}c*{5}{c}@{}}

    \includegraphics[width=\cellW,height=\cellH,keepaspectratio]{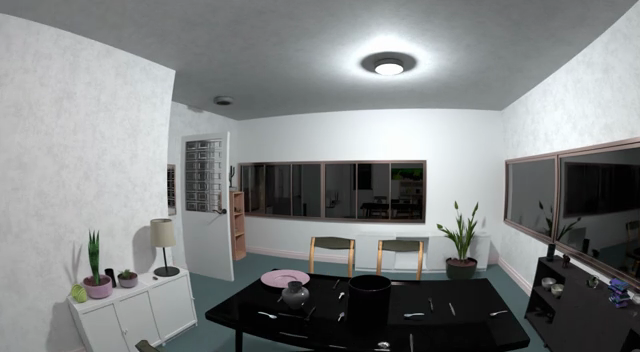} &
    \includegraphics[width=\cellW,height=\cellH,keepaspectratio]{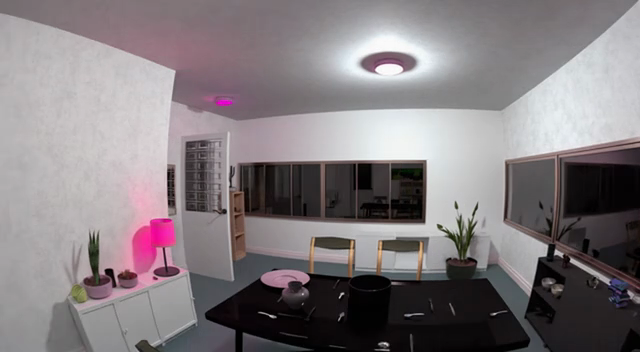} &
    \includegraphics[width=\cellW,height=\cellH,keepaspectratio]{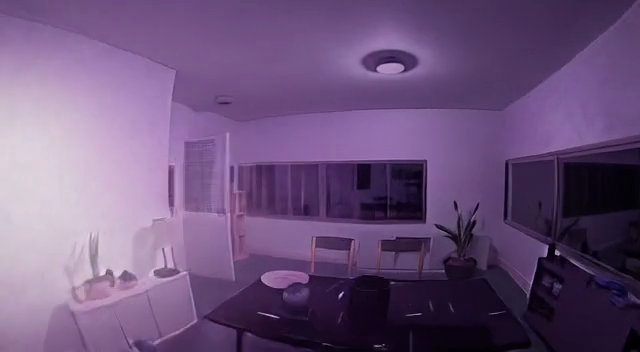} &
    \includegraphics[width=\cellW,height=\cellH,keepaspectratio]{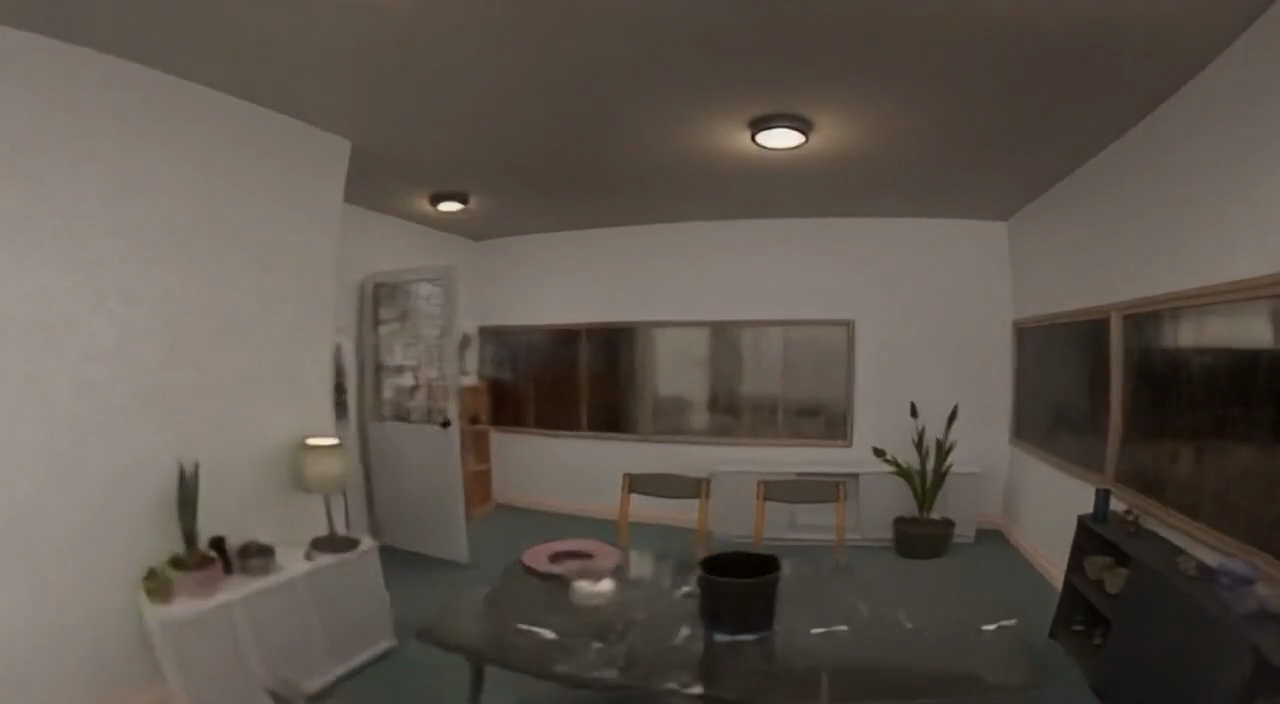} &
    \includegraphics[width=\cellW,height=\cellH,keepaspectratio]{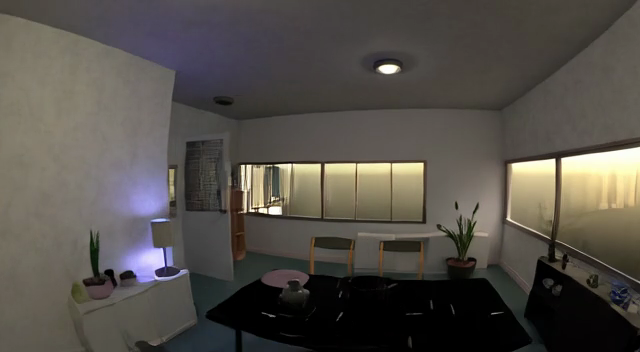} &
    \includegraphics[width=\cellW,height=\cellH,keepaspectratio]{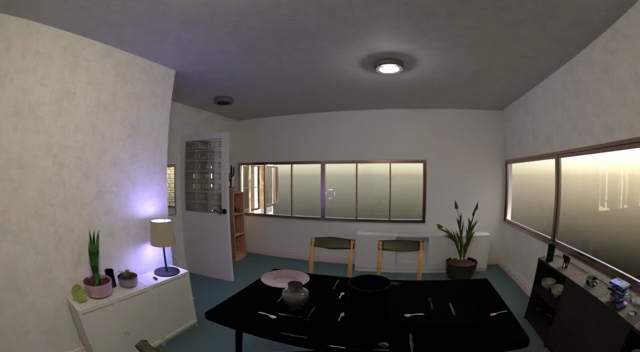}   \\[-2pt]
    \includegraphics[width=\cellW,height=\cellH,keepaspectratio]{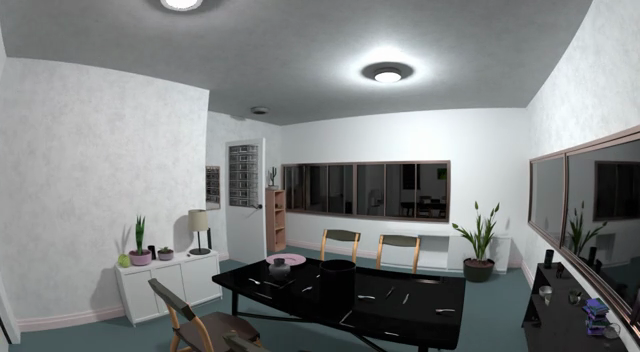} &
    \includegraphics[width=\cellW,height=\cellH,keepaspectratio]{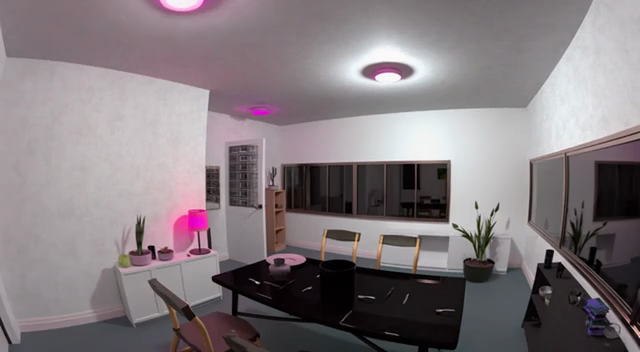} &
    \includegraphics[width=\cellW,height=\cellH,keepaspectratio]{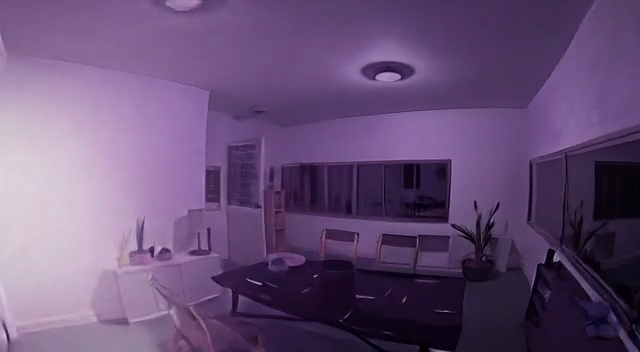} &
    \includegraphics[width=\cellW,height=\cellH,keepaspectratio]{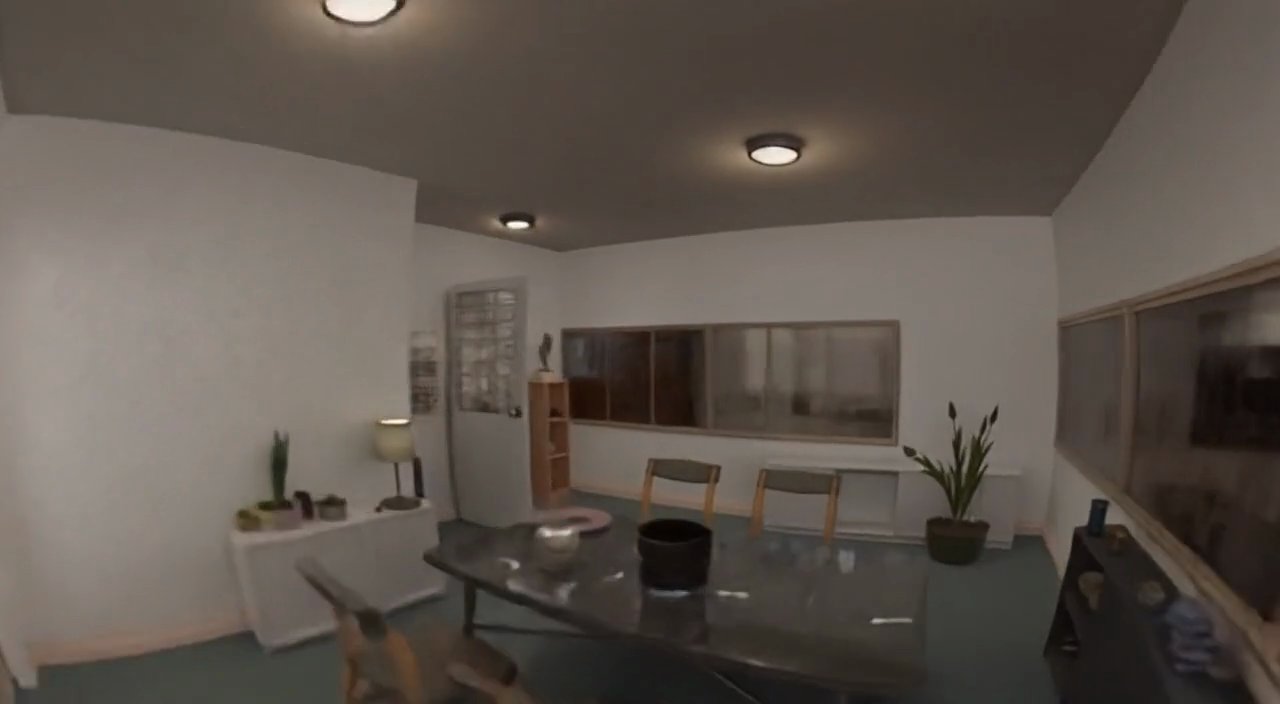} &
    \includegraphics[width=\cellW,height=\cellH,keepaspectratio]{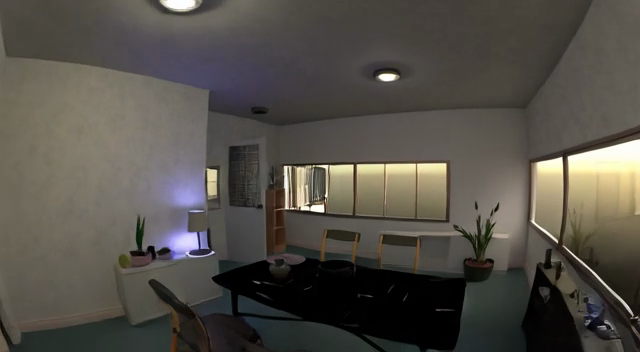} &
    \includegraphics[width=\cellW,height=\cellH,keepaspectratio]{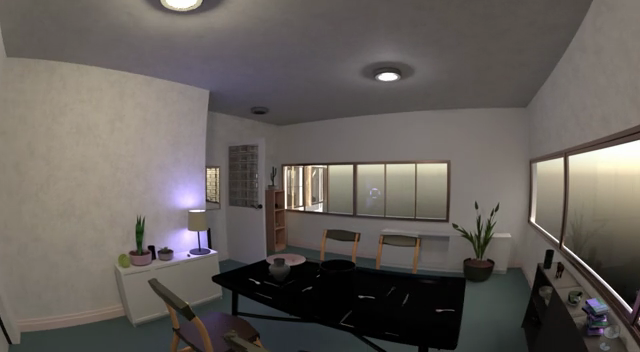}  \\[-2pt]
    \includegraphics[width=\cellW,height=\cellH,keepaspectratio]{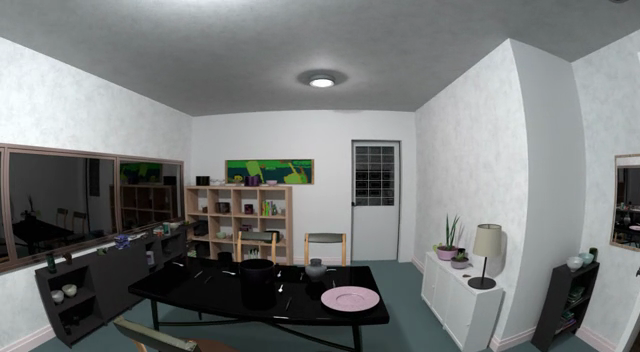} &
    \includegraphics[width=\cellW,height=\cellH,keepaspectratio]{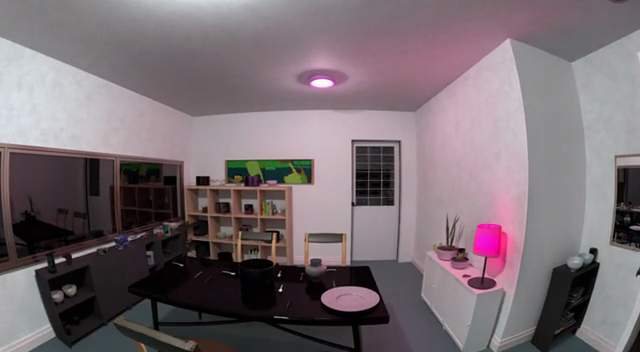} &
    \includegraphics[width=\cellW,height=\cellH,keepaspectratio]{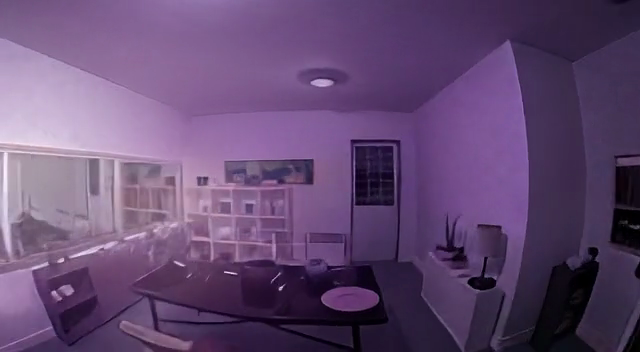} &
    \includegraphics[width=\cellW,height=\cellH,keepaspectratio]{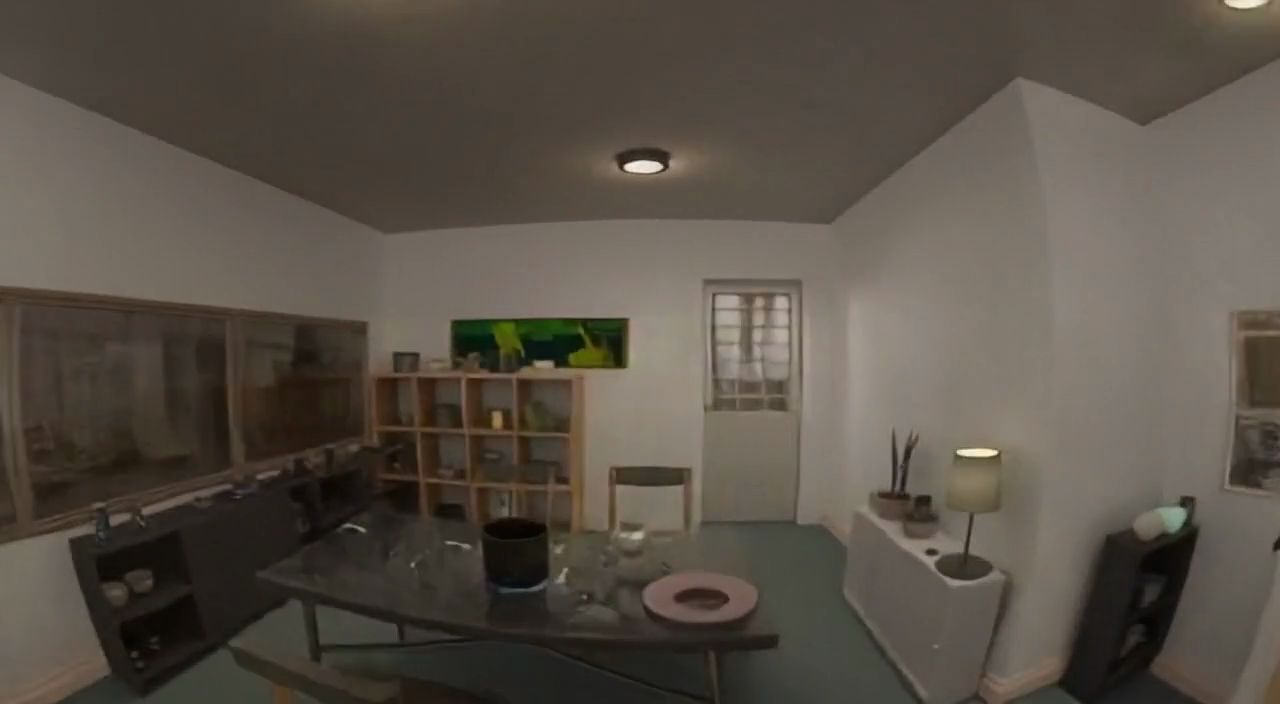} &
    \includegraphics[width=\cellW,height=\cellH,keepaspectratio]{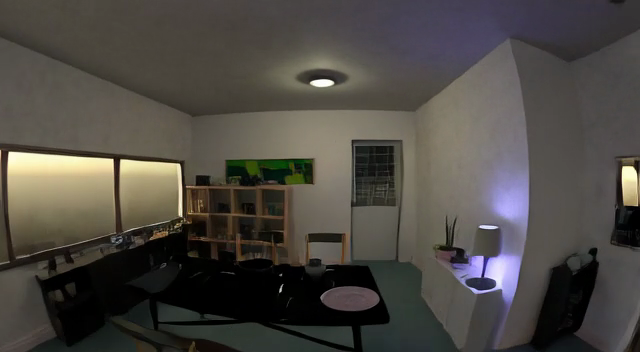} &
    \includegraphics[width=\cellW,height=\cellH,keepaspectratio]{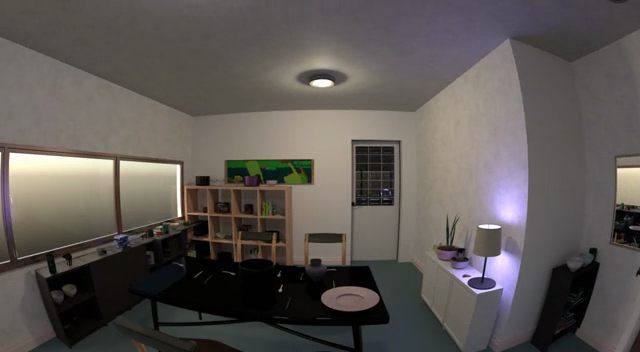}  \\[-2pt]
    \includegraphics[width=\cellW,height=\cellH,keepaspectratio]{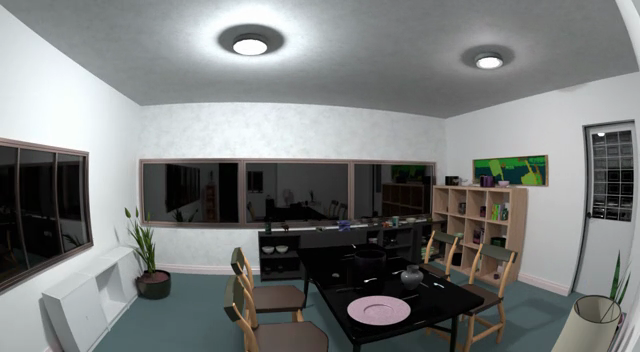} &
    \includegraphics[width=\cellW,height=\cellH,keepaspectratio]{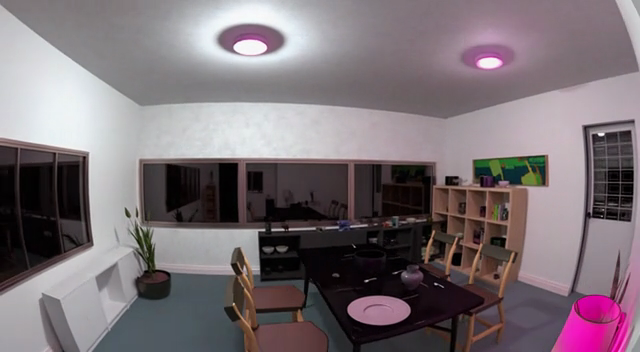} &
    \includegraphics[width=\cellW,height=\cellH,keepaspectratio]{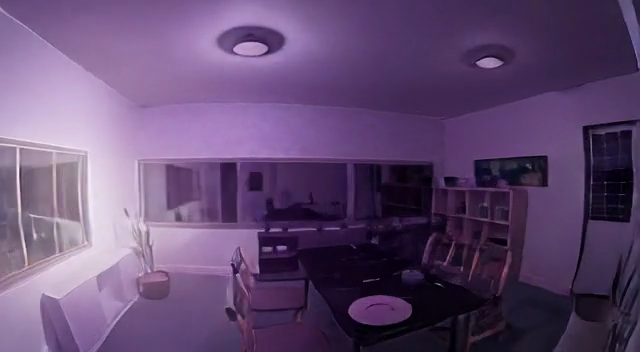} &
    \includegraphics[width=\cellW,height=\cellH,keepaspectratio]{figures/figs_resource/failed.pdf} &
    \includegraphics[width=\cellW,height=\cellH,keepaspectratio]{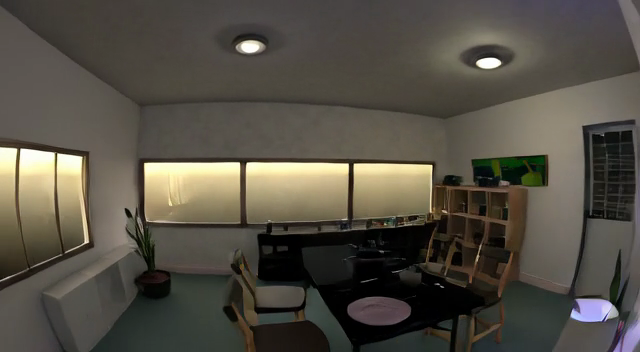} &
    \includegraphics[width=\cellW,height=\cellH,keepaspectratio]{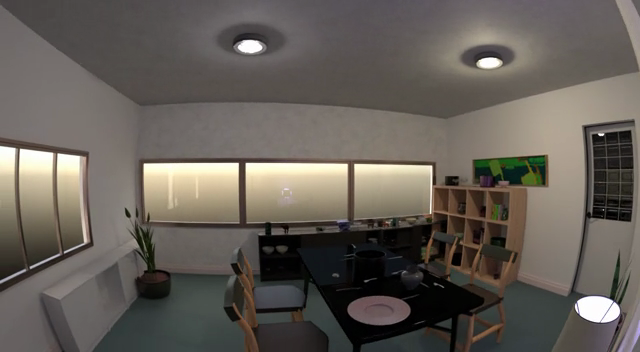} \\
    
        \parbox[c][\baselineskip][c]{\cellW}{\centering\small Input} &
    \parbox[c][\baselineskip][c]{\cellW}{\centering\small LightLab} &
    \parbox[c][\baselineskip][c]{\cellW}{\centering\small Light-A-Video} &
    \parbox[c][\baselineskip][c]{\cellW}{\centering\small DiffusionRenderer} &
    \parbox[c][\baselineskip][c]{\cellW}{\centering\small Ours} &
    \parbox[c][\baselineskip][c]{\cellW}{\centering\small Ground Truth}
  \end{tabular}}
  \caption{\textbf{Results on a held-out synthetic scene.} Here we show qualitative comparisons between our method and existing approaches—including LightLab, Light-A-Video, and DiffusionRenderer. These are conducted on a test scene from our Infinigen dataset. LightLab exhibits reduced flickering due to the relatively stable exposure in synthetic scenes; however, it fails to turn on the exterior lighting and does not reproduce accurate color tones. Light-A-Video successfully turns on the exterior light but struggles to recover the correct indoor lighting. DiffusionRenderer, while achieving better temporal consistency than in real-world scenarios, also fails to reproduce accurate color. In contrast, our method not only turns on both the exterior light and the indoor lamp but also achieves accurate color rendering for both, demonstrating superior visual fidelity and lighting control. }
  \label{fig:infinigen_show}
\end{figure*}

\appendix
\clearpage 

\section*{\huge Appendix}

\renewcommand{\contentsname}{Overview of the Supplementary Material} %

\section{Implementation Details}
\subsection{Relighting Diffusion Model}
\label{sec:Train_relighting}
\paragraph{Training.} As described in the main paper, we fine-tune the WAN~2.2 TI2V-5B model using the AdamW optimizer with a learning rate of $10^{-6}$. The additional MLP modules introduced for our method are trained jointly with the diffusion model, using a learning rate of $10^{-5}$. We use Fully Sharded Data Parallel (FSDP) to distribute the model across multiple hosts for efficient large-scale training. 

\paragraph{Inference.} At inference time, we generate an initial random noise tensor matching the spatial and temporal dimensions of the input video latent, and apply 50 denoising steps using the same flow-matching scheduler employed during training. The full inference process requires approximately ${\sim}100$ seconds for generating an 81-frame video at a resolution of $640\times352$ on 8~H100 GPUs.

\subsection{Reconstruction and 3D Distillation}
\label{sec:zipnerf_train_bake}

As mentioned in the main paper, our 3D reconstruction is done using Zip-NeRF~\cite{barron2023zipnerf}. Once the Zip-NeRF model is trained, we render an elliptical camera path and feed it through our relighting model.

\section{Real-world 3D Environment Relighting}
\label{sec:realworld}

We conduct experiments on four scenes from the Eyeful Tower dataset~\cite{xu2023vr}: \texttt{office1a}, \texttt{office\_view2}, \texttt{kitchen}, and \texttt{seating\_area}. For each scene, we first mark the most prominent light sources in 3D. This is done by selecting image pixels in 2D and back-projecting them to 3D, and modifying the color of the 3D points corresponding to the light sources to $1$ and other points to $0$. The resulting 3D light masks are then rendered along the same elliptical camera path to obtain per-frame 2D light masks that are aligned with the input video frames.

\subsection{Details of User Study}
\begin{figure}
    \centering
    \includegraphics[width=\linewidth]{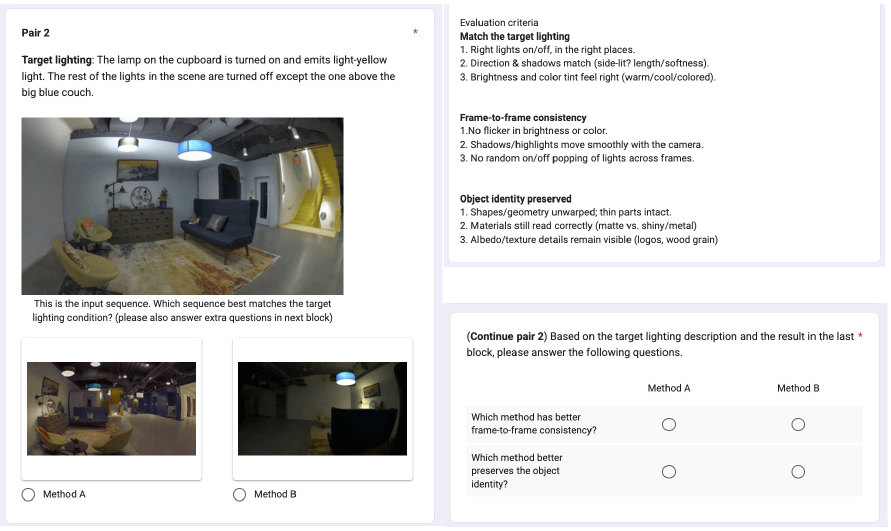}
    \caption{\textbf{Illustration of our user study.} The user is asked to provide feedback of the questions based on the explained criteria.}
    \label{fig:Userstudy}
\end{figure}
In our user study, participants were presented with two videos generated by one of four sources: LightLab~\cite{lightlab}, DiffusionRenderer~\cite{DiffusionRenderer2025} (Cosmos), our method, and the input video. We include the input video as a baseline since, although it does not respect the target lighting condition, it is entirely free of relighting artifacts and therefore often exhibits very high quality. The two videos presented to each user share the same camera path. We provide users with a text description of the target lighting condition to reduce ambiguity, specifying the on/off state, and color of each indoor light source, as well as the exterior lighting condition.

Figure~\ref{fig:Userstudy} shows the interface used in our user study. Each of 12 users is asked to make 18 comparisons, making up a total of 216 responses. Users were asked to compare the two videos and determine which better satisfies three criteria: frame-to-frame consistency, object identity preservation, and lighting similarity to the target condition. Each comparison was presented in randomized order to mitigate position bias, and responses were aggregated across all raters. The lower win rate against Input (68.5\%) is expected, as the original video naturally preserves temporal consistency and object identity perfectly—yet users still preferred our method, indicating they value the improved lighting fidelity our approach provides.

\subsection{Qualitative Comparison}
\label{sec:realworld_qual}
We present some of the qualitative results in Figures~\ref{fig:office1a}, and~\ref{fig:seating} (Light-A-Video, DiffusionRenderer (Cosmos), LightLab and Ours). Please refer to the \href{gr3en-relight.github.io/index.html}{webpage} for more video examples.

\subsection{Fine-grained Lighting Control}
\label{sec:fine_control}
We show the fine-grained individual light source control in Figures~\ref{fig:extra_color} and~\ref{fig:controllable_lighting}. 
Please refer to the \href{gr3en-relight.github.io/fine_control.html}{webpage} to see more examples.

\section{Data Generation Pipeline Visualization}
\label{sec:data_gen}

Figure \ref{fig:datagen1} illustrates the training data used in our approach. Built upon the Infinigen framework \cite{raistrick2024infinigen,raistrick2023infinite}, our training set comprises diverse indoor scenes featuring various floor plans and materials. Our OLAT (One-Light-at-a-Time) data generation pipeline simulates a wide range of lighting conditions within each scene. Notably, both interior lighting (including color, intensity, and on/off states) and exterior lighting are systematically varied across different instances.

\paragraph{\textbf{Images are shown on the following pages.}}

\clearpage

\setlength{\cellW}{0.16\textwidth}

\setlength{\cellH}{0.5625\cellW}

\setlength{\rowLabelW}{0.03\textwidth}

\begin{figure*}[p]
  \centering
  \setlength{\tabcolsep}{2pt}          %
  \renewcommand{\arraystretch}{0.8}    %
    \begin{tabular}{@{}c*{5}{c}@{}}

    \includegraphics[width=\cellW,height=\cellH,keepaspectratio]{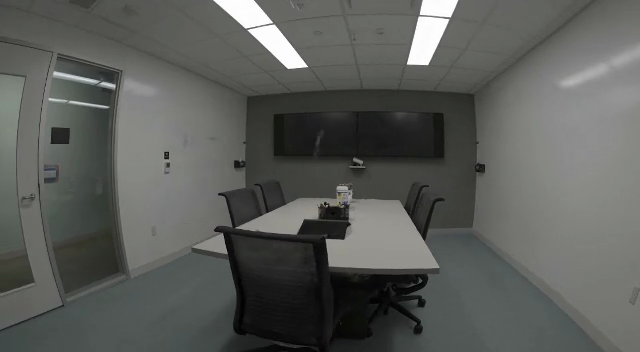} &
    \includegraphics[width=\cellW,height=\cellH,keepaspectratio]{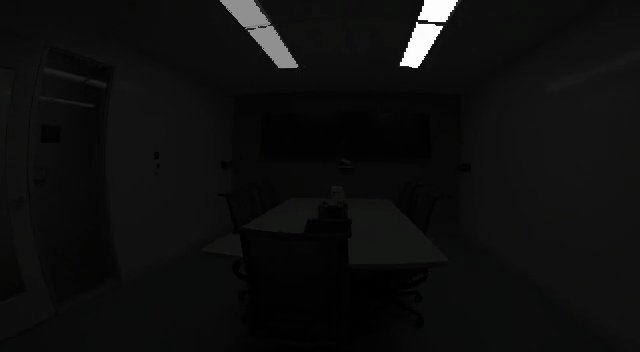} &
    \includegraphics[width=\cellW,height=\cellH,keepaspectratio]{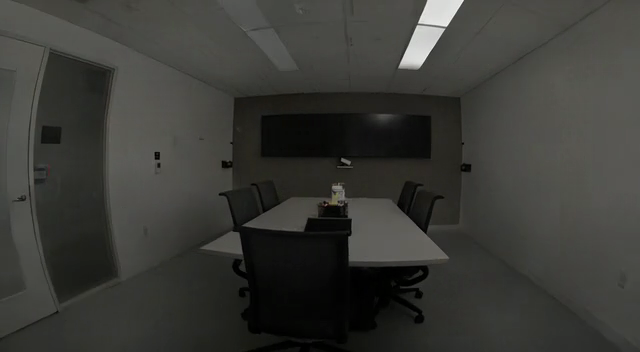} &
    \includegraphics[width=\cellW,height=\cellH,keepaspectratio]{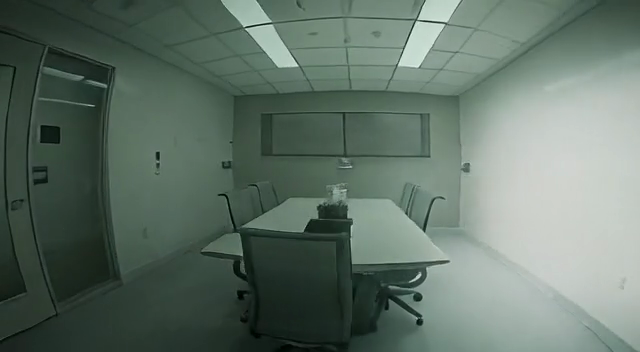} &
    \includegraphics[width=\cellW,height=\cellH,keepaspectratio]{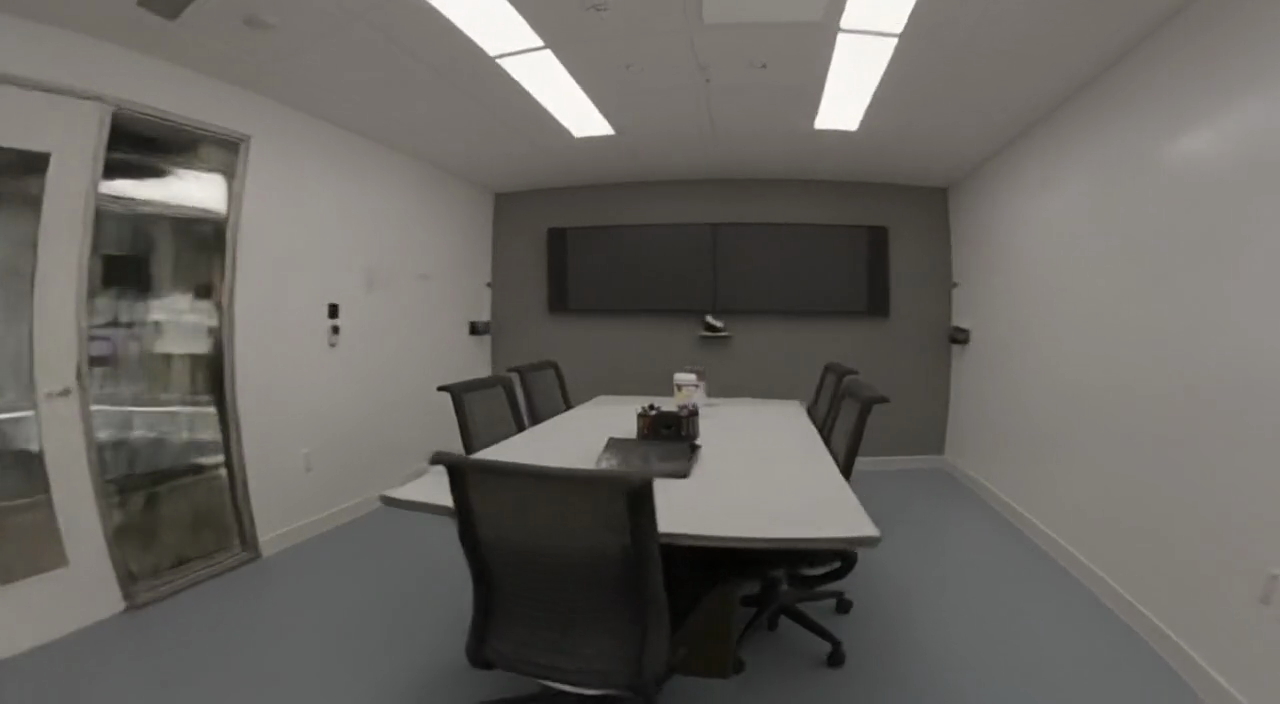} &
    \includegraphics[width=\cellW,height=\cellH,keepaspectratio]{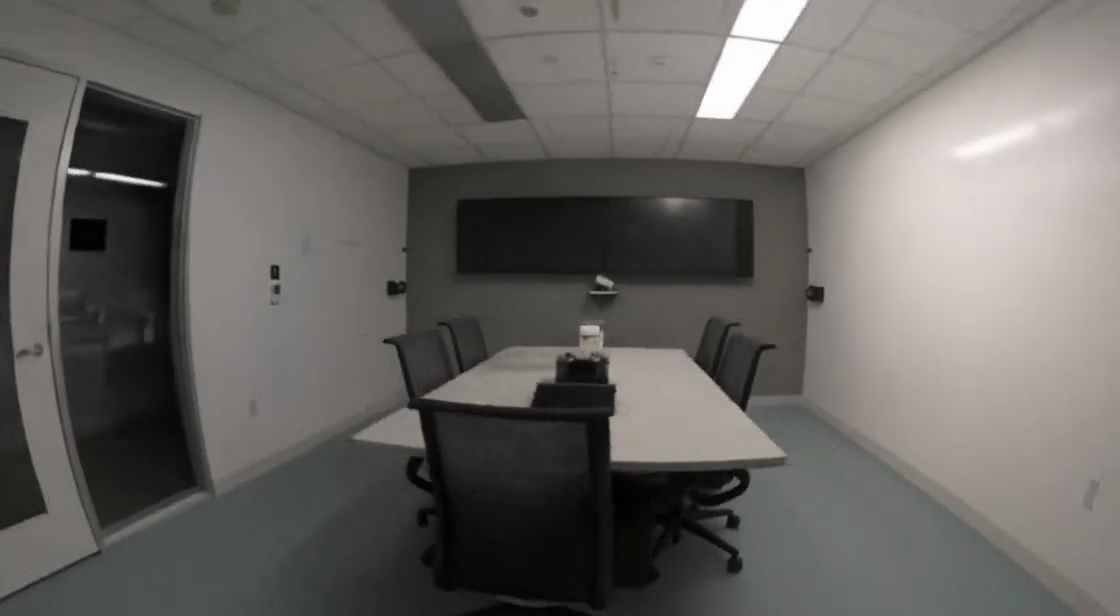}  \\[-2pt]
    \includegraphics[width=\cellW,height=\cellH,keepaspectratio]{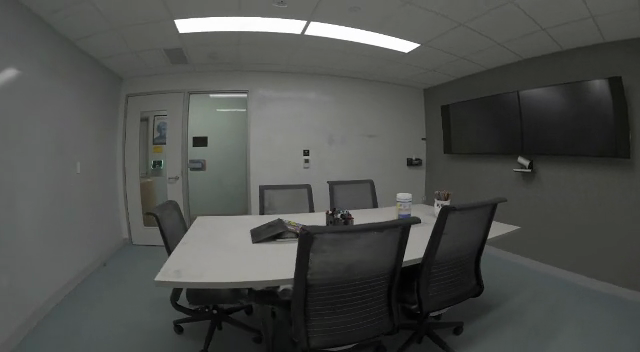} &
    \includegraphics[width=\cellW,height=\cellH,keepaspectratio]{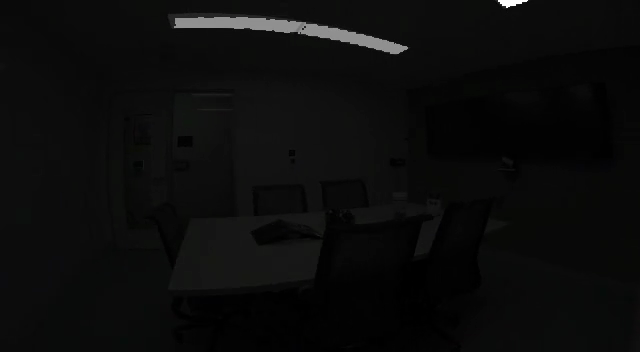} &
    \includegraphics[width=\cellW,height=\cellH,keepaspectratio]{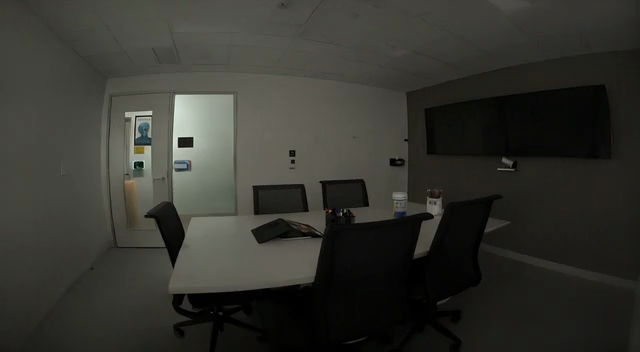} &
    \includegraphics[width=\cellW,height=\cellH,keepaspectratio]{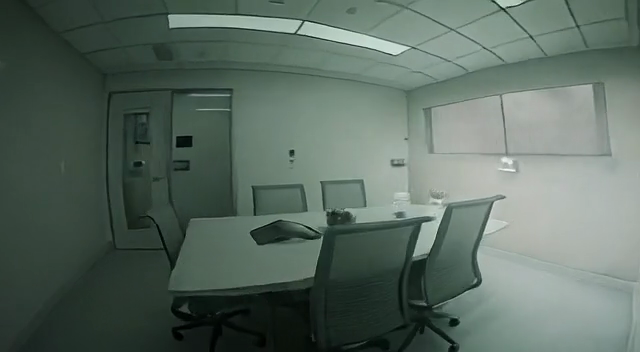} &
    \includegraphics[width=\cellW,height=\cellH,keepaspectratio]{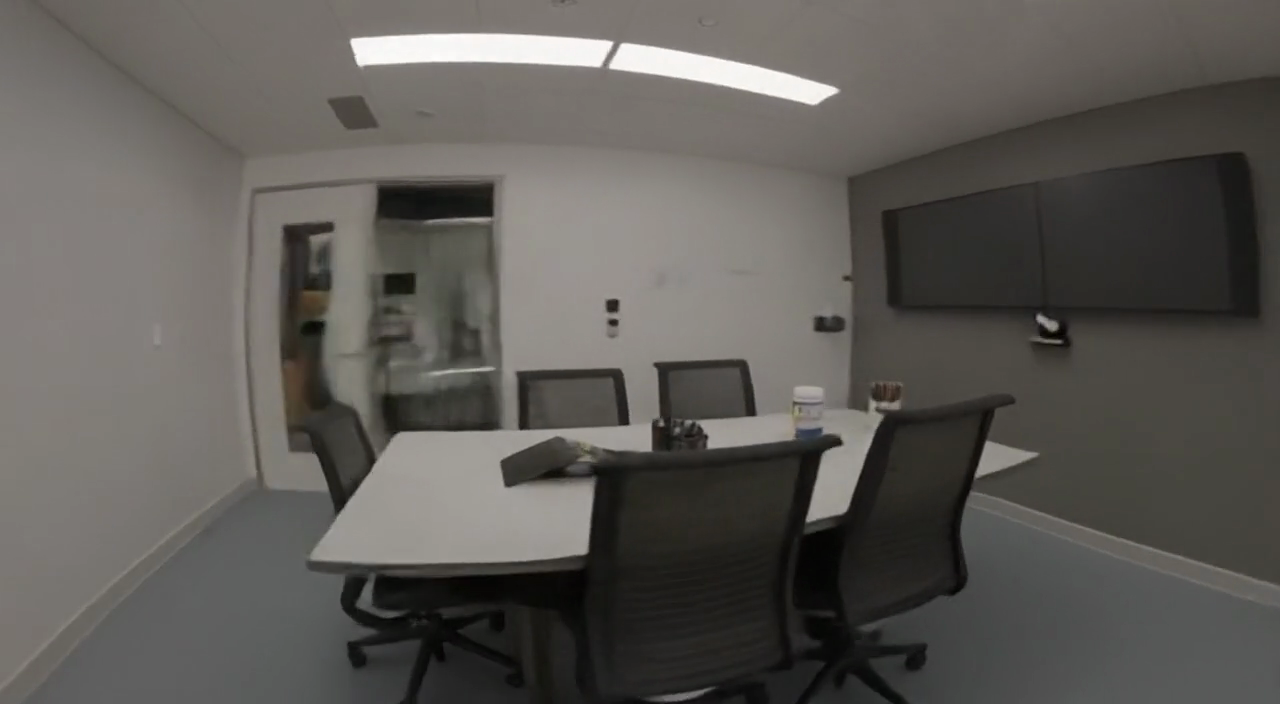} &
    \includegraphics[width=\cellW,height=\cellH,keepaspectratio]{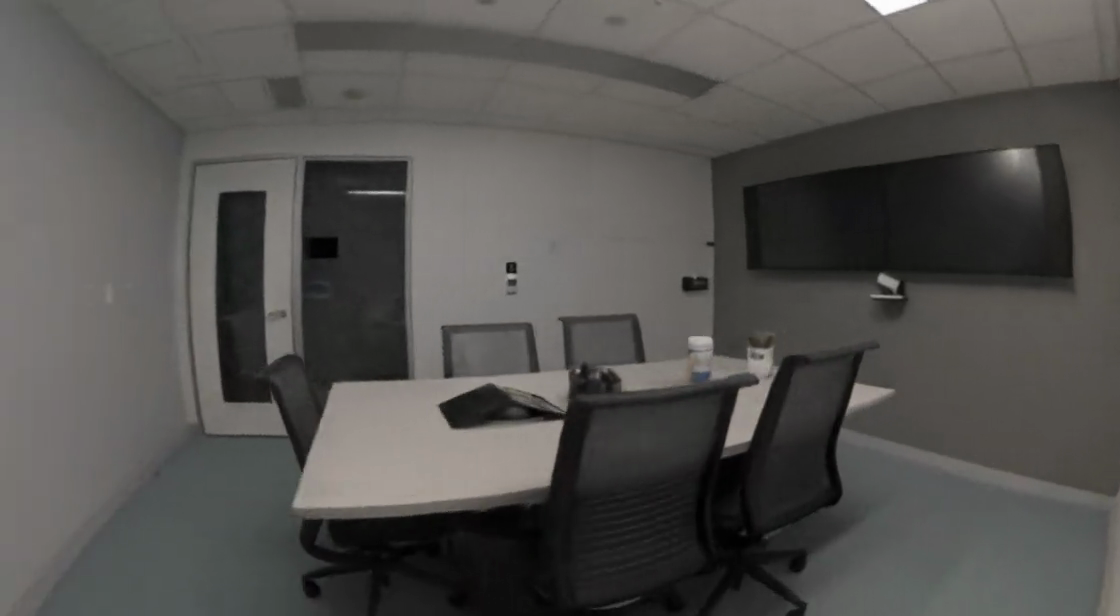} \\[-2pt]
    \includegraphics[width=\cellW,height=\cellH,keepaspectratio]{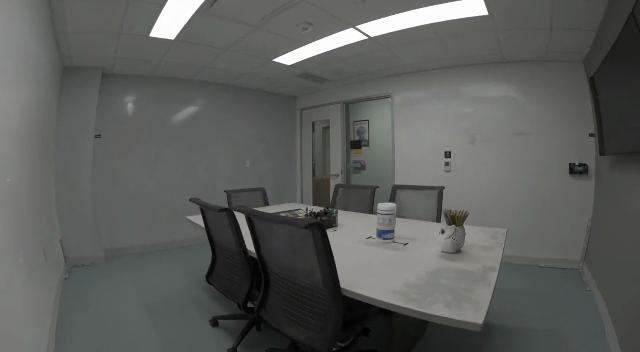} &
    \includegraphics[width=\cellW,height=\cellH,keepaspectratio]{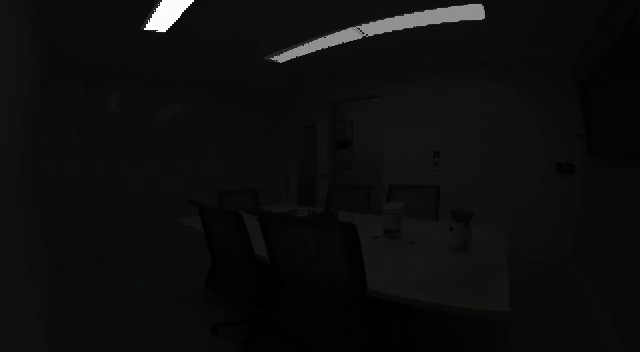} &
    \includegraphics[width=\cellW,height=\cellH,keepaspectratio]{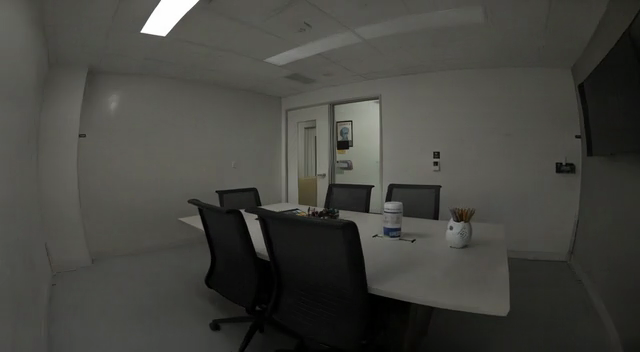} &
    \includegraphics[width=\cellW,height=\cellH,keepaspectratio]{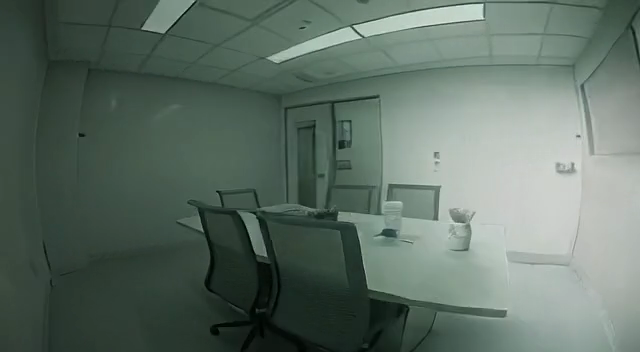} &
    \includegraphics[width=\cellW,height=\cellH,keepaspectratio]{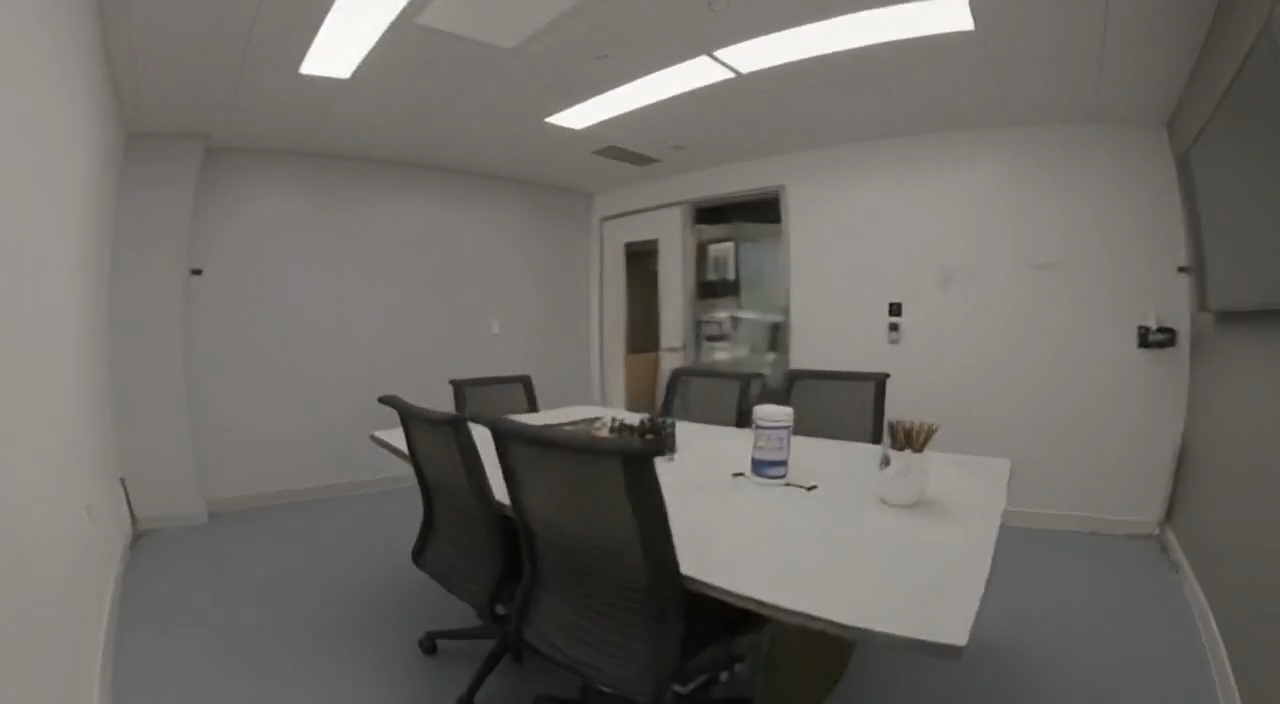} &
    \includegraphics[width=\cellW,height=\cellH,keepaspectratio]{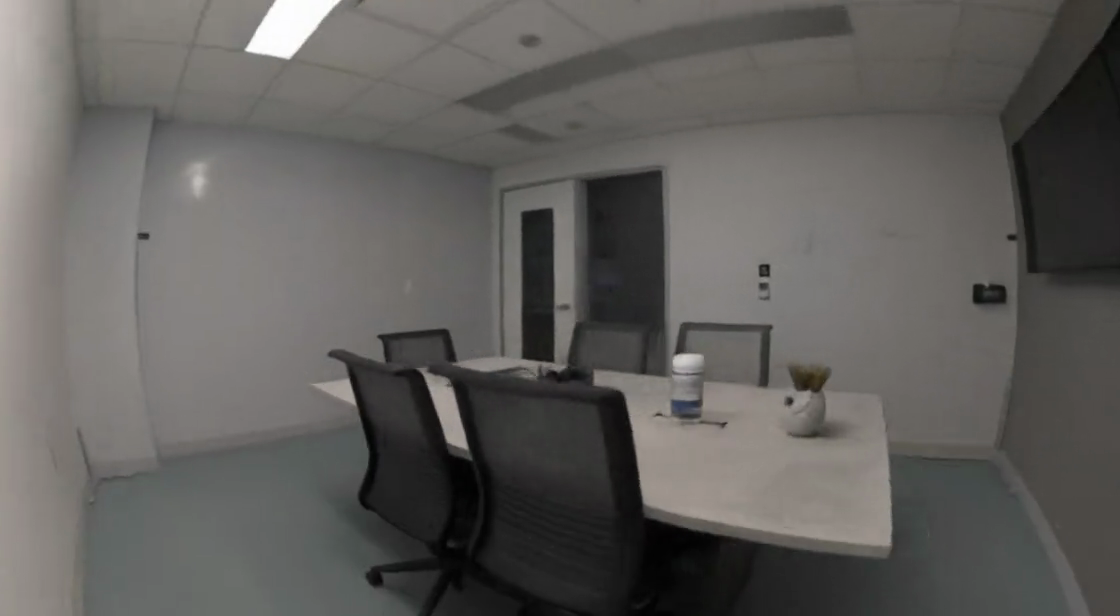} \\[-2pt]
    \includegraphics[width=\cellW,height=\cellH,keepaspectratio]{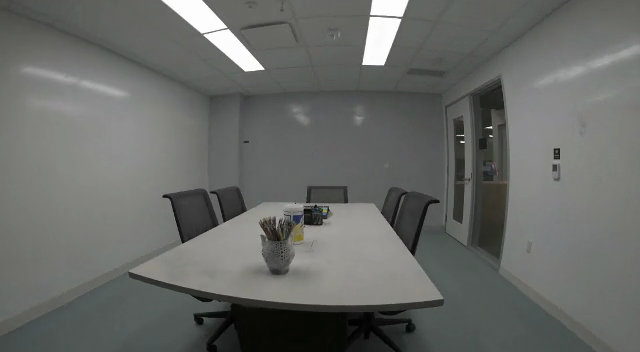} &
    \includegraphics[width=\cellW,height=\cellH,keepaspectratio]{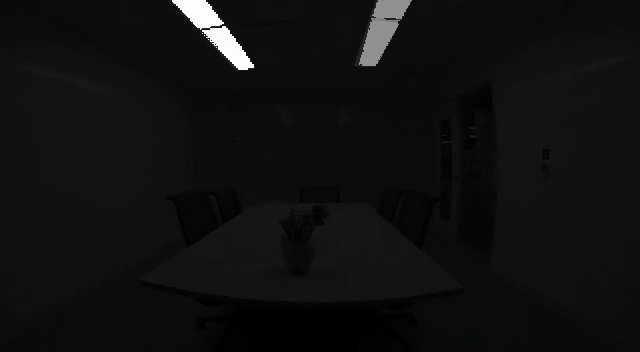} &
    \includegraphics[width=\cellW,height=\cellH,keepaspectratio]{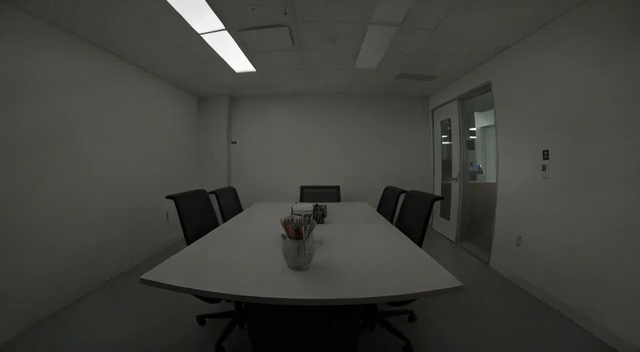} &
    \includegraphics[width=\cellW,height=\cellH,keepaspectratio]{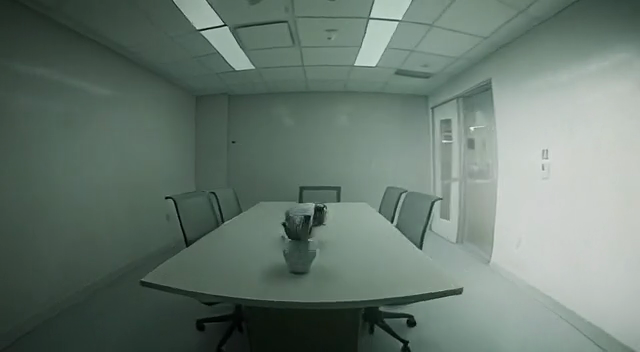} &
    \includegraphics[width=\cellW,height=\cellH,keepaspectratio]{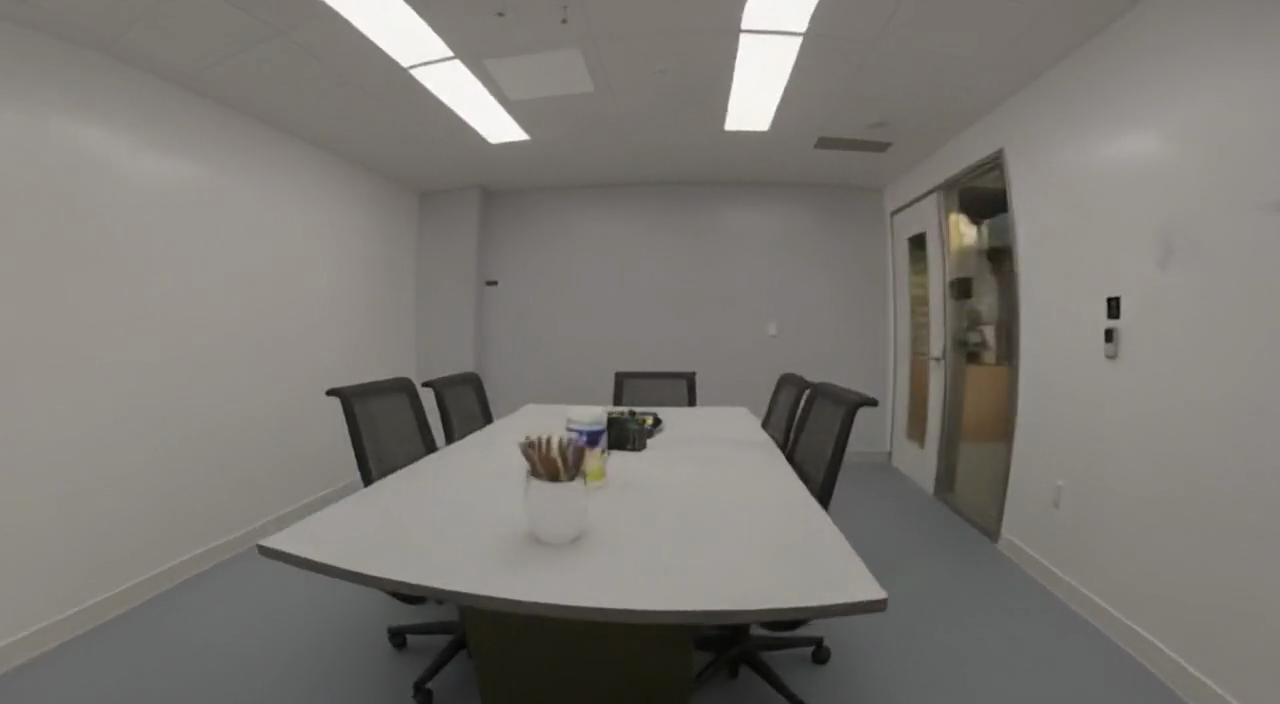} &
    \includegraphics[width=\cellW,height=\cellH,keepaspectratio]{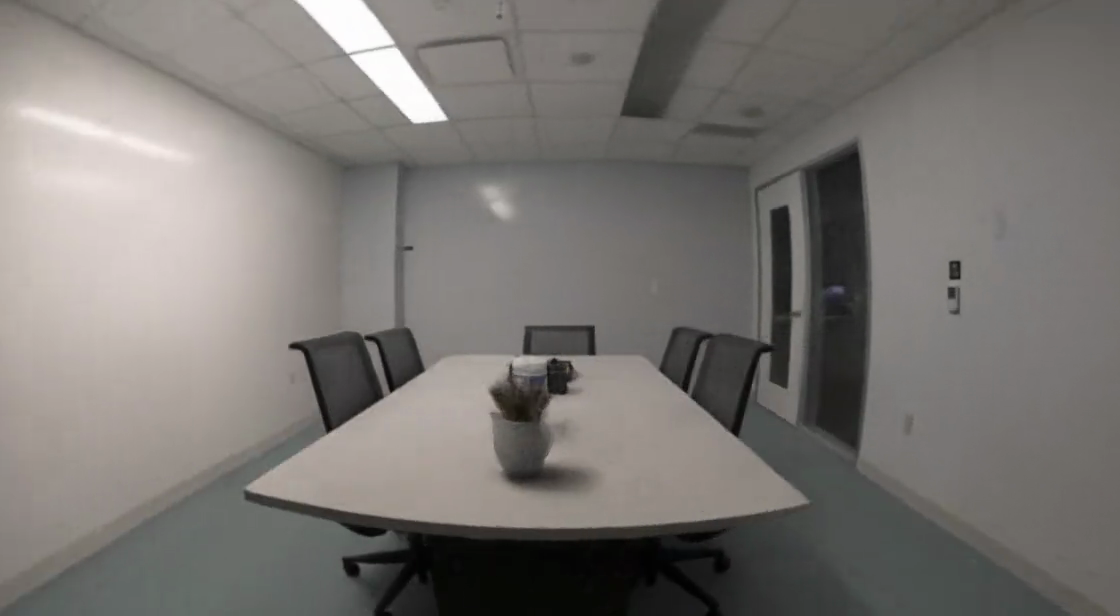} 
    \\[-2pt]
    \includegraphics[width=\cellW,height=\cellH,keepaspectratio]{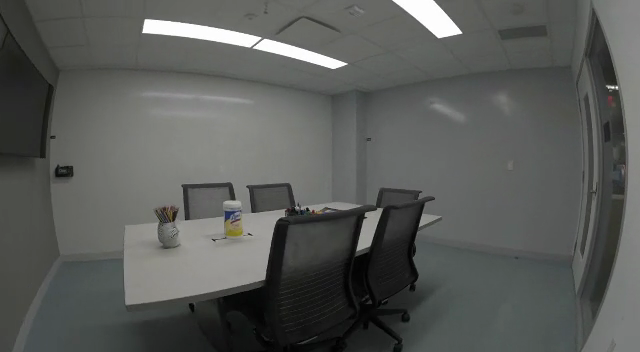} &
    \includegraphics[width=\cellW,height=\cellH,keepaspectratio]{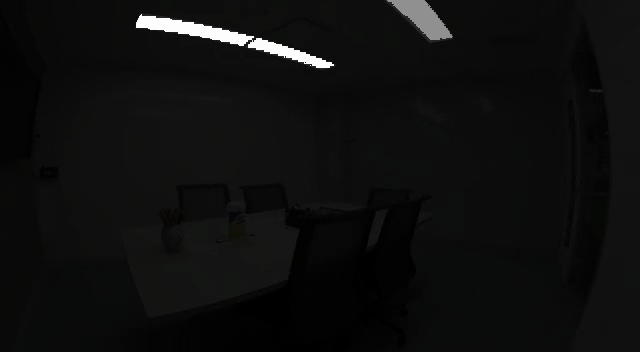} &
    \includegraphics[width=\cellW,height=\cellH,keepaspectratio]{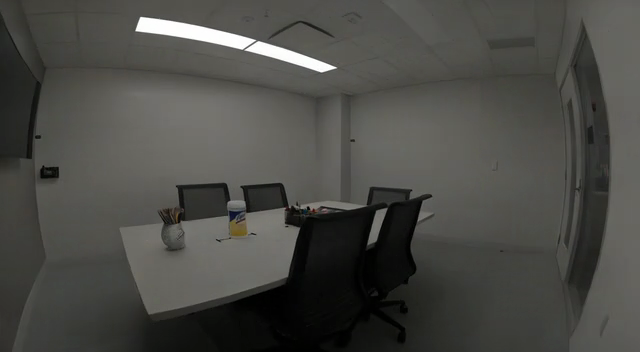} &
    \includegraphics[width=\cellW,height=\cellH,keepaspectratio]{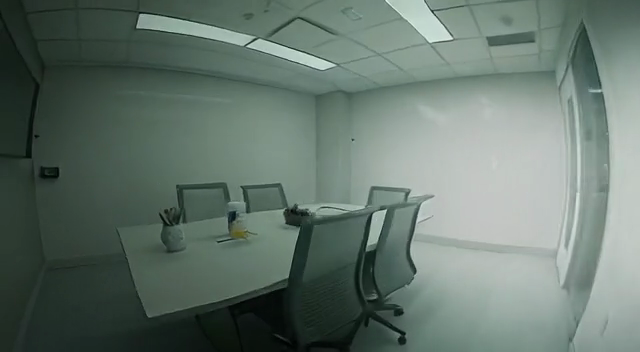} &
    \includegraphics[width=\cellW,height=\cellH,keepaspectratio]{figures/figs_resource/failed.pdf} &
    \includegraphics[width=\cellW,height=\cellH,keepaspectratio]{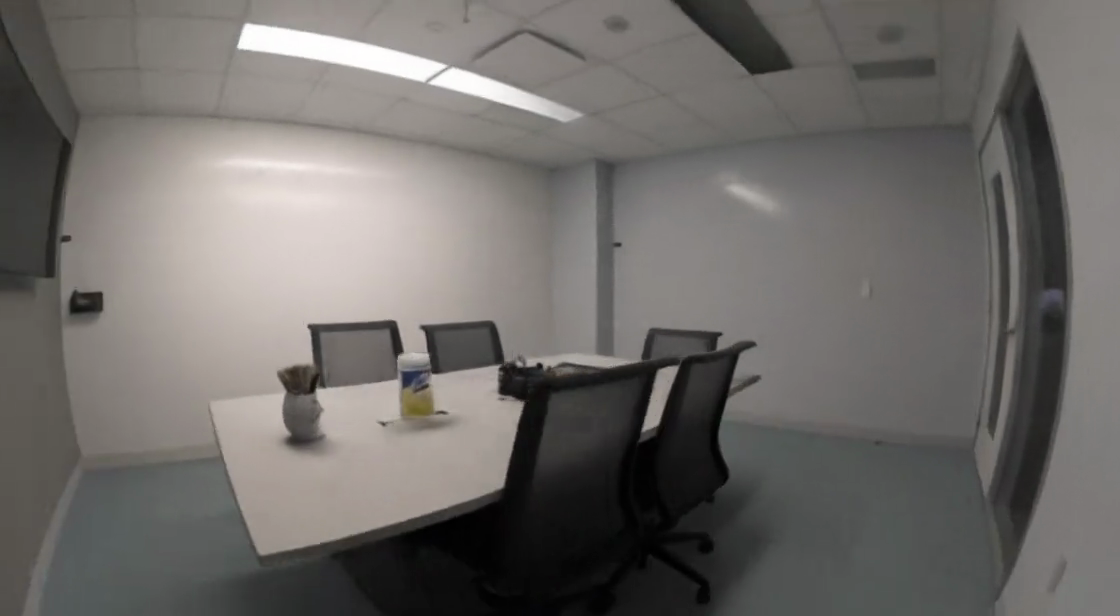} \\
    \parbox[c][\baselineskip][c]{\cellW}{\centering\small Input} &
    \parbox[c][\baselineskip][c]{\cellW}{\centering\small Condition} &
    \parbox[c][\baselineskip][c]{\cellW}{\centering\small LightLab} &
    \parbox[c][\baselineskip][c]{\cellW}{\centering\small Light-A-Video} &
    \parbox[c][\baselineskip][c]{\cellW}{\centering\small DiffusionRenderer} &
    \parbox[c][\baselineskip][c]{\cellW}{\centering\small Ours} 
  \end{tabular}
\caption{\textbf{Additional real-world results.} Qualitative comparisons of our method with LightLab, Light-A-Video, and DiffusionRenderer on the \texttt{office1a} scene from the Eyeful Tower dataset. LightLab independently relights each frame, resulting in noticeable temporal flickering artifacts. Light-A-Video offers limited lighting control and produces low-quality results. DiffusionRenderer achieves temporal consistency but suffers from blurriness and low visual fidelity. In contrast, our method realistically relights the scene based on the target lighting conditions. Notably, it accurately generates cast shadows for the object on the table and effectively removes undesired reflectance caused by lighting.}
  \label{fig:office1a}
\end{figure*}

\setlength{\cellW}{0.16\textwidth}

\setlength{\cellH}{0.5625\cellW}

\setlength{\rowLabelW}{0.03\textwidth}

\begin{figure*}[p]
  \centering
  \setlength{\tabcolsep}{2pt}          %
  \renewcommand{\arraystretch}{0.8}    %
  \begin{tabular}{@{}c*{5}{c}@{}}

    \includegraphics[width=\cellW,height=\cellH,keepaspectratio]{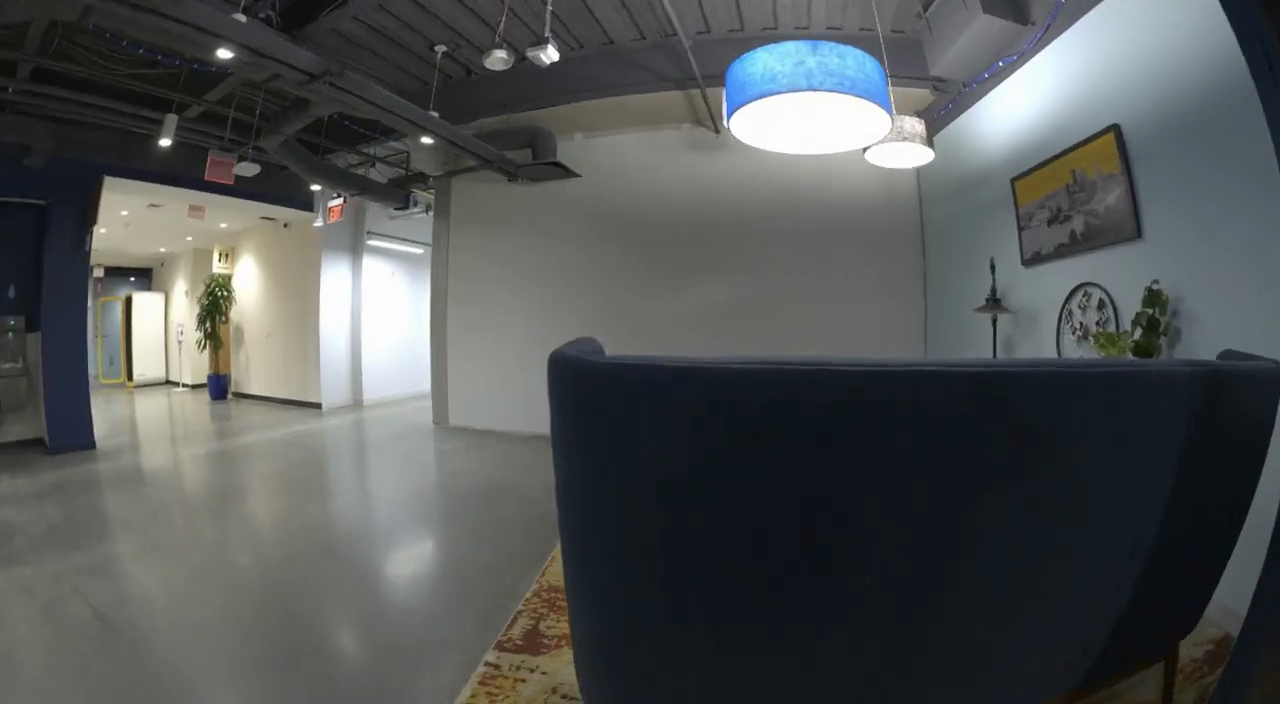} &
    \includegraphics[width=\cellW,height=\cellH,keepaspectratio]{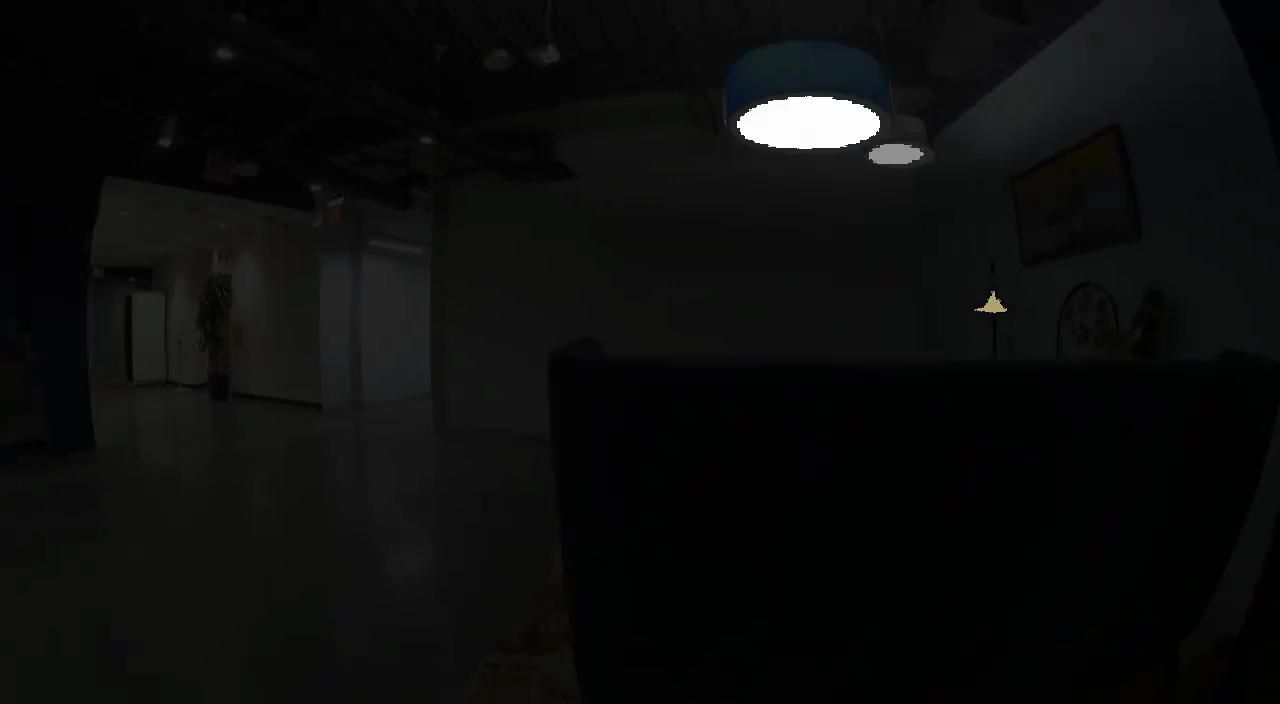} &
    \includegraphics[width=\cellW,height=\cellH,keepaspectratio]{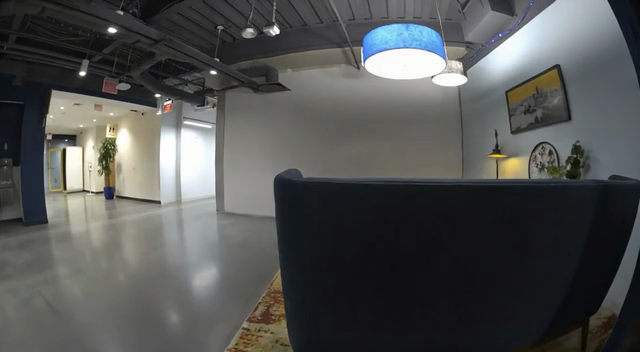} &
    \includegraphics[width=\cellW,height=\cellH,keepaspectratio]{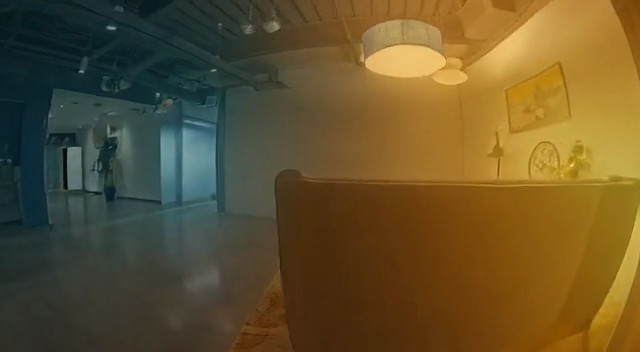} &
    \includegraphics[width=\cellW,height=\cellH,keepaspectratio]{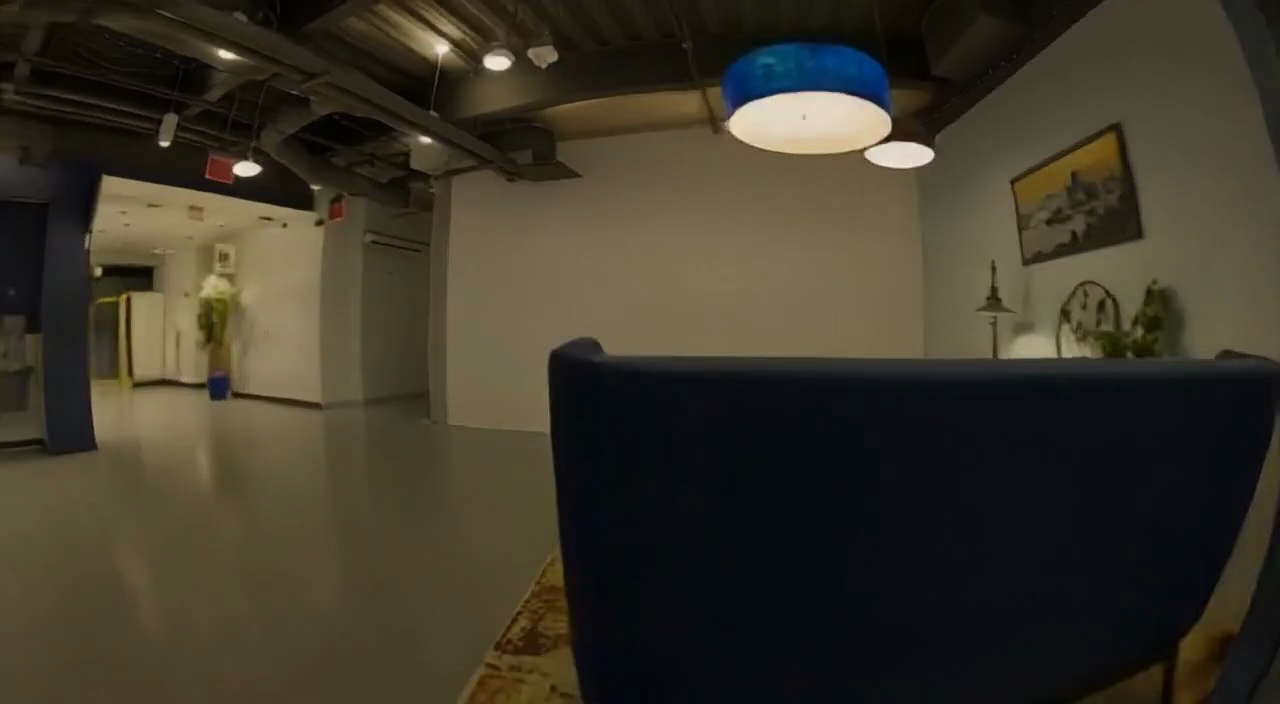} &
    \includegraphics[width=\cellW,height=\cellH,keepaspectratio]{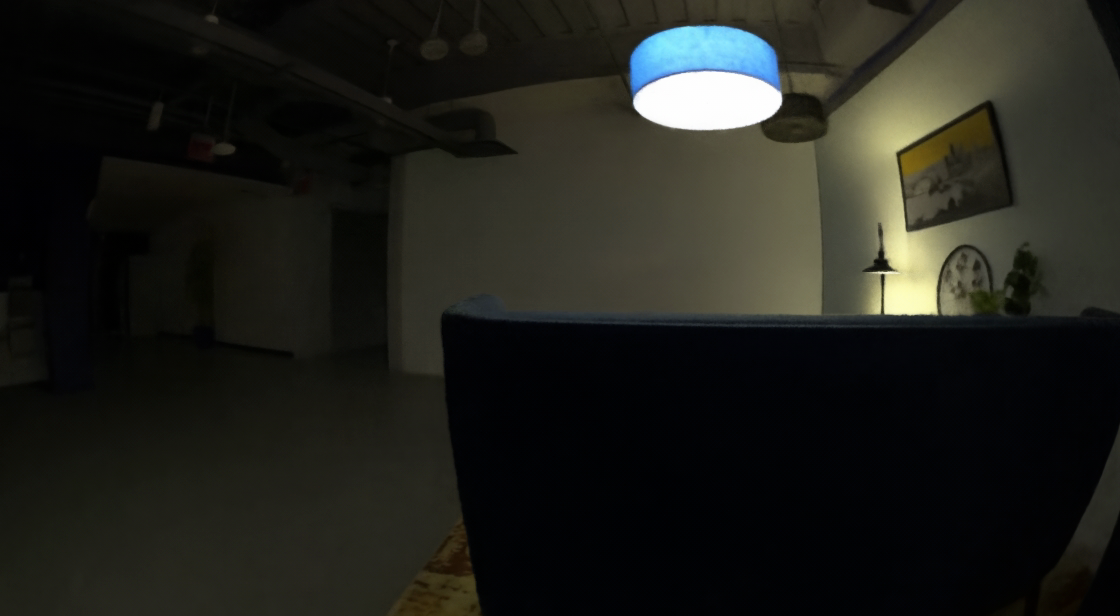}  \\[-2pt]
    \includegraphics[width=\cellW,height=\cellH,keepaspectratio]{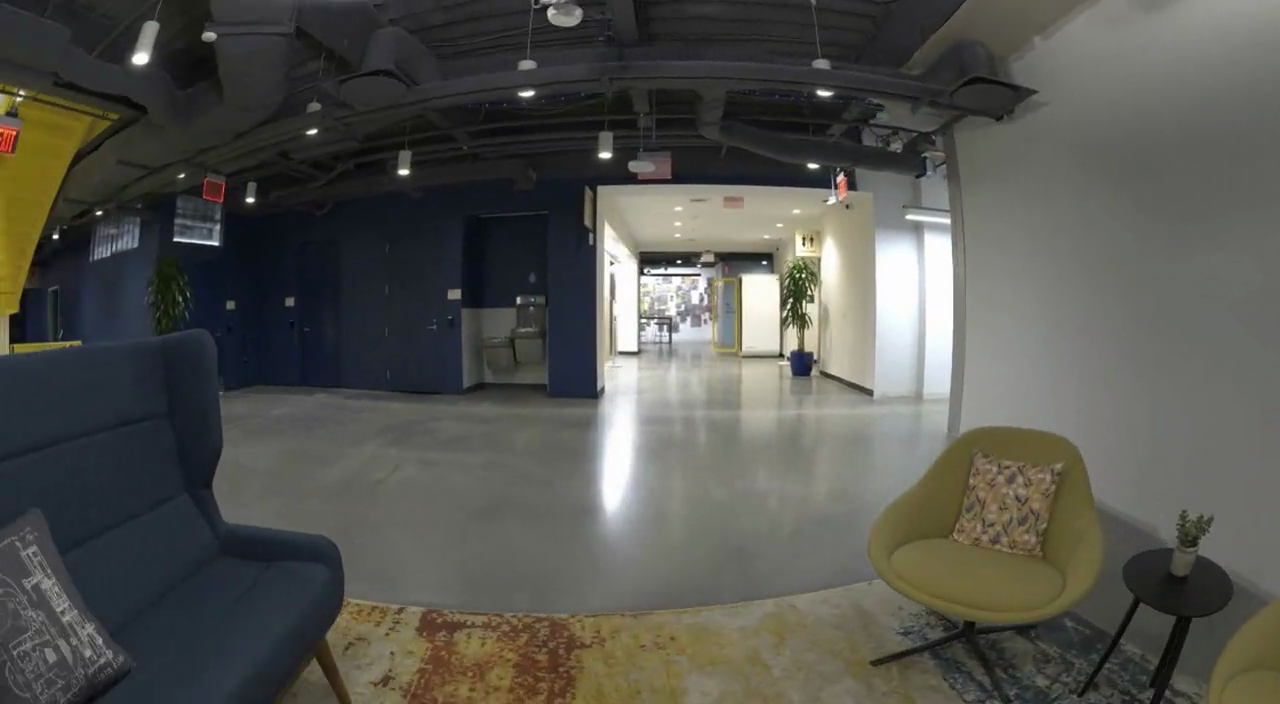} &
    \includegraphics[width=\cellW,height=\cellH,keepaspectratio]{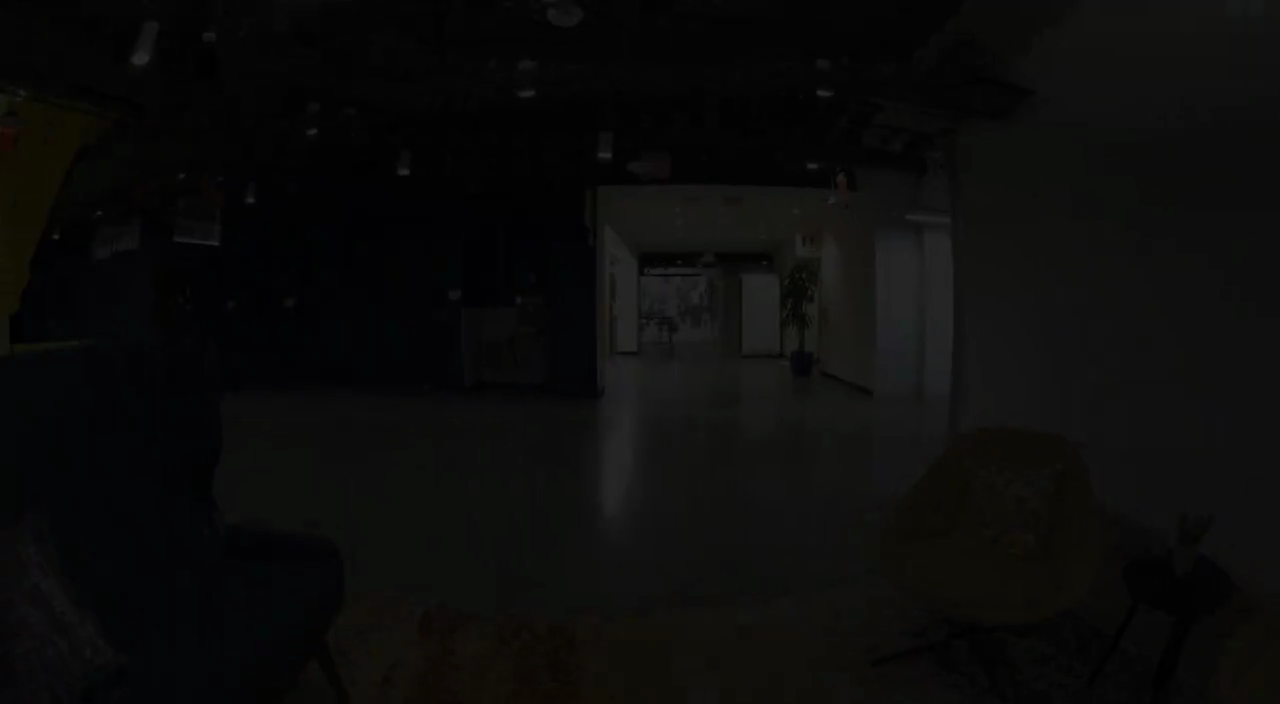} &
    \includegraphics[width=\cellW,height=\cellH,keepaspectratio]{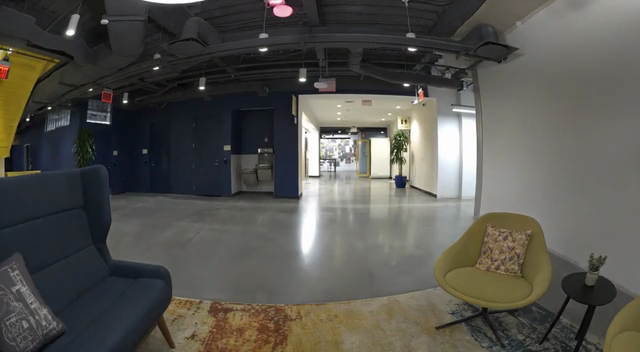} &
    \includegraphics[width=\cellW,height=\cellH,keepaspectratio]{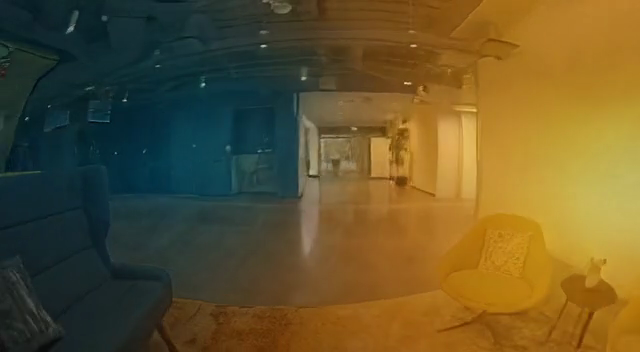} &
    \includegraphics[width=\cellW,height=\cellH,keepaspectratio]{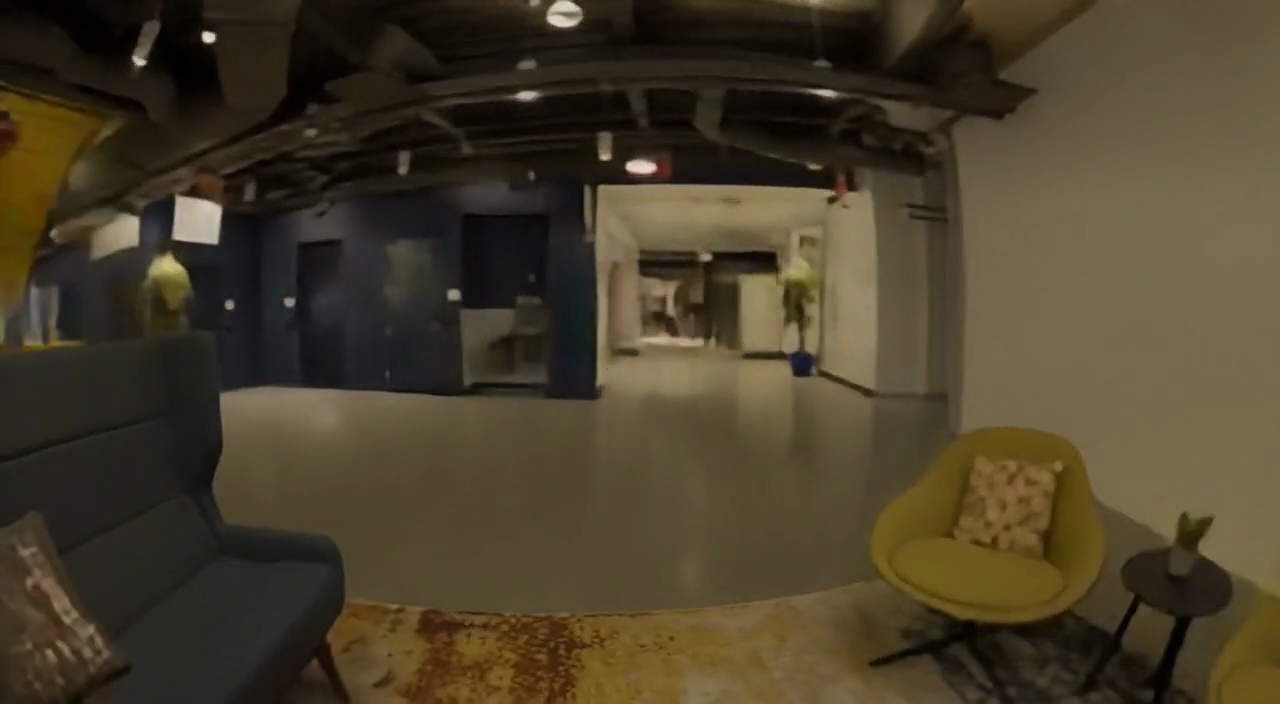} &
    \includegraphics[width=\cellW,height=\cellH,keepaspectratio]{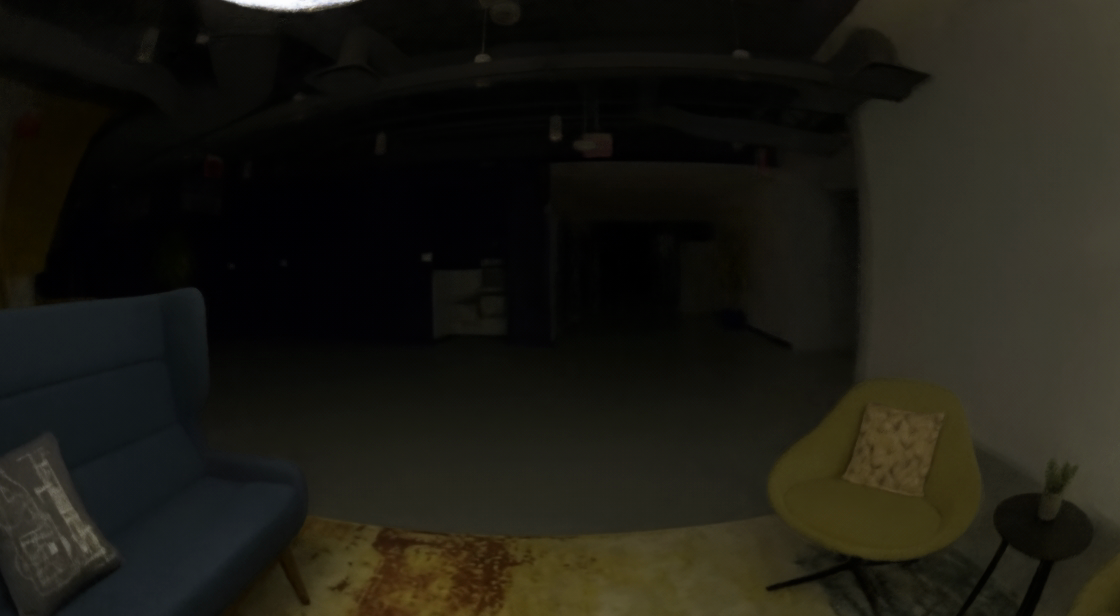} \\[-2pt]
    \includegraphics[width=\cellW,height=\cellH,keepaspectratio]{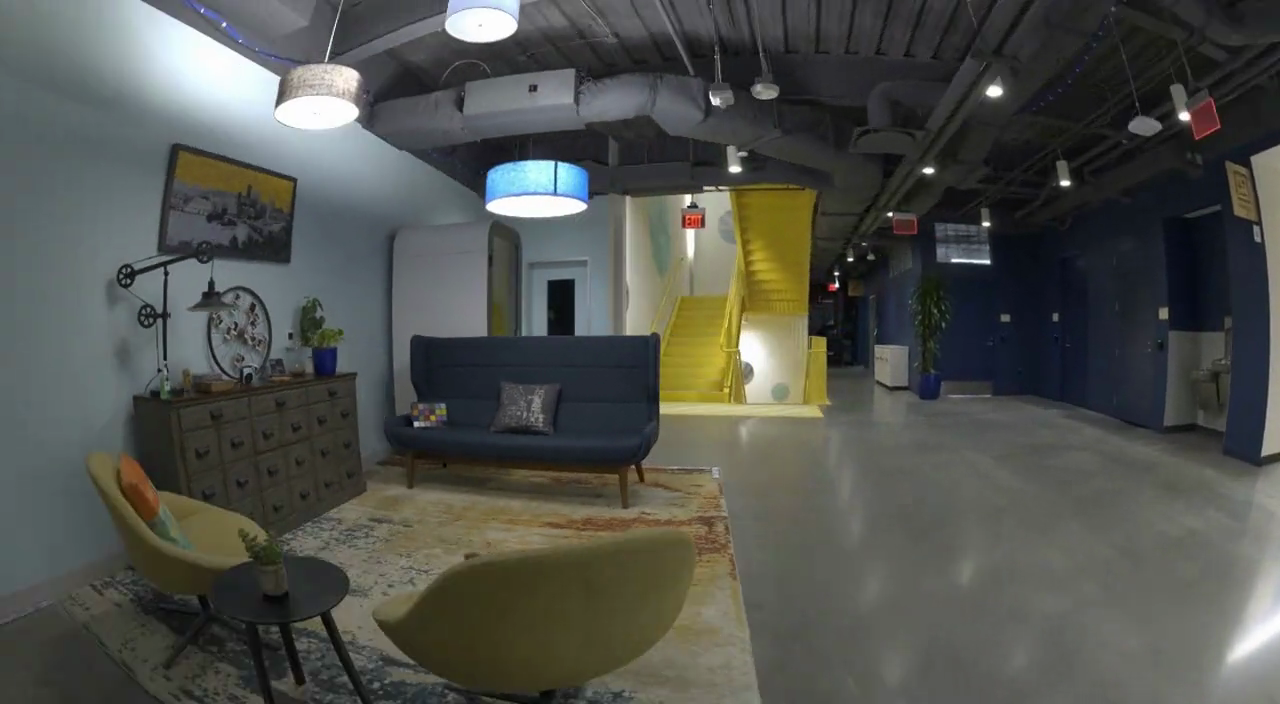} &
    \includegraphics[width=\cellW,height=\cellH,keepaspectratio]{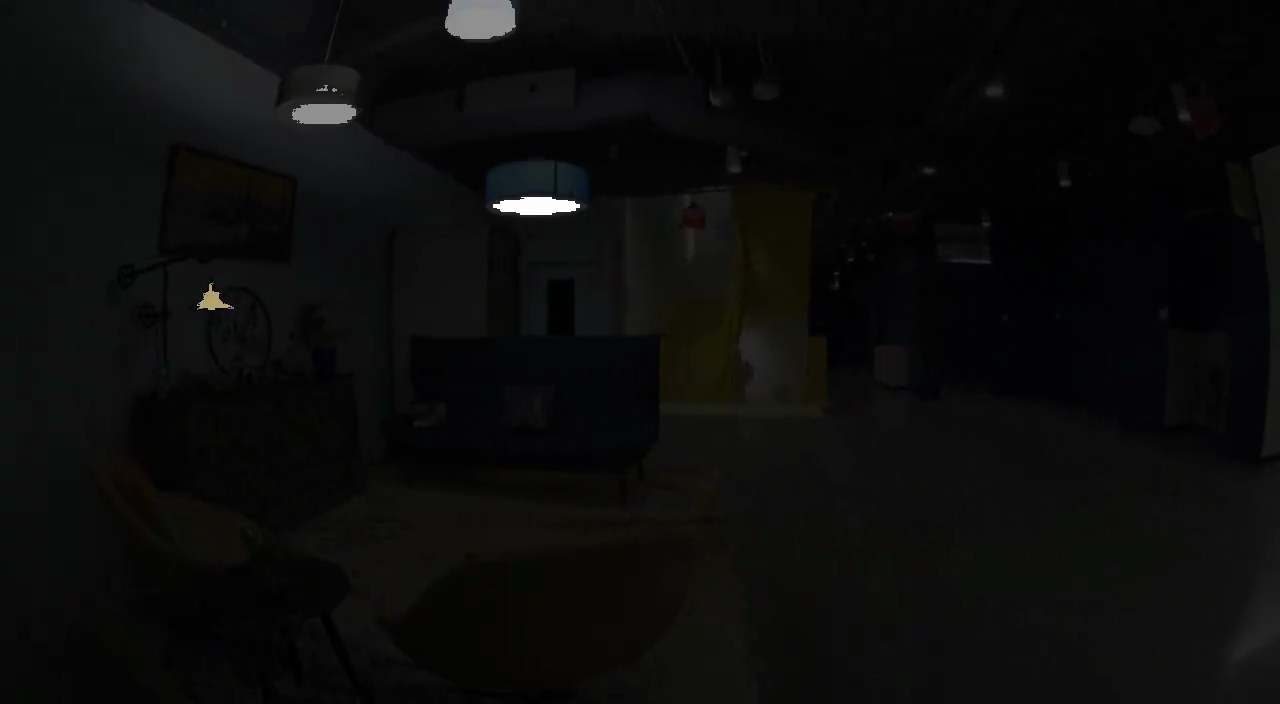} &
    \includegraphics[width=\cellW,height=\cellH,keepaspectratio]{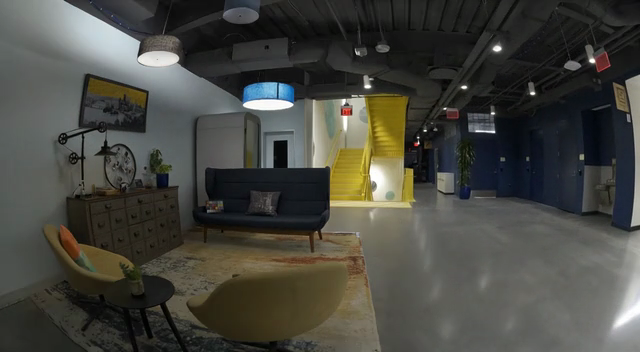} &
    \includegraphics[width=\cellW,height=\cellH,keepaspectratio]{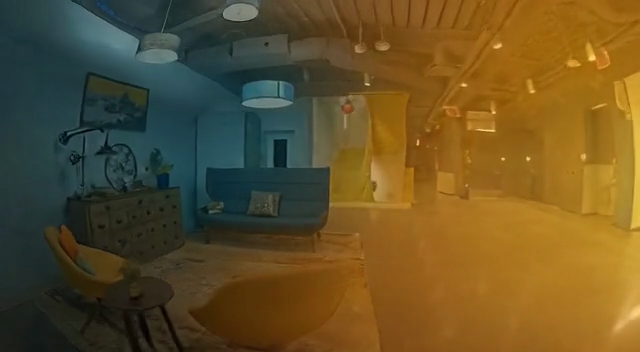} &
    \includegraphics[width=\cellW,height=\cellH,keepaspectratio]{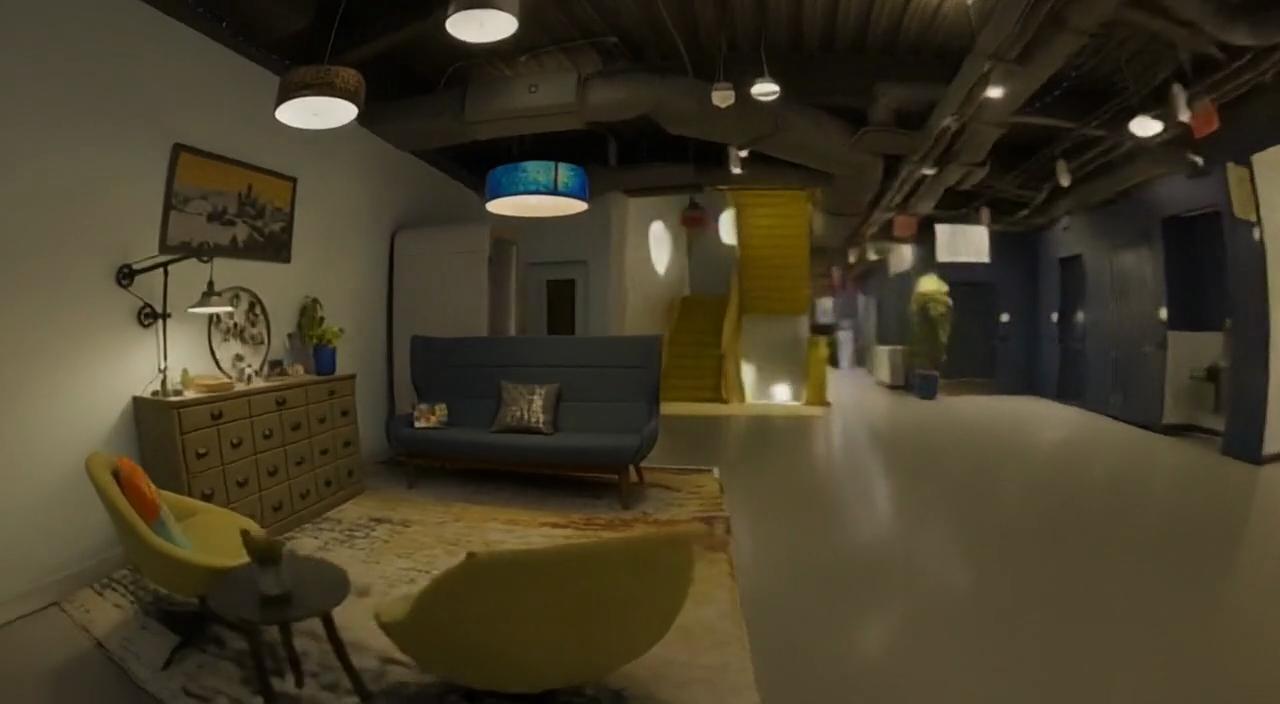} &
    \includegraphics[width=\cellW,height=\cellH,keepaspectratio]{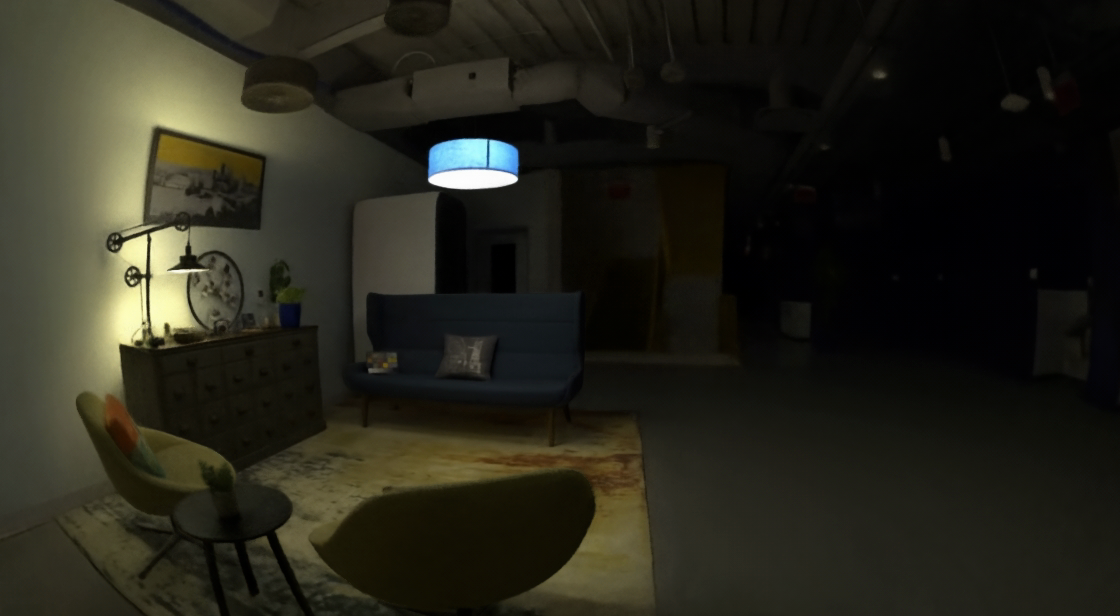} \\[-2pt]
    \includegraphics[width=\cellW,height=\cellH,keepaspectratio]{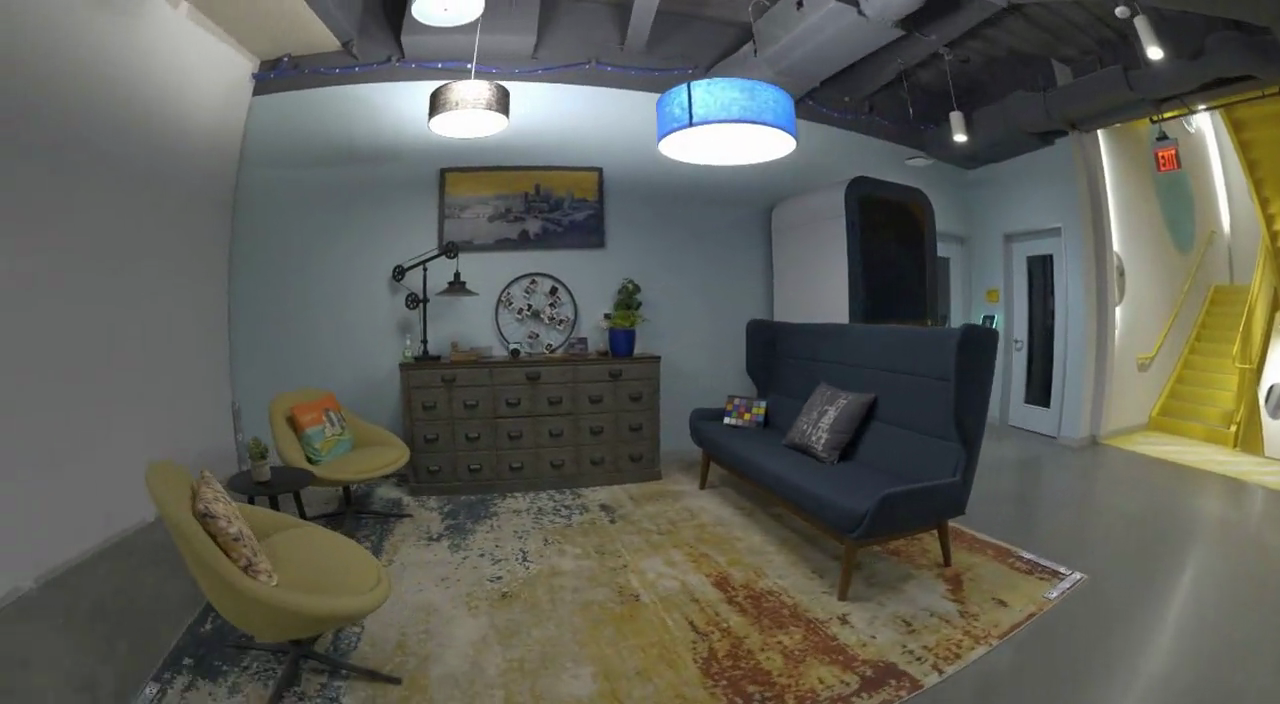} &
    \includegraphics[width=\cellW,height=\cellH,keepaspectratio]{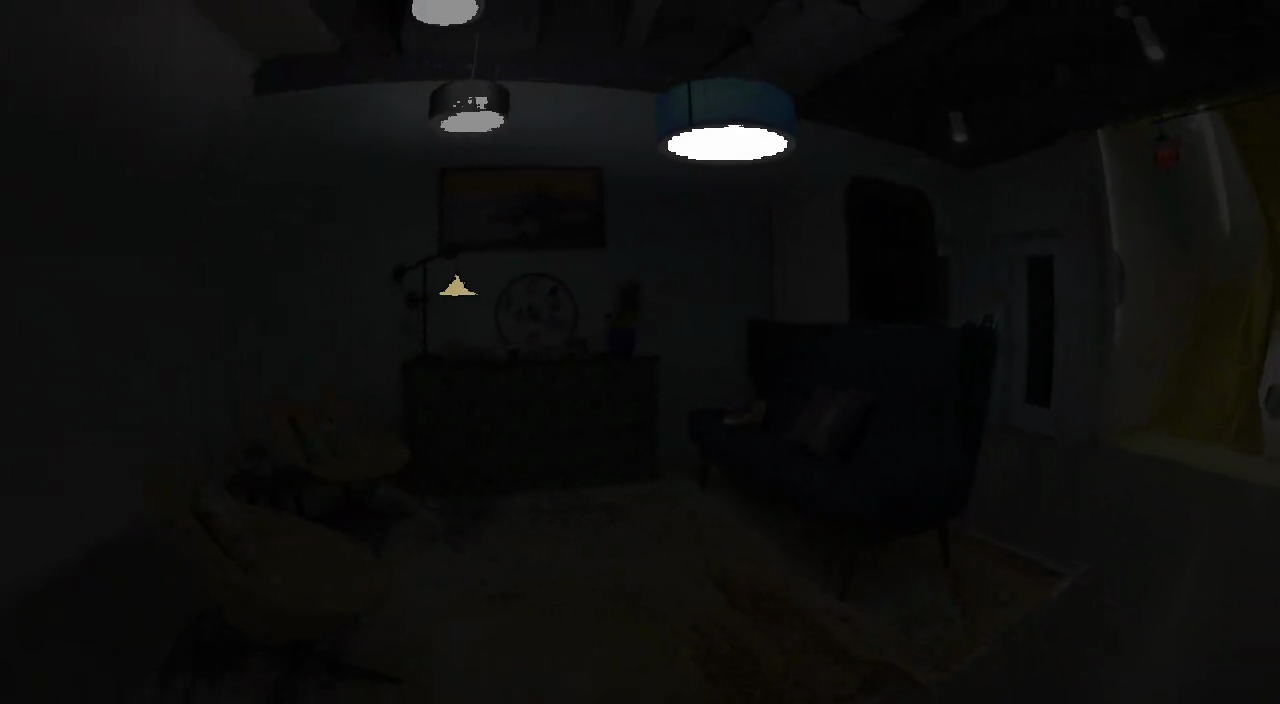} &
    \includegraphics[width=\cellW,height=\cellH,keepaspectratio]{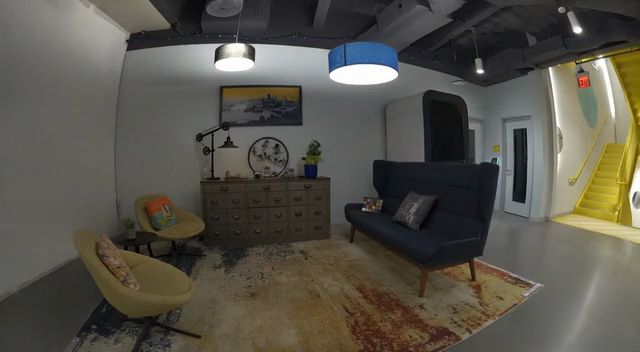} &
    \includegraphics[width=\cellW,height=\cellH,keepaspectratio]{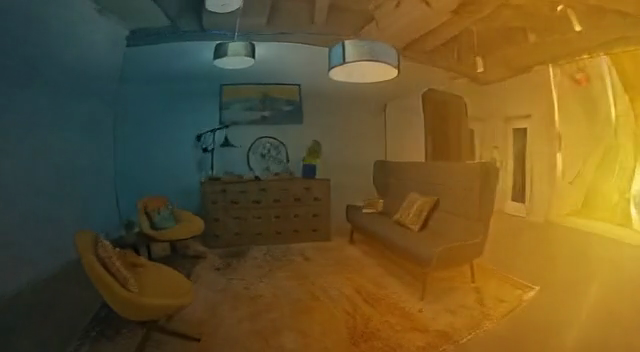} &
    \includegraphics[width=\cellW,height=\cellH,keepaspectratio]{figures/figs_resource/failed.pdf} &
    \includegraphics[width=\cellW,height=\cellH,keepaspectratio]{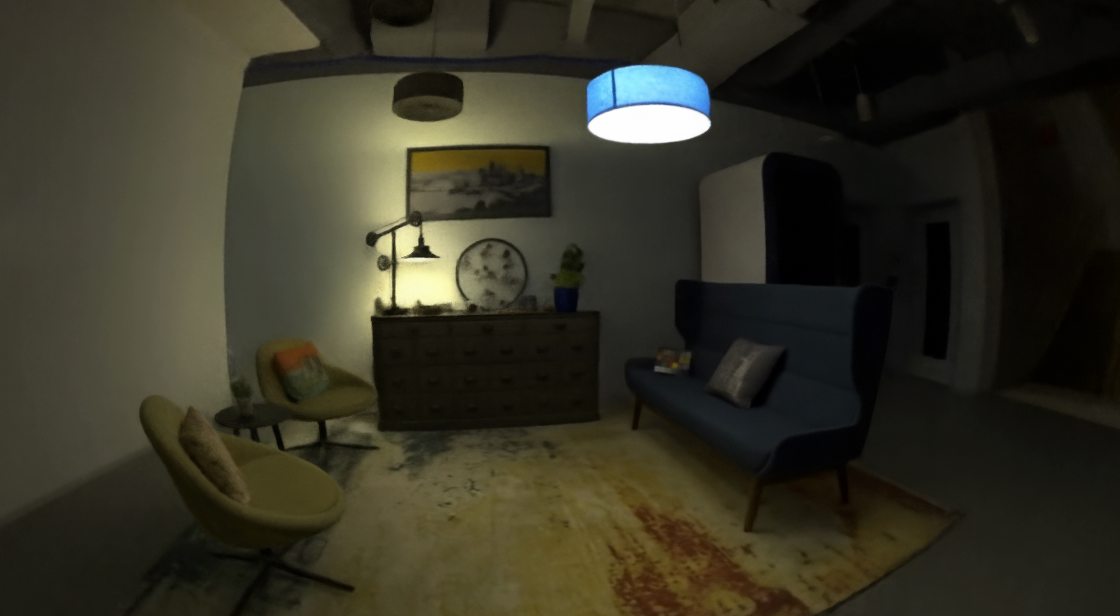} 
    \\[-2pt]
    \includegraphics[width=\cellW,height=\cellH,keepaspectratio]{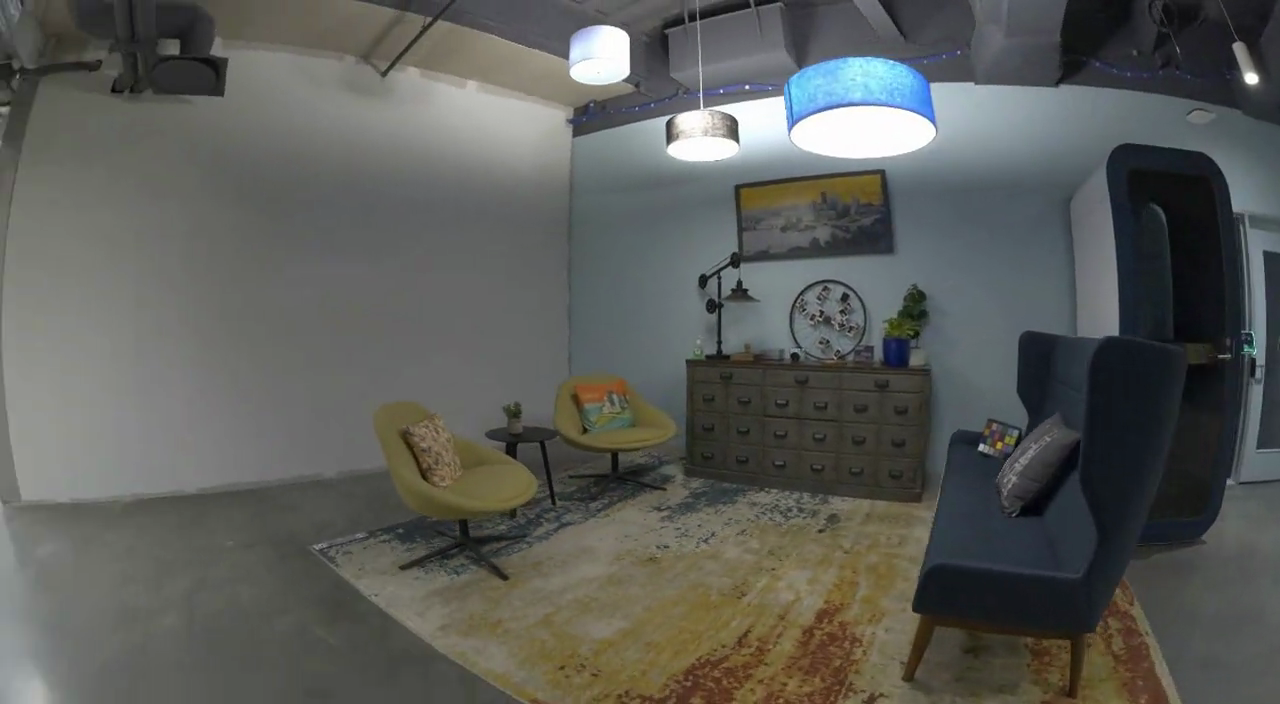} &
    \includegraphics[width=\cellW,height=\cellH,keepaspectratio]{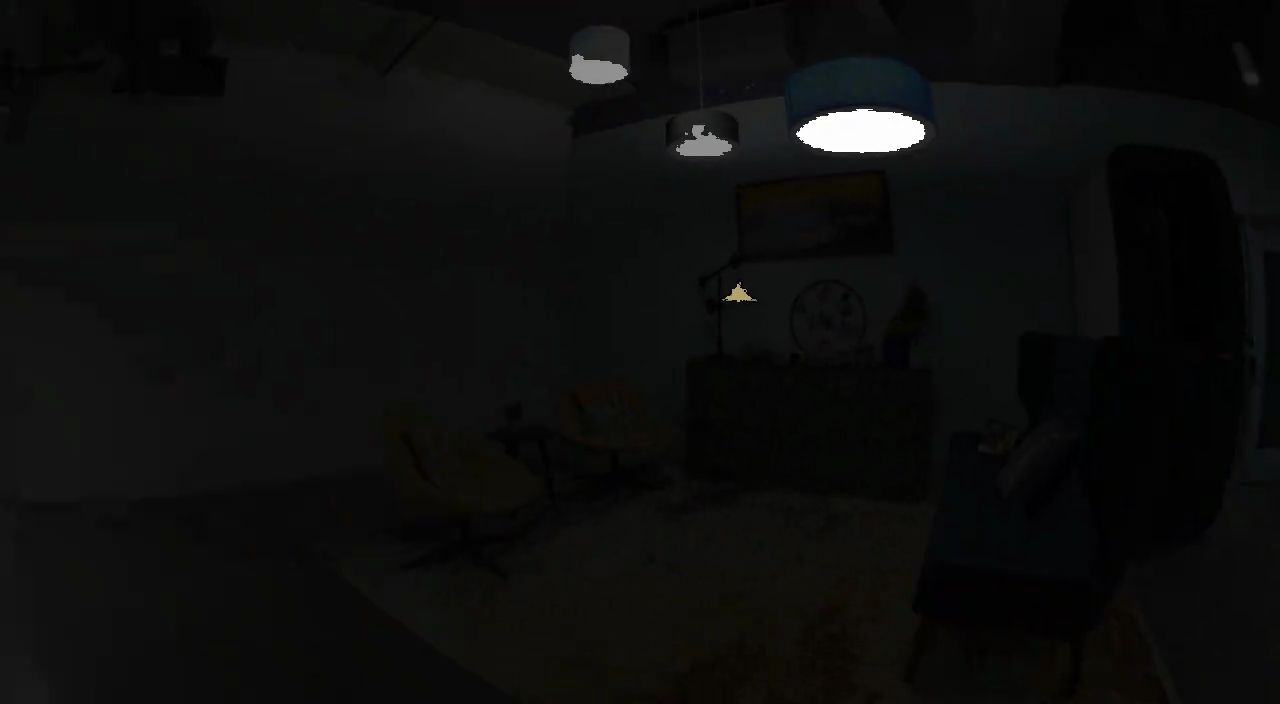} &
    \includegraphics[width=\cellW,height=\cellH,keepaspectratio]{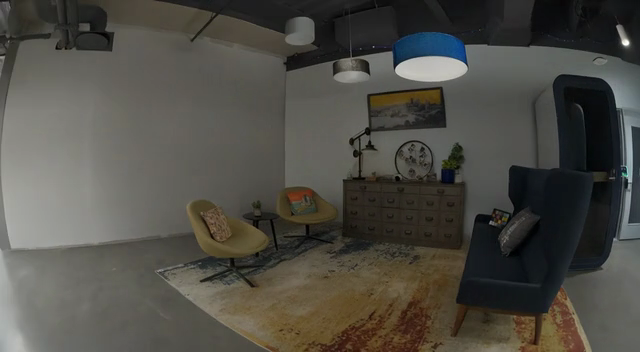} &
    \includegraphics[width=\cellW,height=\cellH,keepaspectratio]{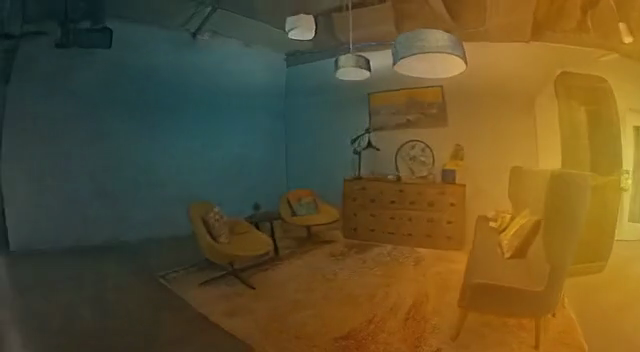} &
    \includegraphics[width=\cellW,height=\cellH,keepaspectratio]{figures/figs_resource/failed.pdf} &
    \includegraphics[width=\cellW,height=\cellH,keepaspectratio]{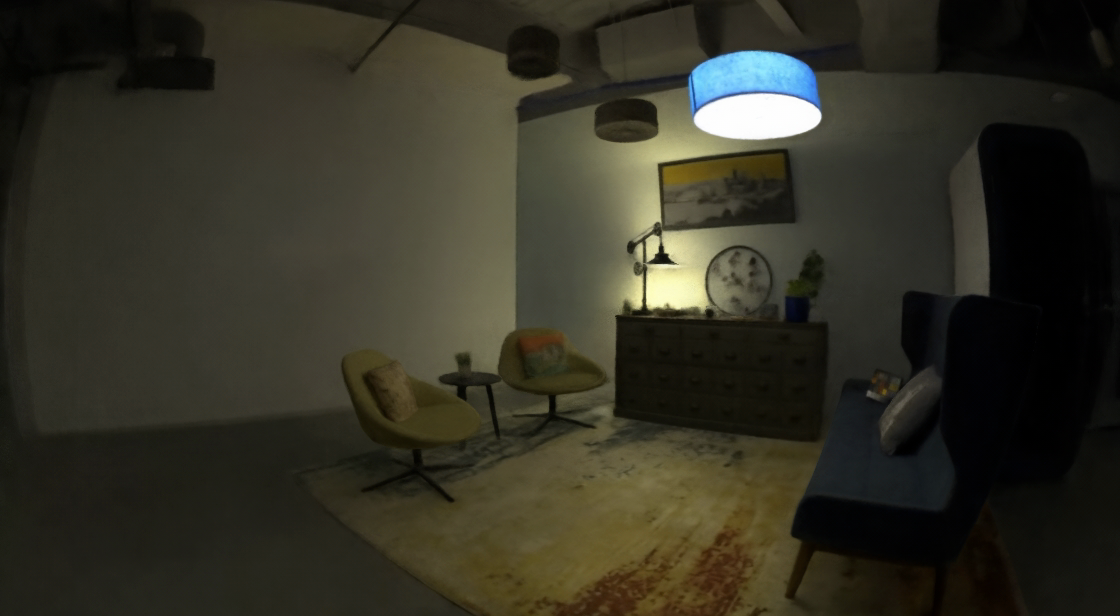} \\
        \parbox[c][\baselineskip][c]{\cellW}{\centering\small Input} &
    \parbox[c][\baselineskip][c]{\cellW}{\centering\small Condition} &
    \parbox[c][\baselineskip][c]{\cellW}{\centering\small LightLab} &
    \parbox[c][\baselineskip][c]{\cellW}{\centering\small Light-A-Video} &
    \parbox[c][\baselineskip][c]{\cellW}{\centering\small DiffusionRenderer} &
    \parbox[c][\baselineskip][c]{\cellW}{\centering\small Ours}
  \end{tabular}
  \caption{\textbf{Additional real-world results.} Qualitative comparisons of our method with LightLab, Light-A-Video, and DiffusionRenderer on the \texttt{seating\_area} scene from the Eyeful Tower dataset. Since LightLab separately relights each frame of the video it suffers from flickering temporal artifacts. Light-A-Video provides very limited control and low-quality results. DiffusionRenderer provides consistent but low-quality and blurry results. Our approach is able to realistically relight the scene according to the provided target lights.}
  \label{fig:seating}
\end{figure*}

\setlength{\cellW}{0.16\textwidth}

\setlength{\cellH}{0.5625\cellW}

\setlength{\rowLabelW}{0.03\textwidth}

\begin{figure*}[t]
  \centering
  \setlength{\tabcolsep}{2pt}          %
  \renewcommand{\arraystretch}{0.8}    %
  \begin{tabular}{@{}c*{5}{c}@{}}

    \includegraphics[width=\cellW,height=\cellH,keepaspectratio]{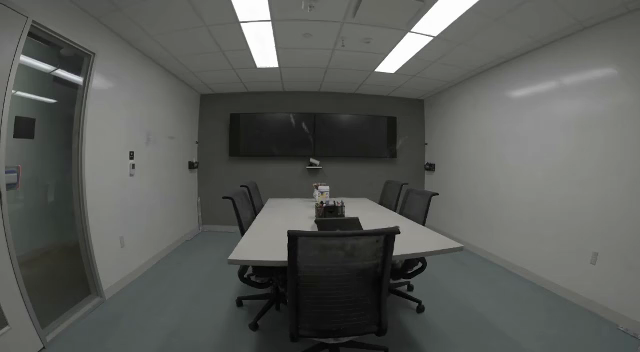} &
    \includegraphics[width=\cellW,height=\cellH,keepaspectratio]{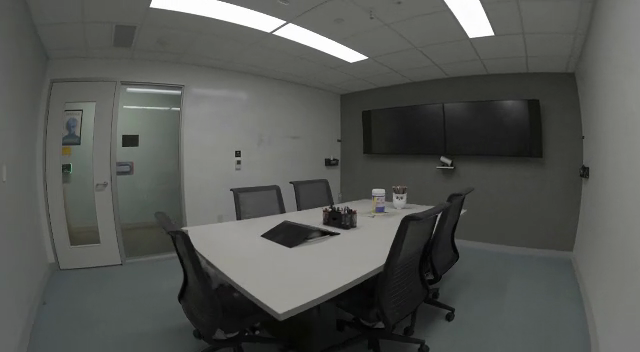} &
    \includegraphics[width=\cellW,height=\cellH,keepaspectratio]{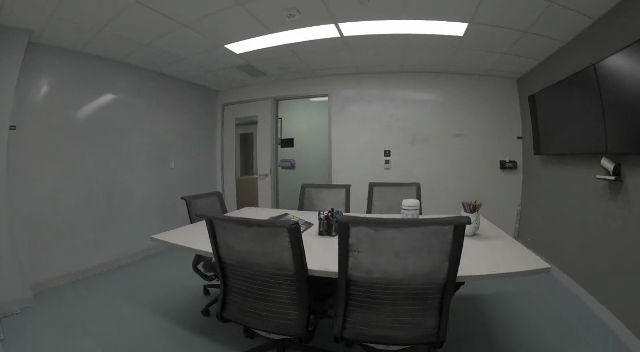} &
    \includegraphics[width=\cellW,height=\cellH,keepaspectratio]{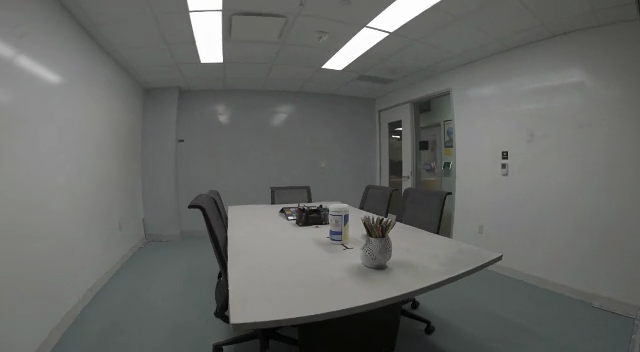} &
    \includegraphics[width=\cellW,height=\cellH,keepaspectratio]{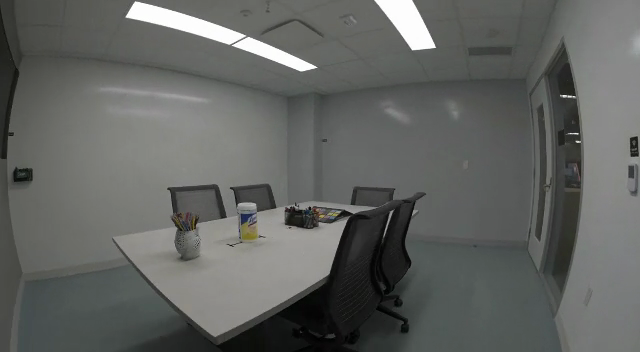} &
    \includegraphics[width=\cellW,height=\cellH,keepaspectratio]{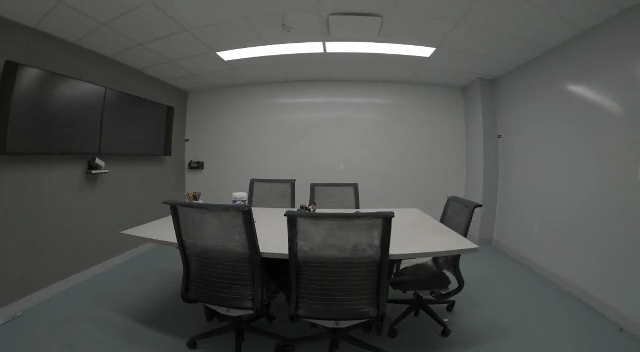}  \\[-2pt]
    \includegraphics[width=\cellW,height=\cellH,keepaspectratio]{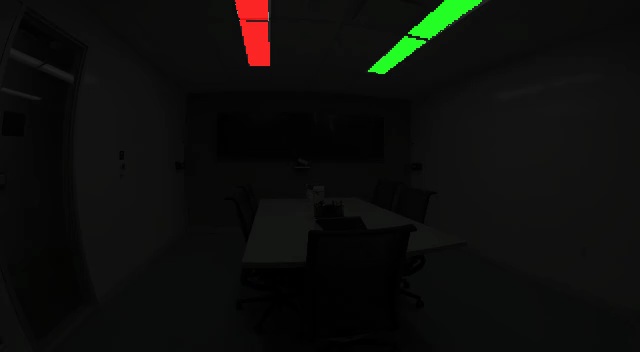} &
    \includegraphics[width=\cellW,height=\cellH,keepaspectratio]{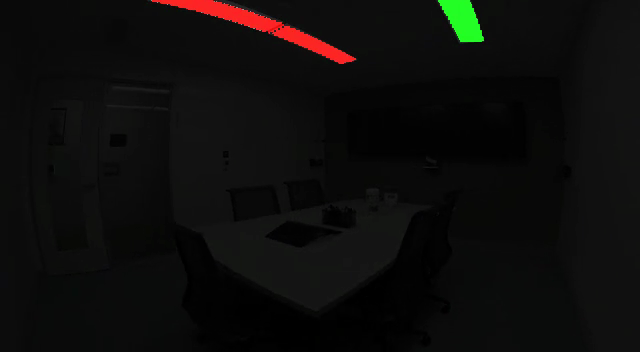} &
    \includegraphics[width=\cellW,height=\cellH,keepaspectratio]{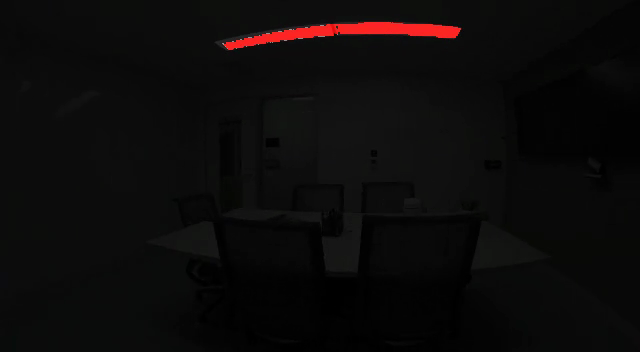} &
    \includegraphics[width=\cellW,height=\cellH,keepaspectratio]{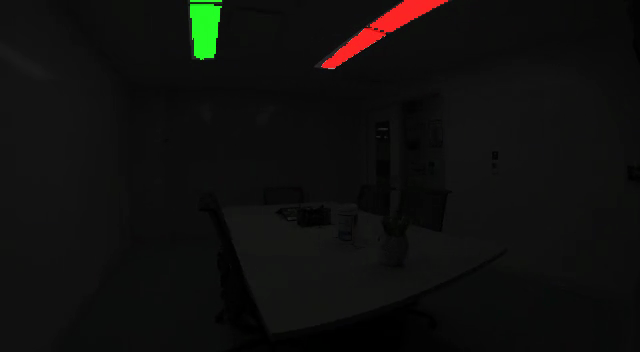} &
    \includegraphics[width=\cellW,height=\cellH,keepaspectratio]{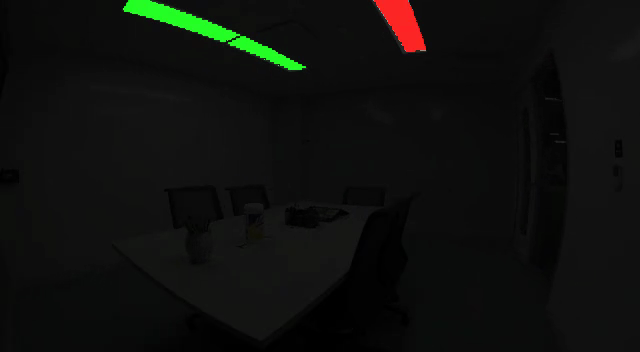} &
    \includegraphics[width=\cellW,height=\cellH,keepaspectratio]{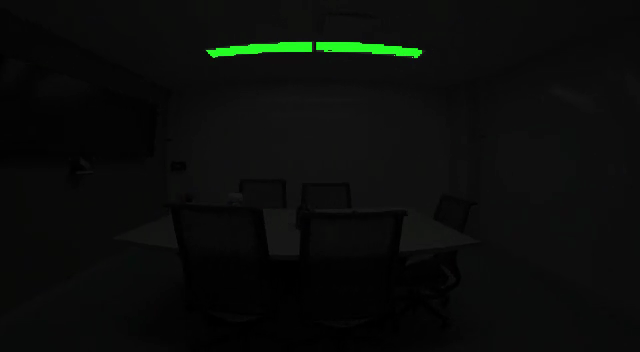} \\[-2pt]
    \includegraphics[width=\cellW,height=\cellH,keepaspectratio]{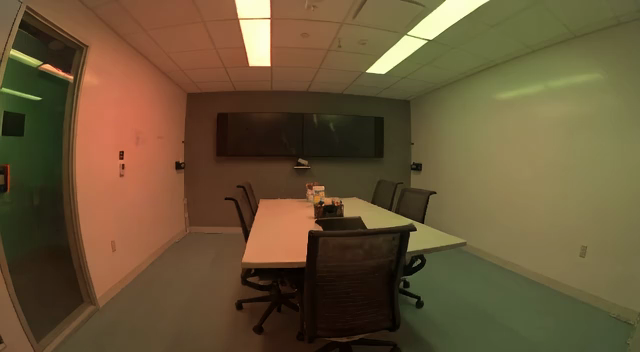} &
    \includegraphics[width=\cellW,height=\cellH,keepaspectratio]{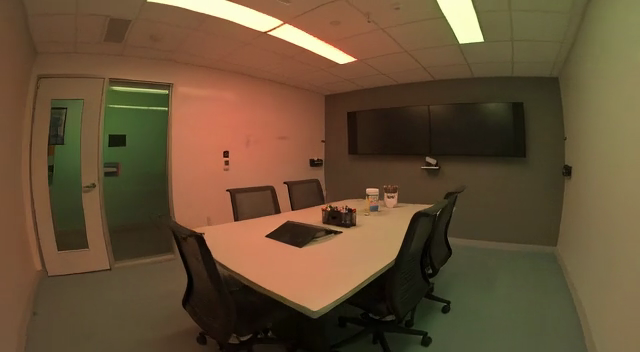} &
    \includegraphics[width=\cellW,height=\cellH,keepaspectratio]{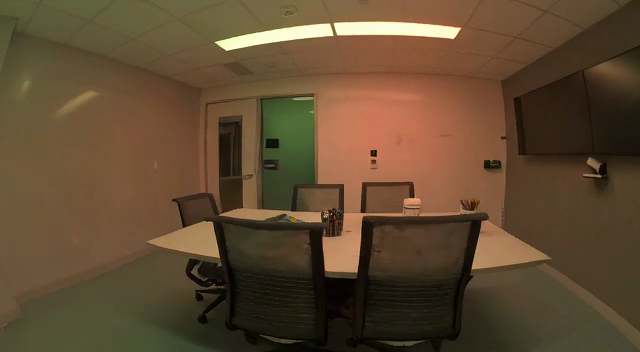} &
    \includegraphics[width=\cellW,height=\cellH,keepaspectratio]{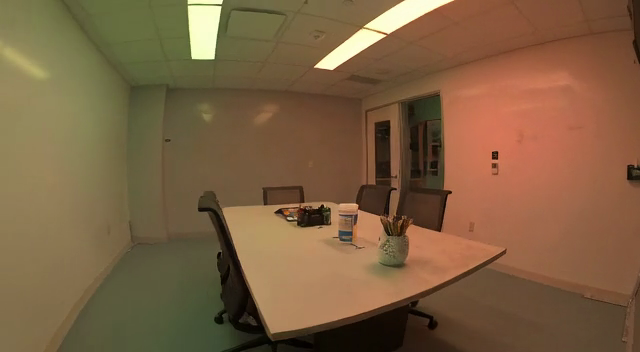} &
    \includegraphics[width=\cellW,height=\cellH,keepaspectratio]{figures/figs_resource/office_1a_color/rg_our0045.png} &
    \includegraphics[width=\cellW,height=\cellH,keepaspectratio]{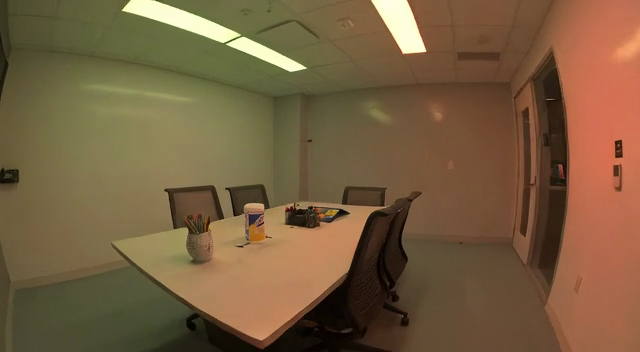} \\
  \end{tabular}
  \caption{\textbf{Fine-grained color control on a real-world scene.} Top row: input video frames. Middle row: target lighting conditions indicating the ceiling lamps are emitting red and green respectively. Bottom row: our relit results. Our method is able to plausibly relight the scene even when the target lighting colors are non-ordinary.}%
  \label{fig:extra_color}
\end{figure*}

\setlength{\cellW}{0.16\textwidth}

\setlength{\cellH}{0.5625\cellW}

\setlength{\rowLabelW}{0.03\textwidth}

\begin{figure*}[p]
  \centering
  \setlength{\tabcolsep}{2pt}          %
  \renewcommand{\arraystretch}{0.8}    %
  \begin{tabular}{@{}c*{5}{c}@{}}
    \includegraphics[width=\cellW,height=\cellH,keepaspectratio]{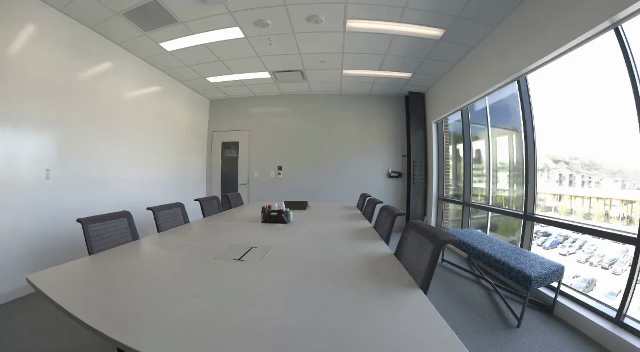} &
    \includegraphics[width=\cellW,height=\cellH,keepaspectratio]{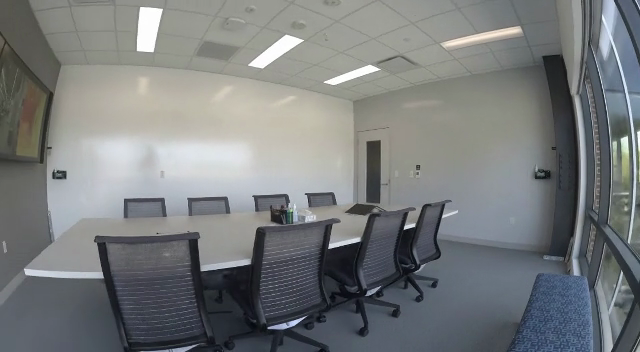} &
    \includegraphics[width=\cellW,height=\cellH,keepaspectratio]{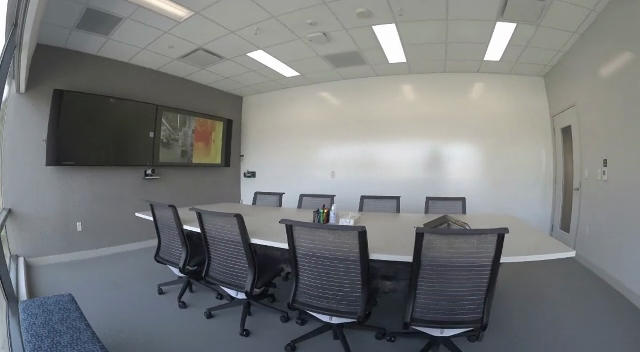} &
    \includegraphics[width=\cellW,height=\cellH,keepaspectratio]{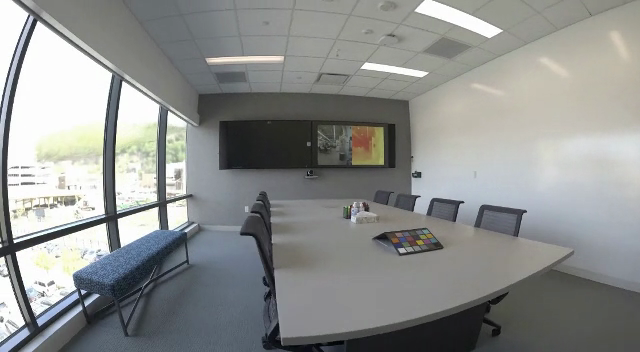} &
    \includegraphics[width=\cellW,height=\cellH,keepaspectratio]{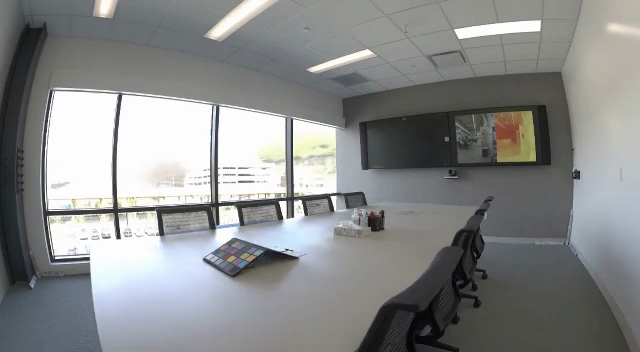} &
    \includegraphics[width=\cellW,height=\cellH,keepaspectratio]{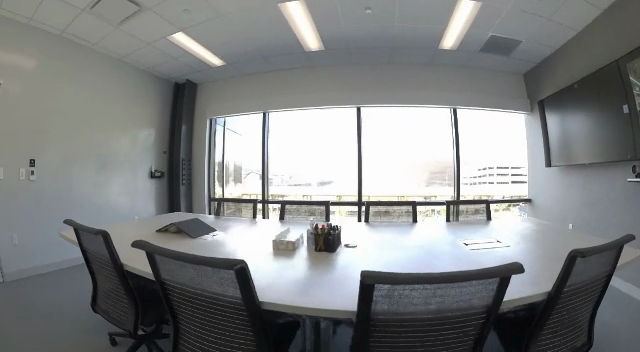}  \\[-1pt]
    \multicolumn{6}{c}{\small (a) Original lighting: all indoor lights on with exterior illumination} \\[1mm]
    \includegraphics[width=\cellW,height=\cellH,keepaspectratio]{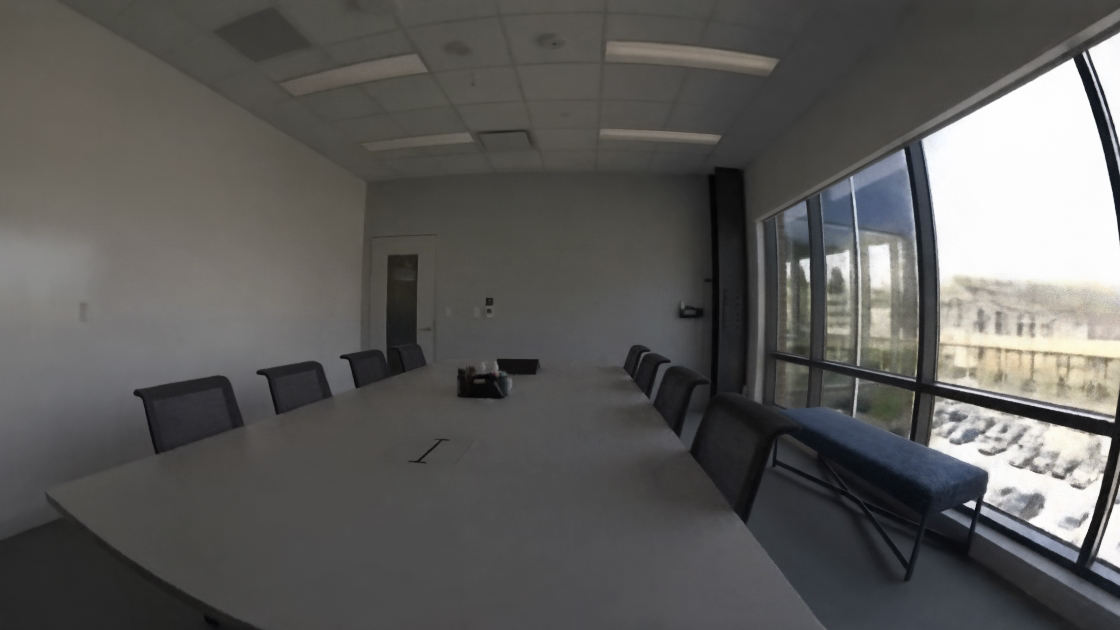} &
    \includegraphics[width=\cellW,height=\cellH,keepaspectratio]{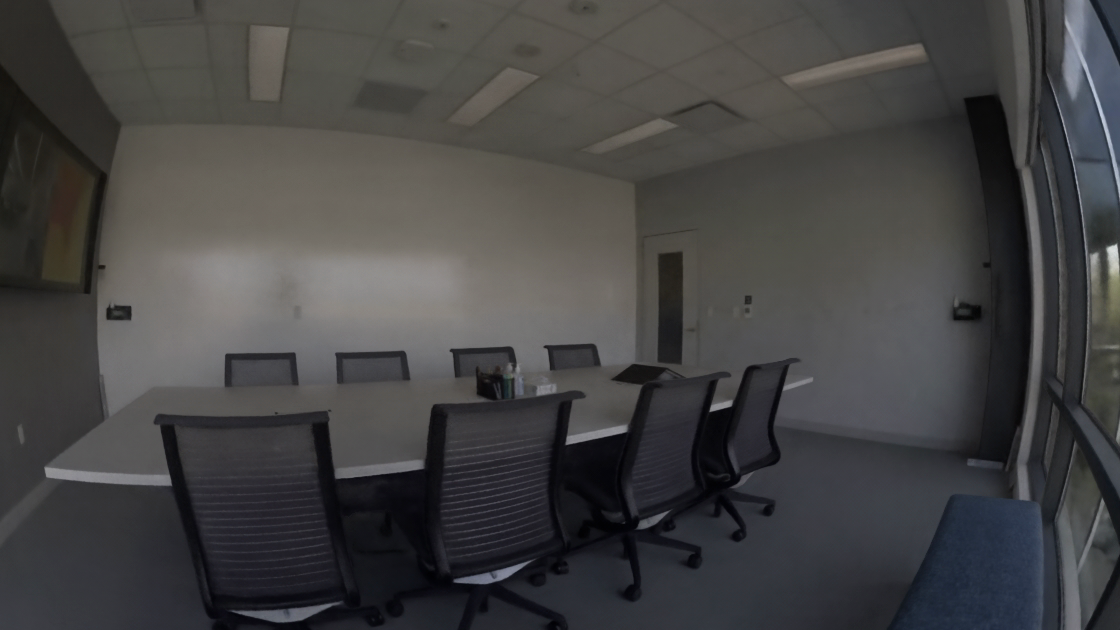} &
    \includegraphics[width=\cellW,height=\cellH,keepaspectratio]{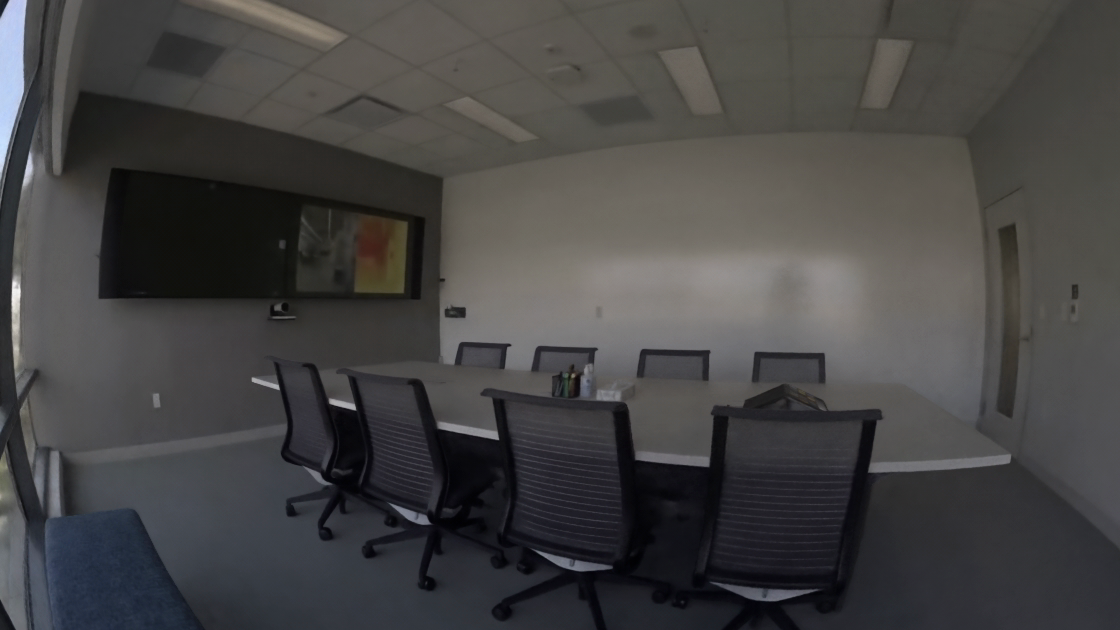} &
    \includegraphics[width=\cellW,height=\cellH,keepaspectratio]{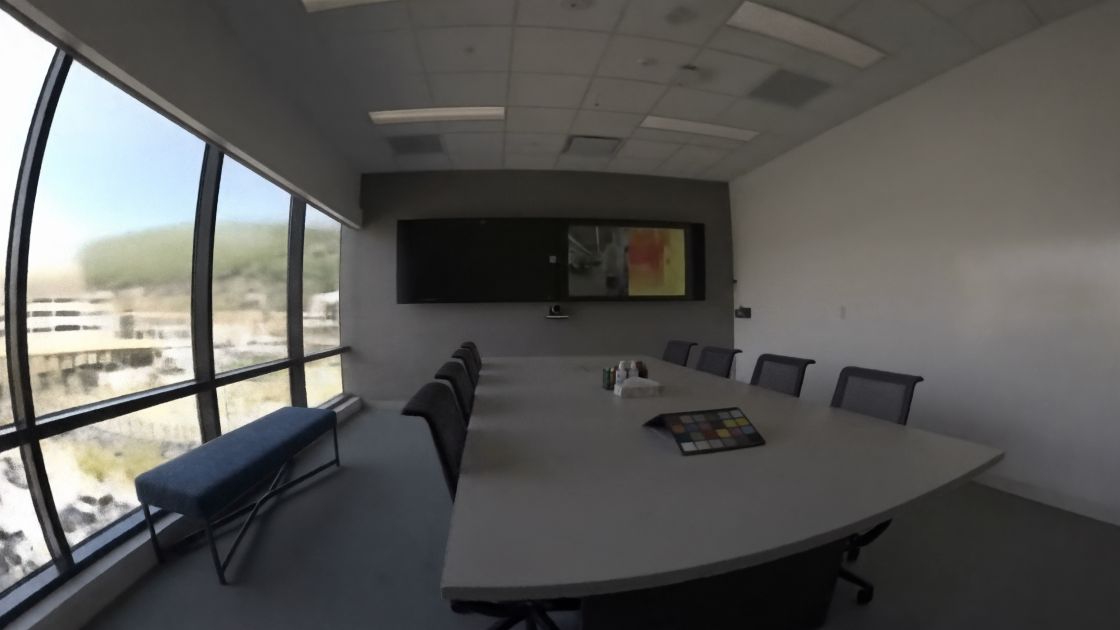} &
    \includegraphics[width=\cellW,height=\cellH,keepaspectratio]{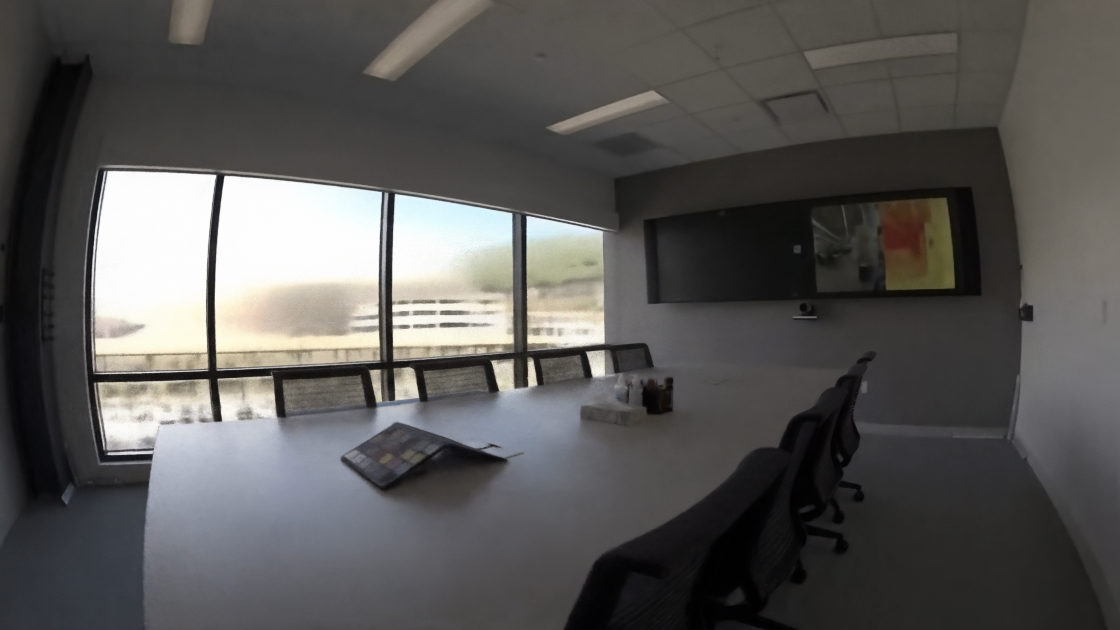} &
    \includegraphics[width=\cellW,height=\cellH,keepaspectratio]{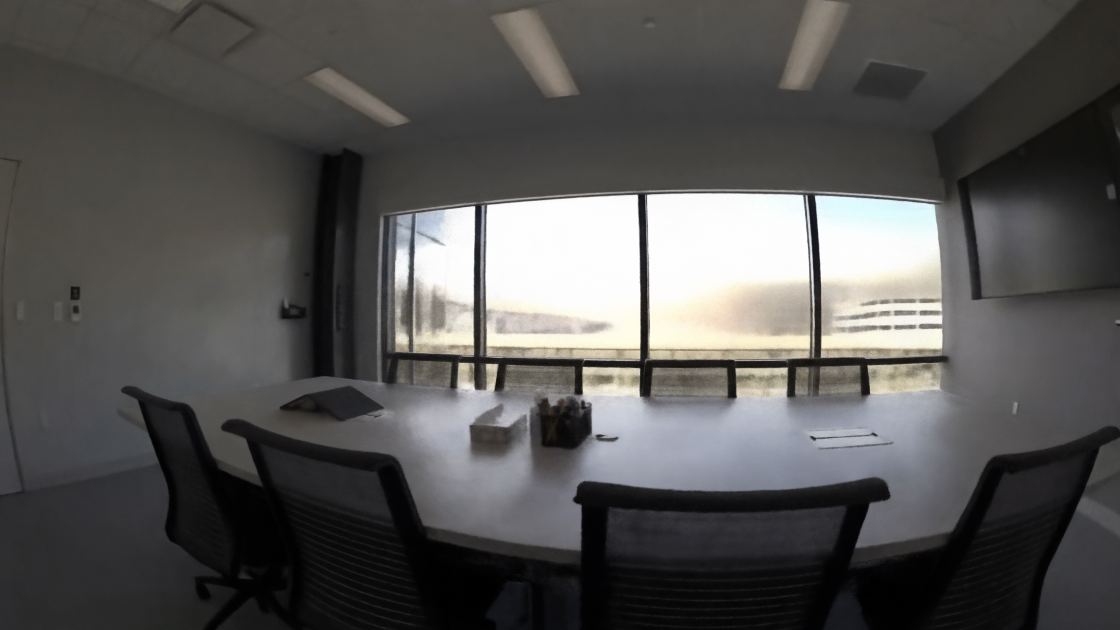} \\[-1pt]
    \multicolumn{6}{c}{\small (b) All indoor lights turned off, exterior light only} \\[1mm]
    \includegraphics[width=\cellW,height=\cellH,keepaspectratio]{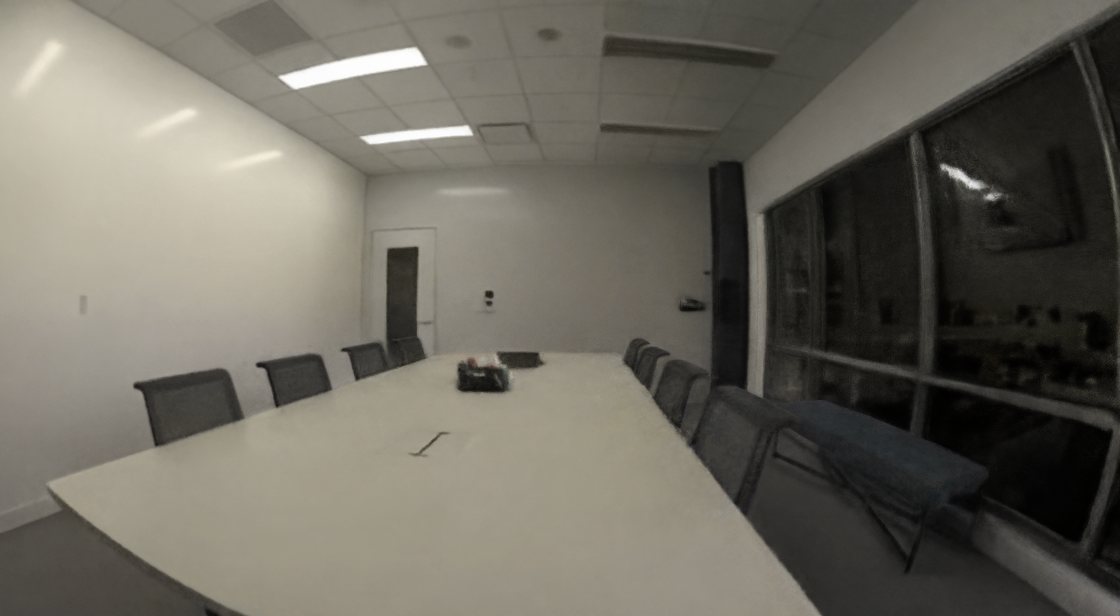} &
    \includegraphics[width=\cellW,height=\cellH,keepaspectratio]{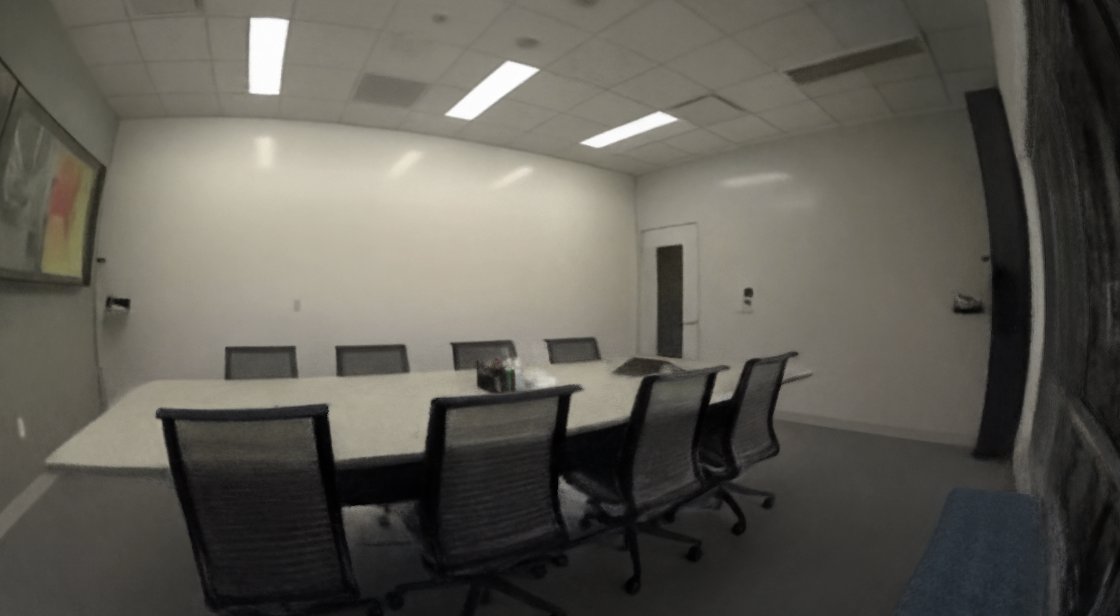} &
    \includegraphics[width=\cellW,height=\cellH,keepaspectratio]{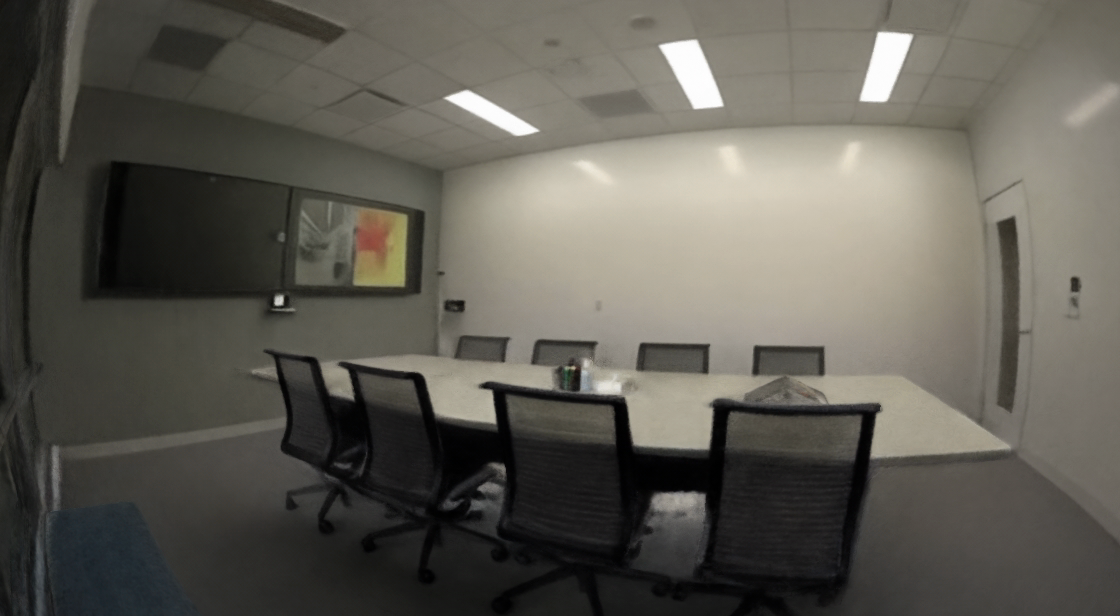} &
    \includegraphics[width=\cellW,height=\cellH,keepaspectratio]{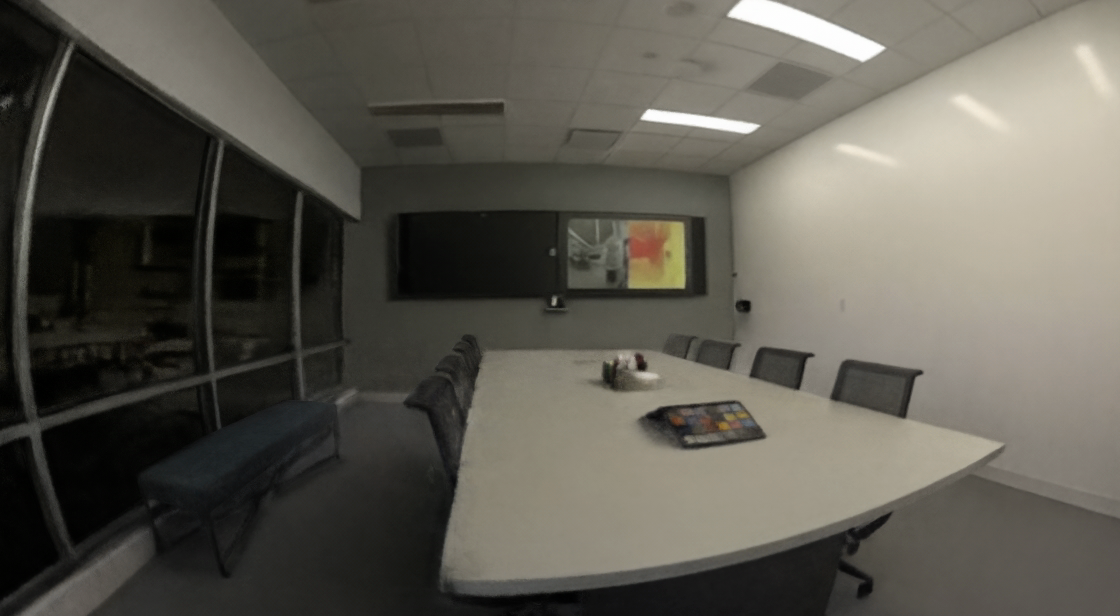} &
    \includegraphics[width=\cellW,height=\cellH,keepaspectratio]{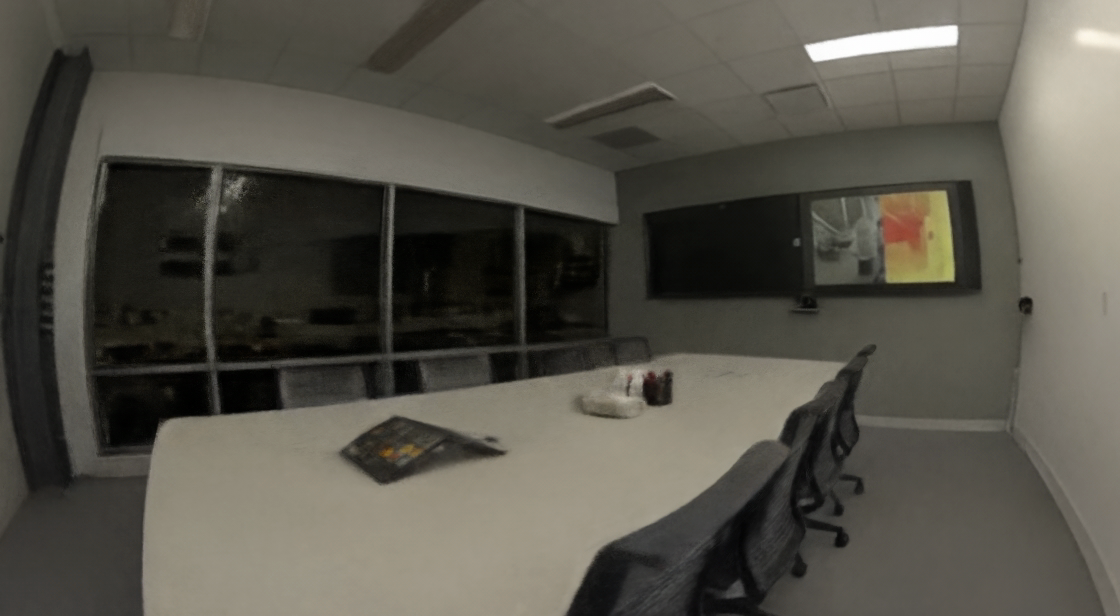} &
    \includegraphics[width=\cellW,height=\cellH,keepaspectratio]{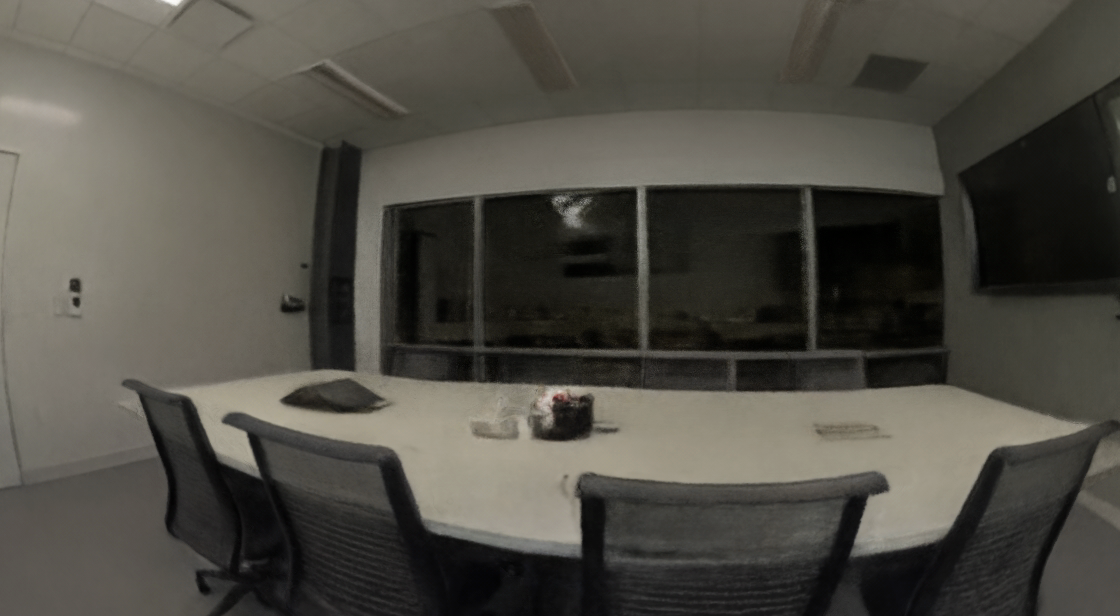} \\[-1pt]
    \multicolumn{6}{c}{\small (c) Partial ceiling lamps turned on, exterior light disabled} \\[1mm]
    \includegraphics[width=\cellW,height=\cellH,keepaspectratio]{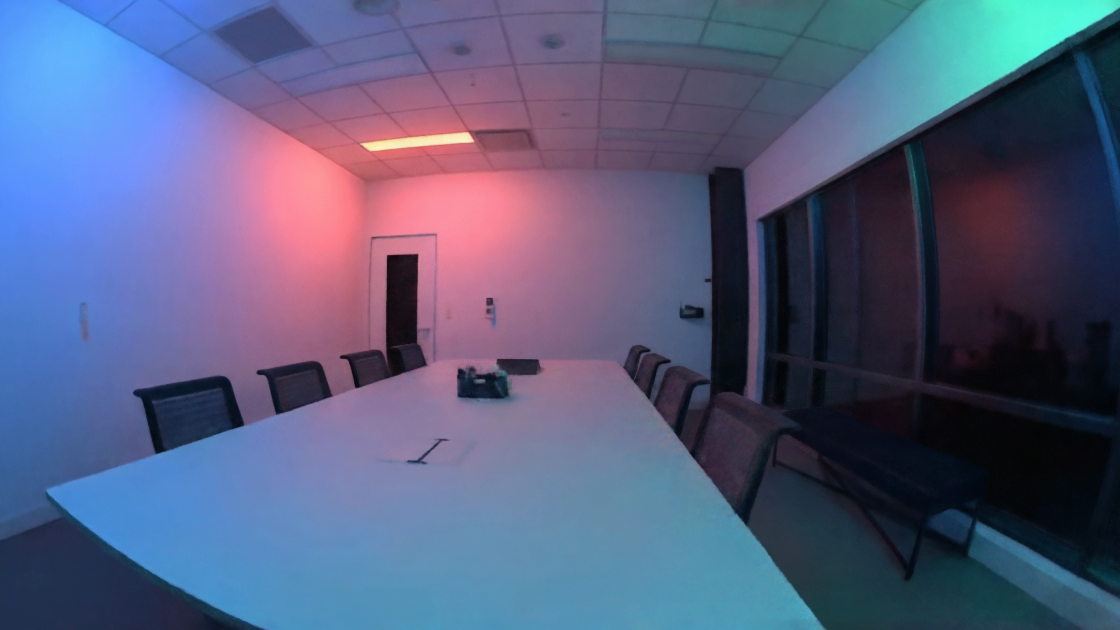} &
    \includegraphics[width=\cellW,height=\cellH,keepaspectratio]{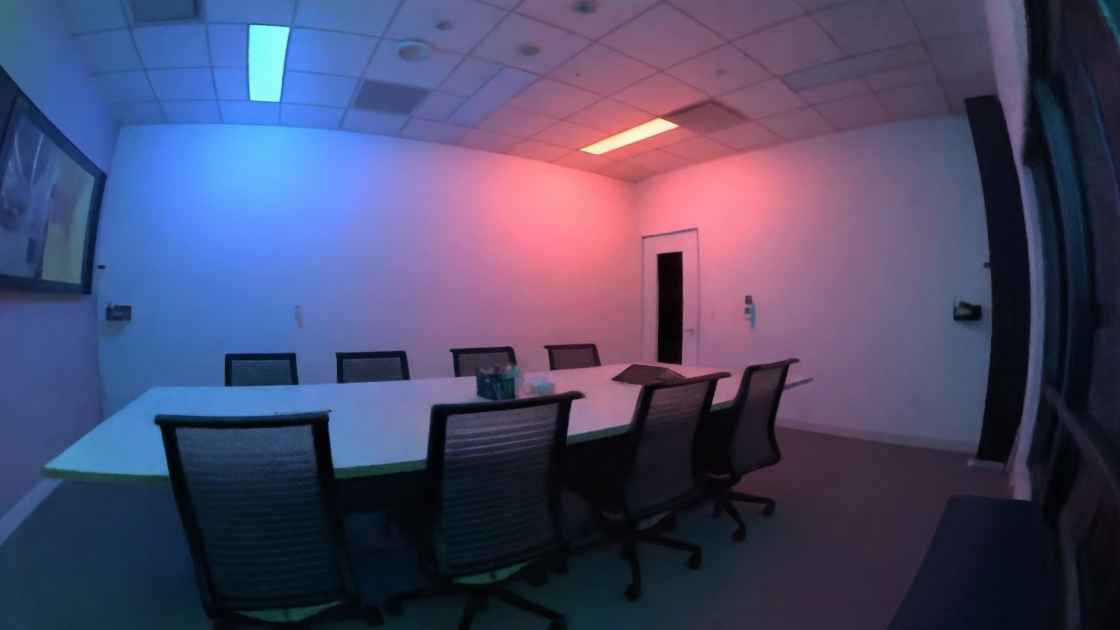} &
    \includegraphics[width=\cellW,height=\cellH,keepaspectratio]{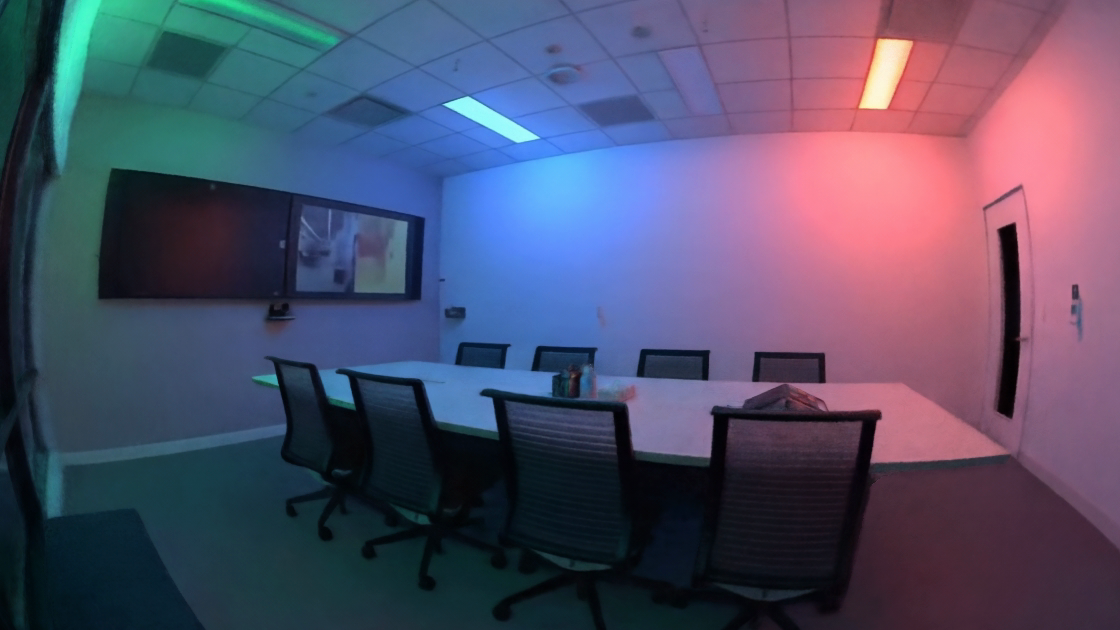} &
    \includegraphics[width=\cellW,height=\cellH,keepaspectratio]{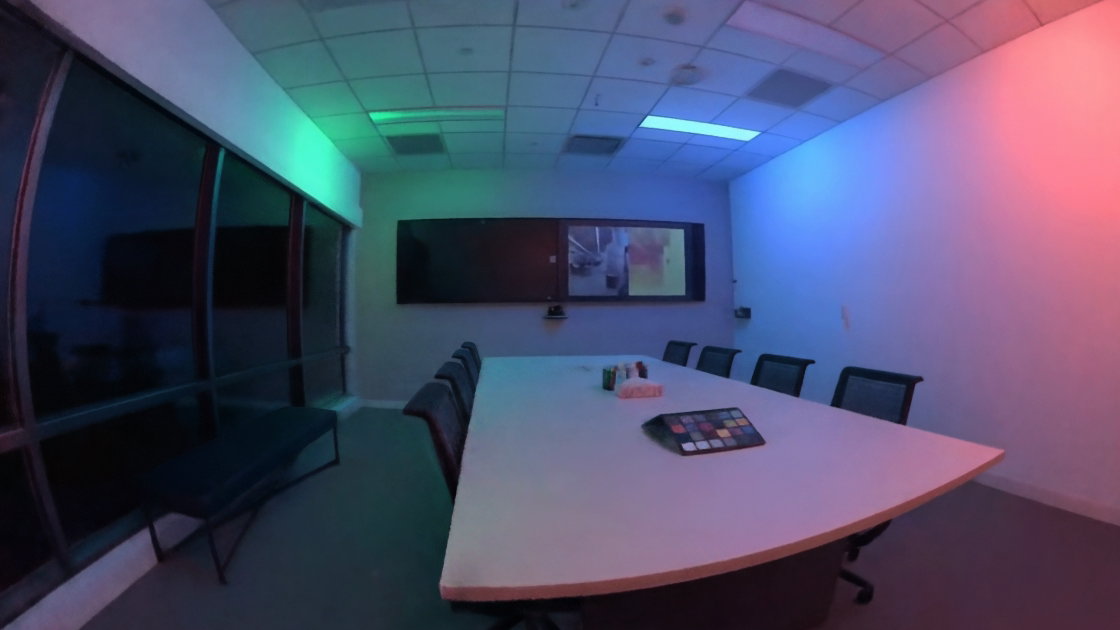} &
    \includegraphics[width=\cellW,height=\cellH,keepaspectratio]{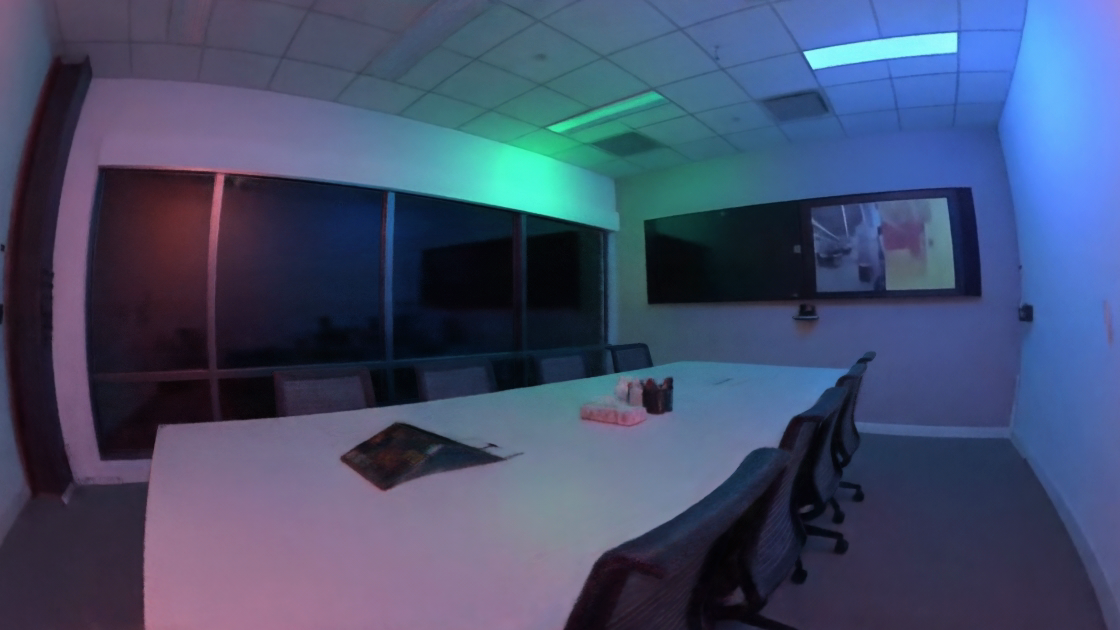} &
    \includegraphics[width=\cellW,height=\cellH,keepaspectratio]{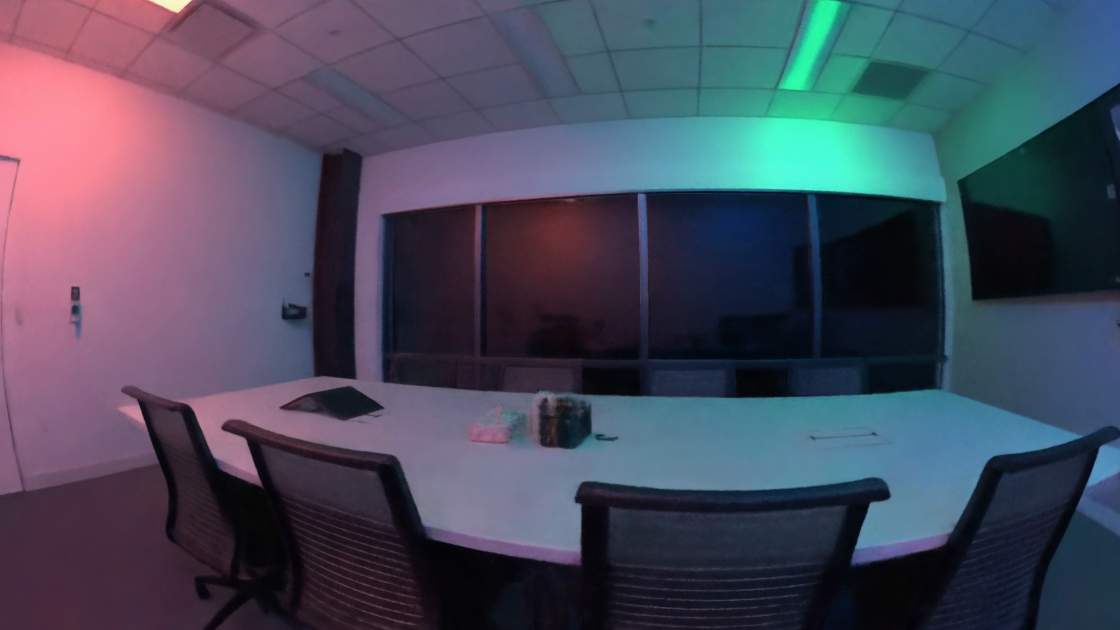} \\[-1pt]
    \multicolumn{6}{c}{\small (d) Selected ceiling lamps on with varying emission colors} \\[1mm]
  \end{tabular}
  \caption{\textbf{Controllable lighting in 3D scenes.} Our method enables fine-grained control over individual light sources in reconstructed environments. Users can selectively toggle indoor and exterior lights (a$\rightarrow$b), activate specific ceiling lamp groups (b$\rightarrow$c), and modify emission colors (c$\rightarrow$d), demonstrating flexible and realistic relighting capabilities.}
  \label{fig:controllable_lighting}
\end{figure*}

\setlength{\cellW}{0.16\textwidth}

\setlength{\cellH}{0.5625\cellW}

\setlength{\rowLabelW}{0.03\textwidth}

\begin{figure*}[p]
  \centering
  \setlength{\tabcolsep}{2pt}          %
  \renewcommand{\arraystretch}{0.8}    %
  \begin{tabular}{@{}c*{5}{c}@{}}

    \includegraphics[width=\cellW,height=\cellH,keepaspectratio]{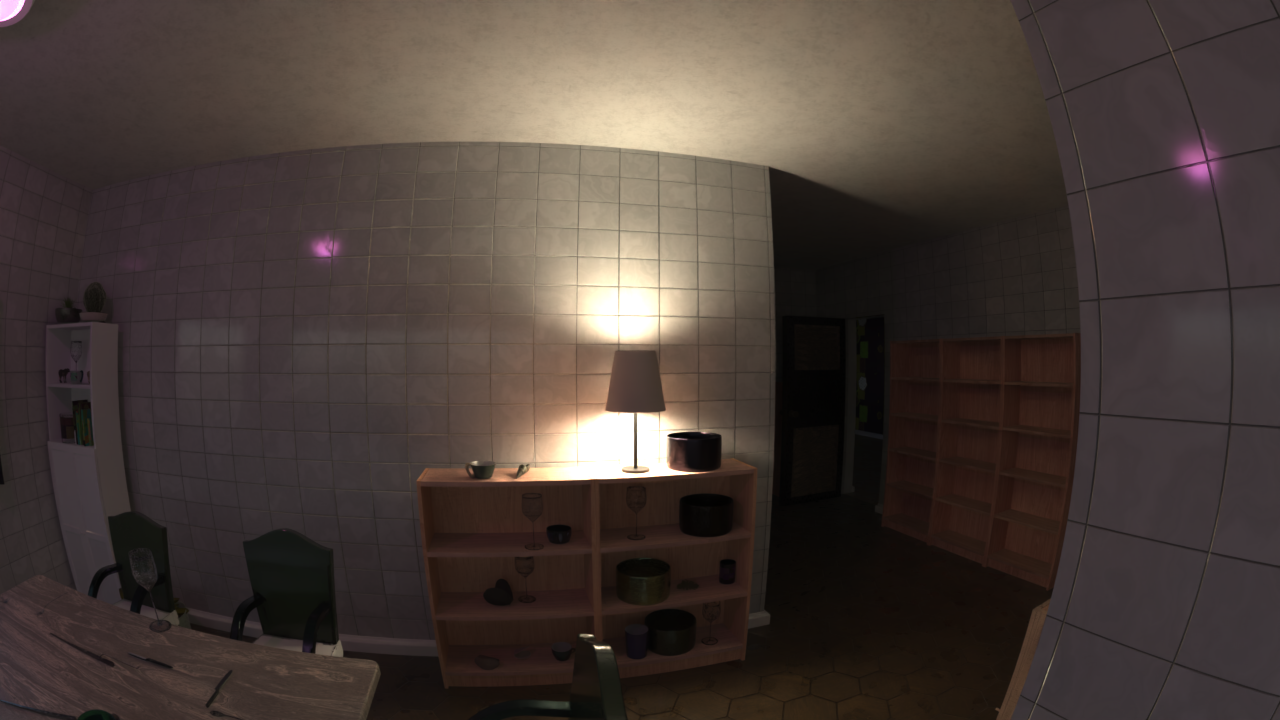} &
    \includegraphics[width=\cellW,height=\cellH,keepaspectratio]{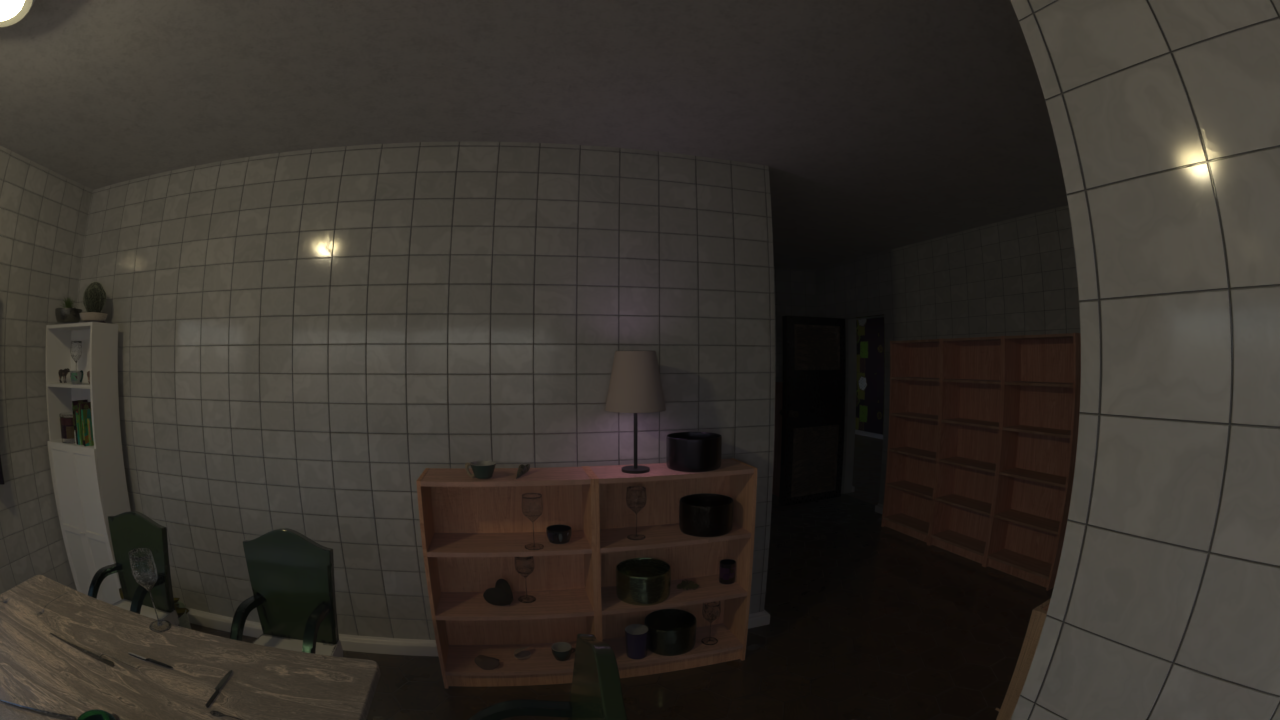} &
    \includegraphics[width=\cellW,height=\cellH,keepaspectratio]{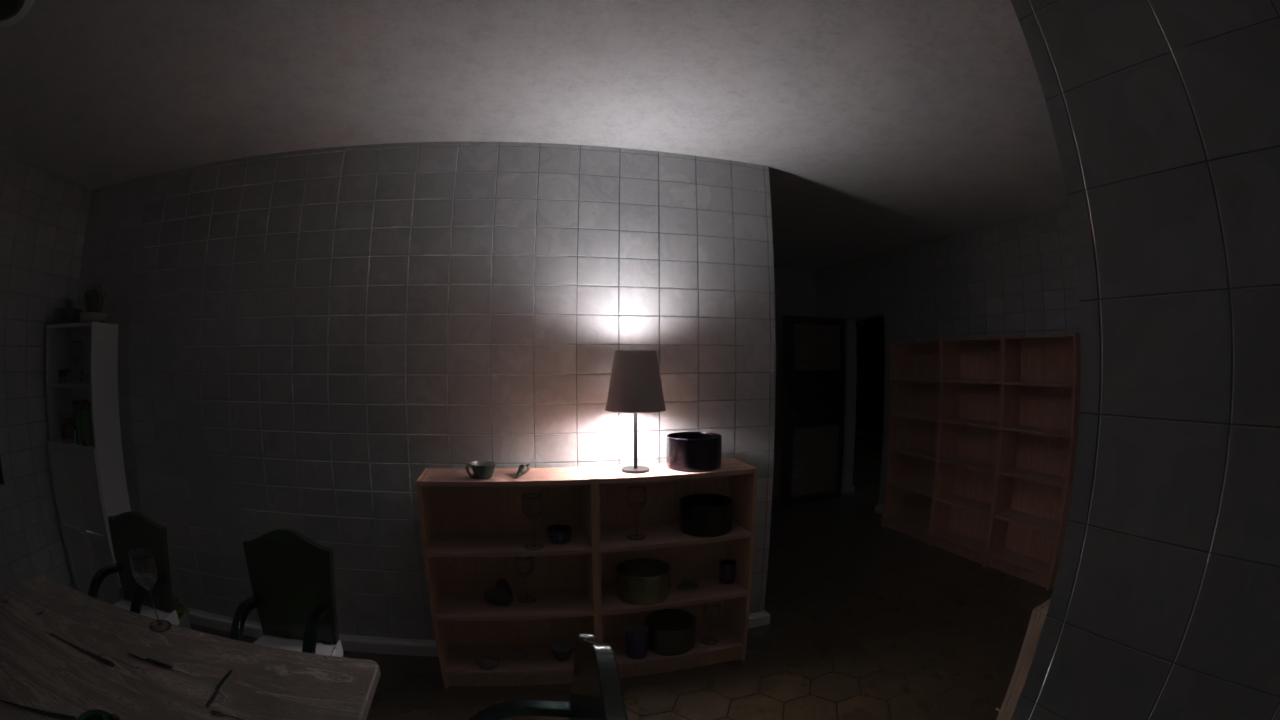} &
    \includegraphics[width=\cellW,height=\cellH,keepaspectratio]{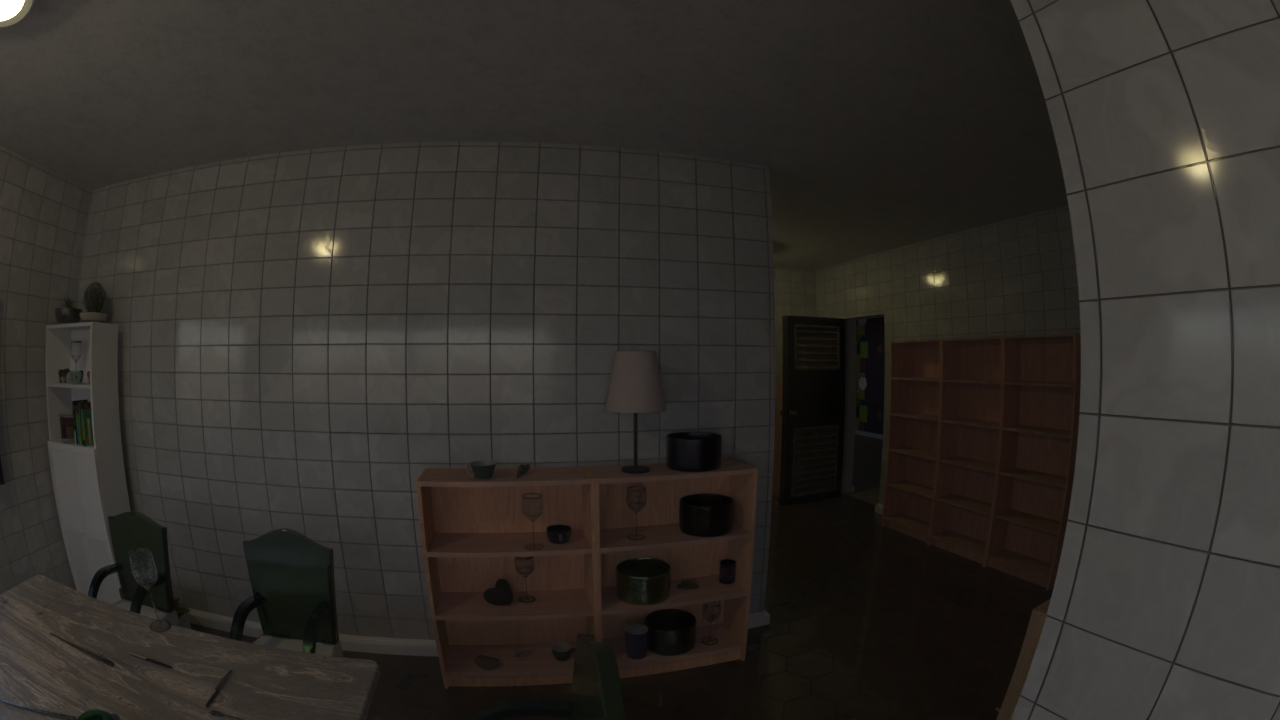} &
    \includegraphics[width=\cellW,height=\cellH,keepaspectratio]{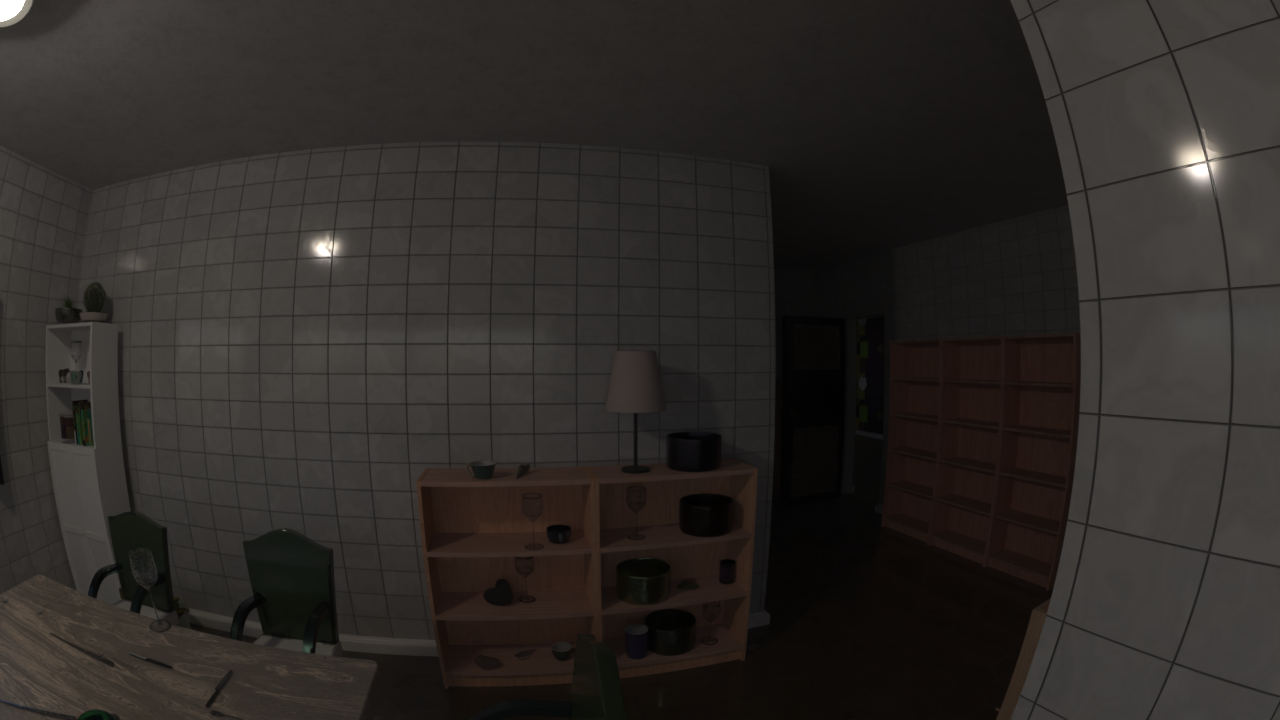} &
    \includegraphics[width=\cellW,height=\cellH,keepaspectratio]{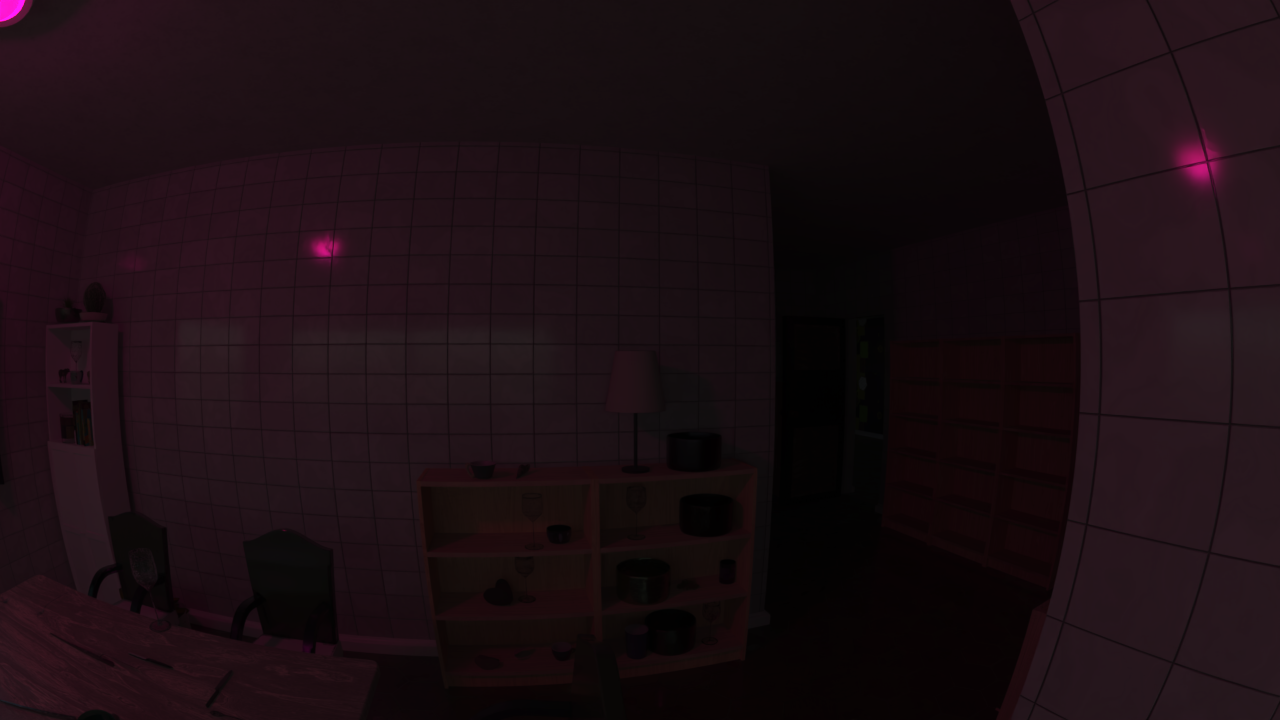}   \\[-2pt]
    \includegraphics[width=\cellW,height=\cellH,keepaspectratio]{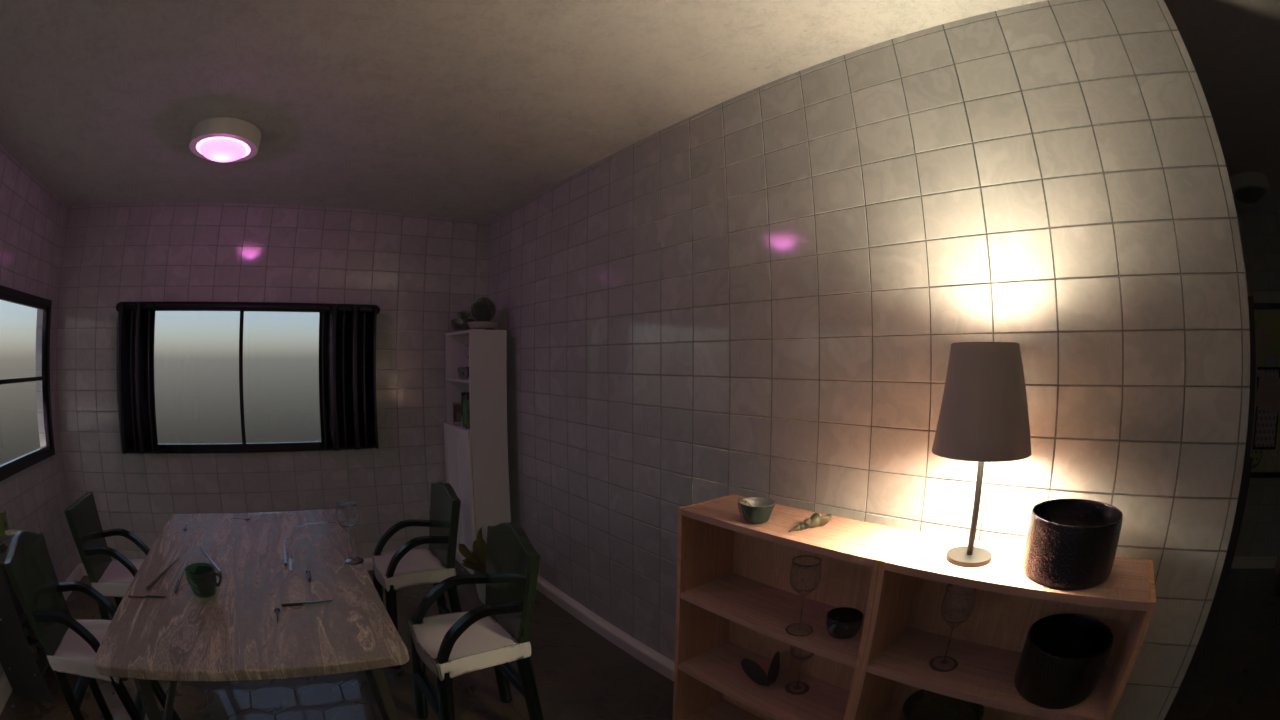} &
    \includegraphics[width=\cellW,height=\cellH,keepaspectratio]{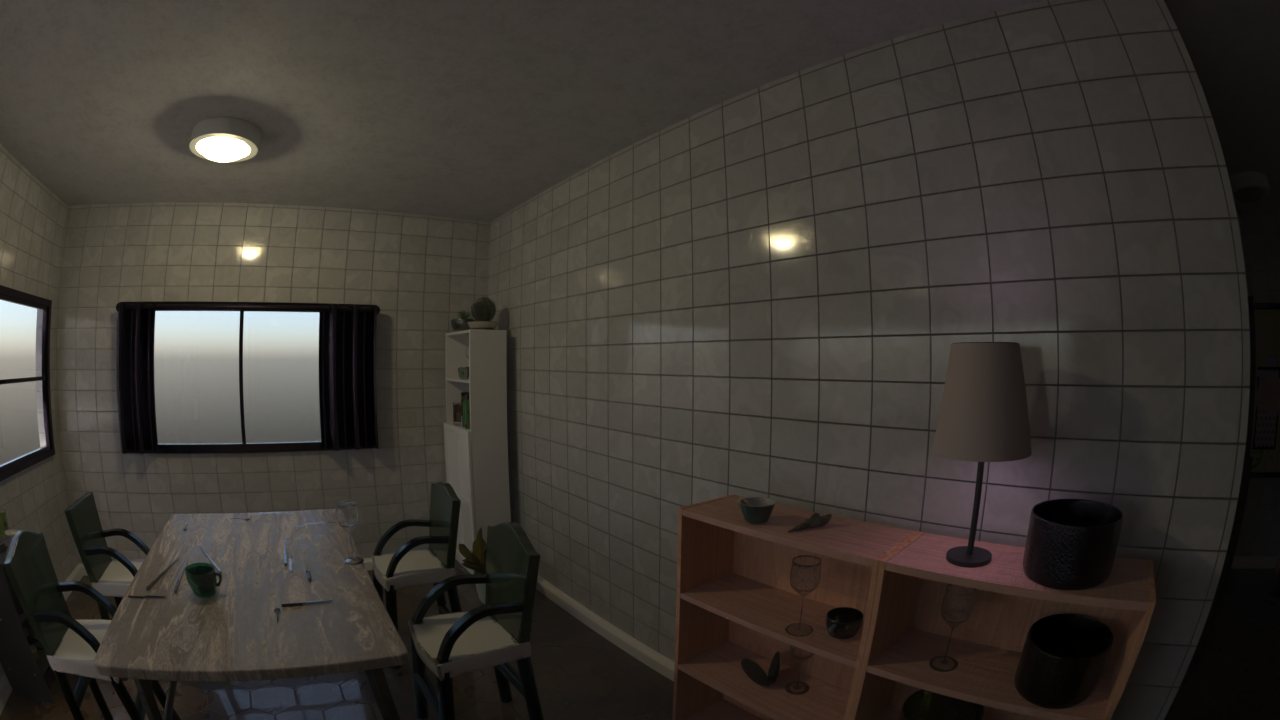} &
    \includegraphics[width=\cellW,height=\cellH,keepaspectratio]{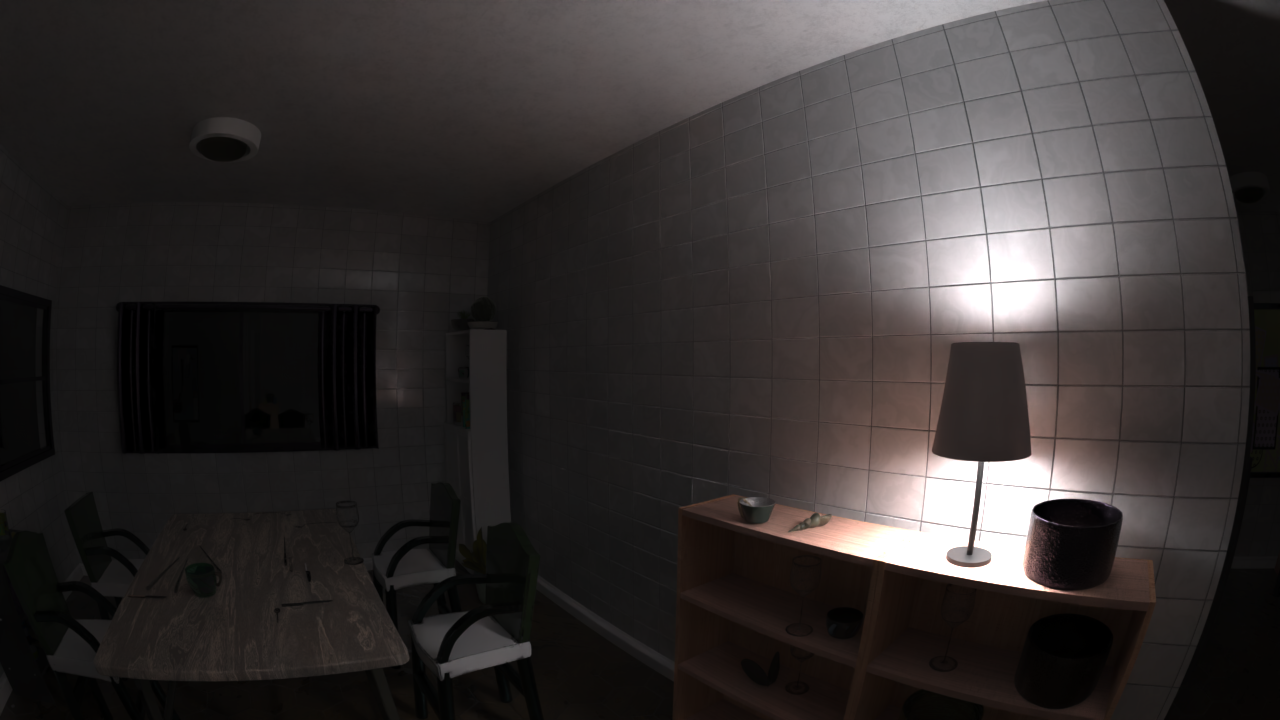} &
    \includegraphics[width=\cellW,height=\cellH,keepaspectratio]{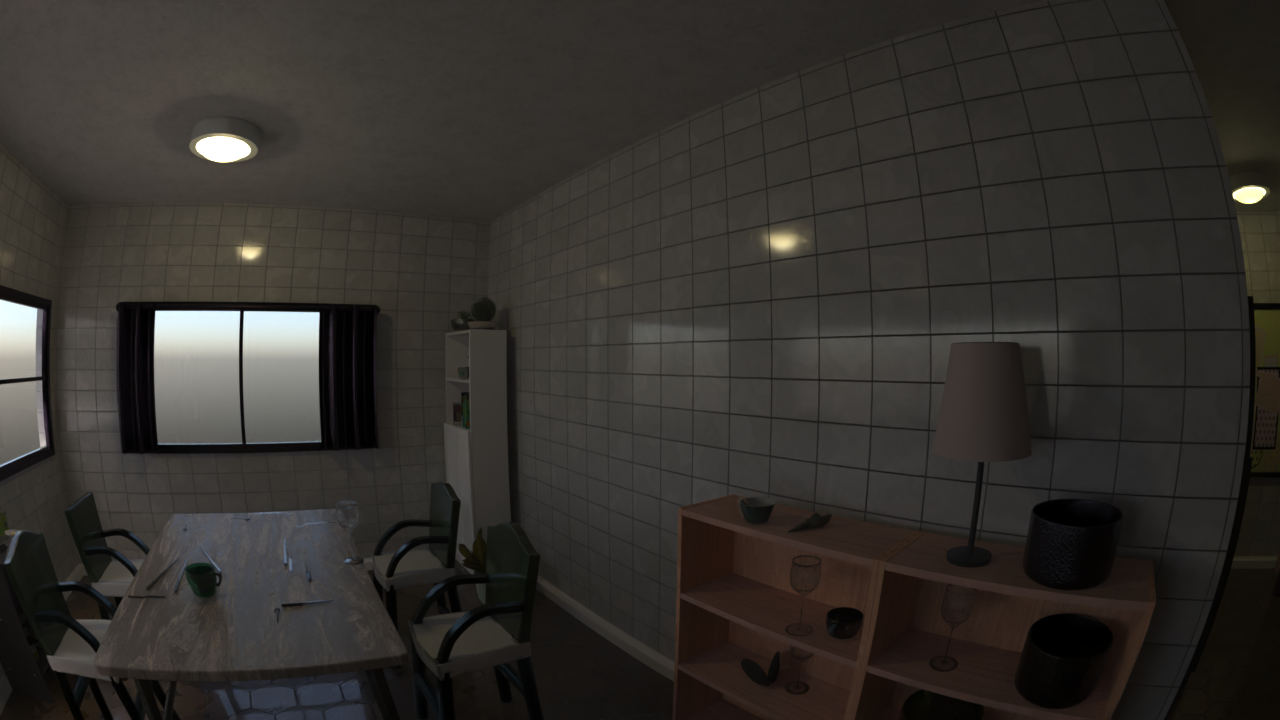} &
    \includegraphics[width=\cellW,height=\cellH,keepaspectratio]{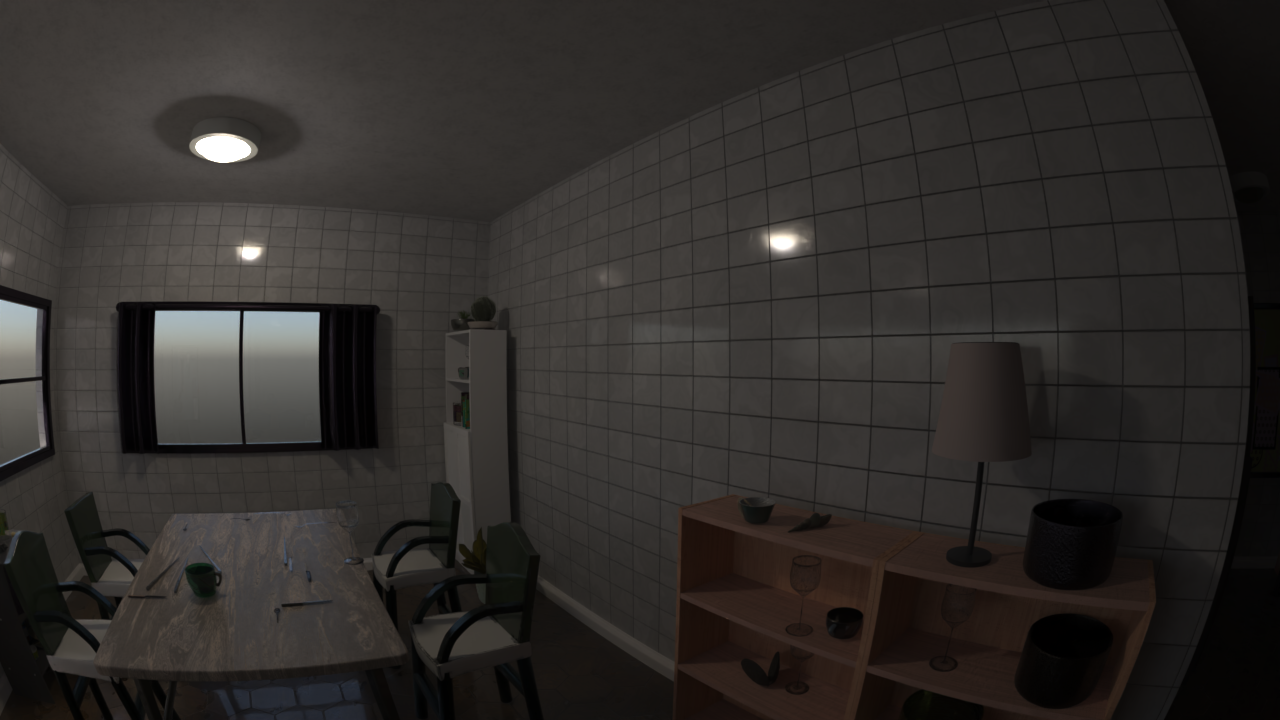} &
    \includegraphics[width=\cellW,height=\cellH,keepaspectratio]{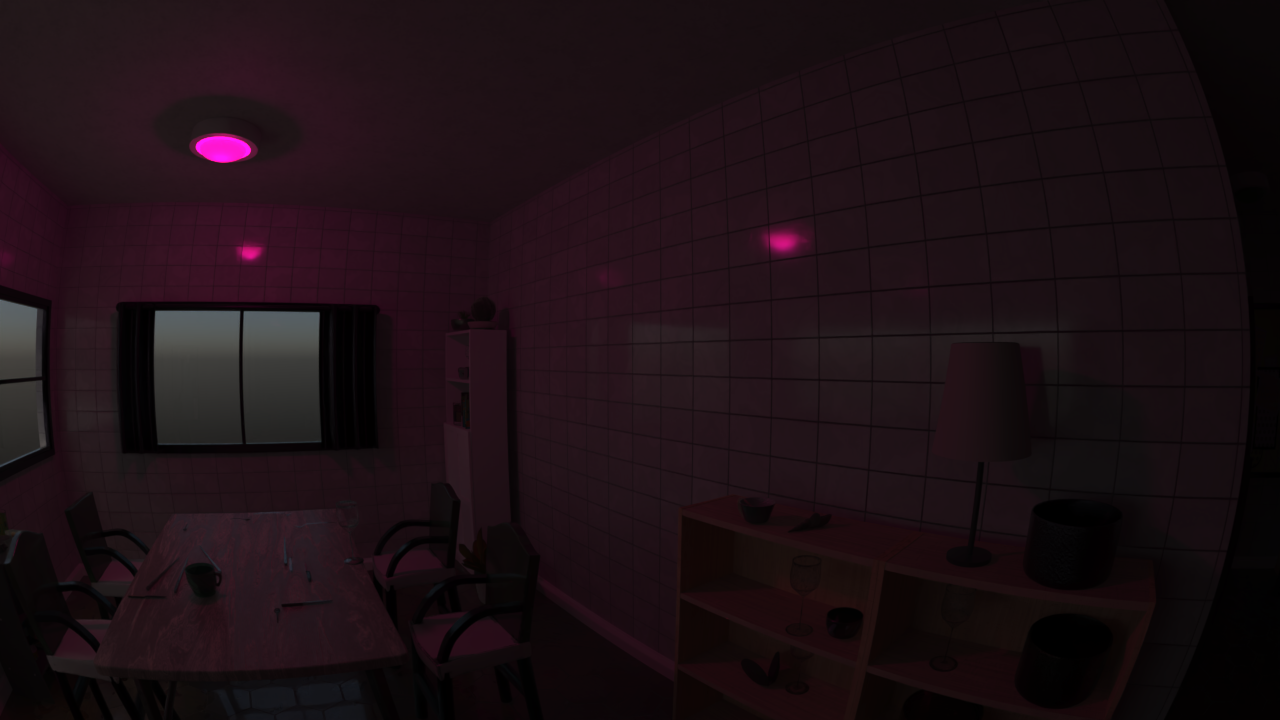}  \\[-2pt]
    \includegraphics[width=\cellW,height=\cellH,keepaspectratio]{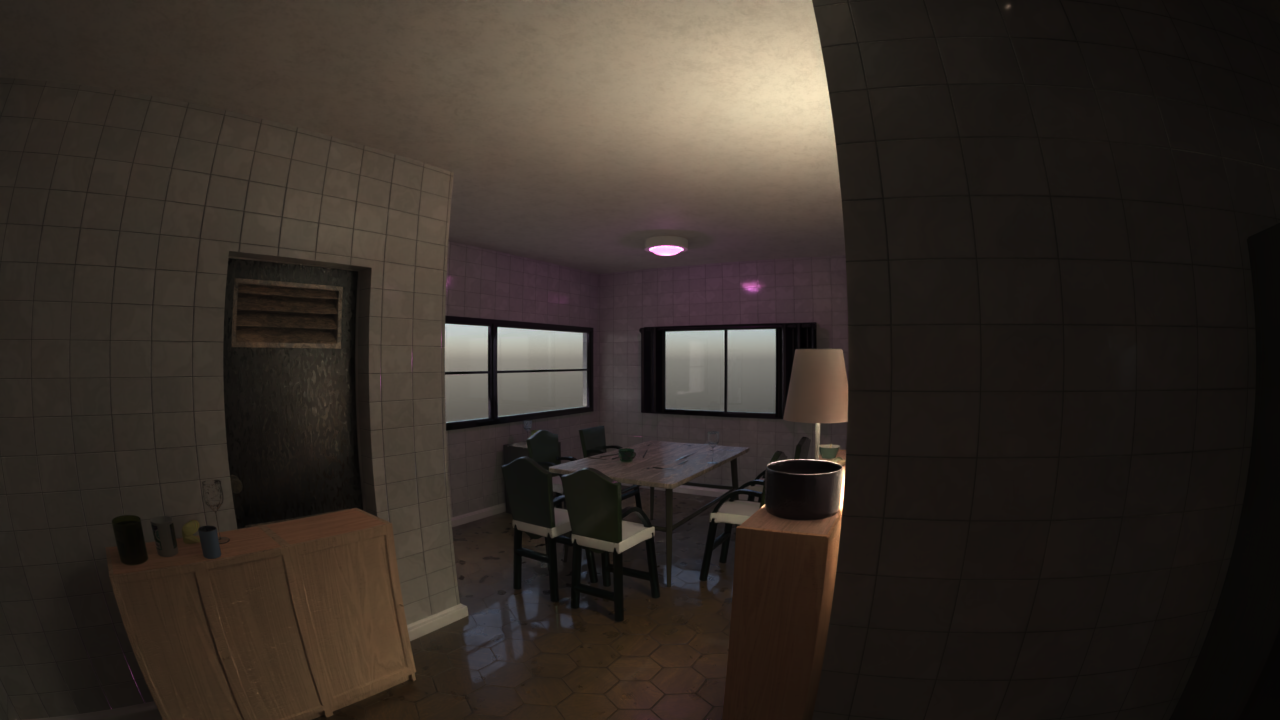} &
    \includegraphics[width=\cellW,height=\cellH,keepaspectratio]{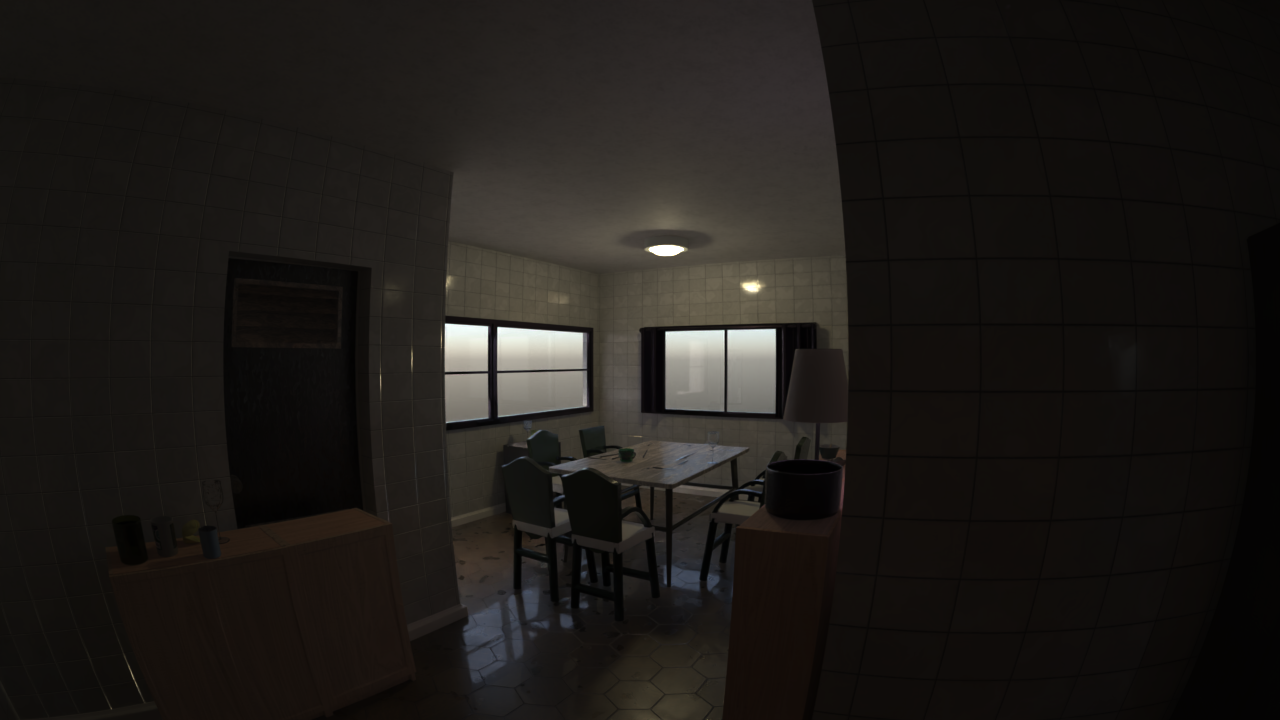} &
    \includegraphics[width=\cellW,height=\cellH,keepaspectratio]{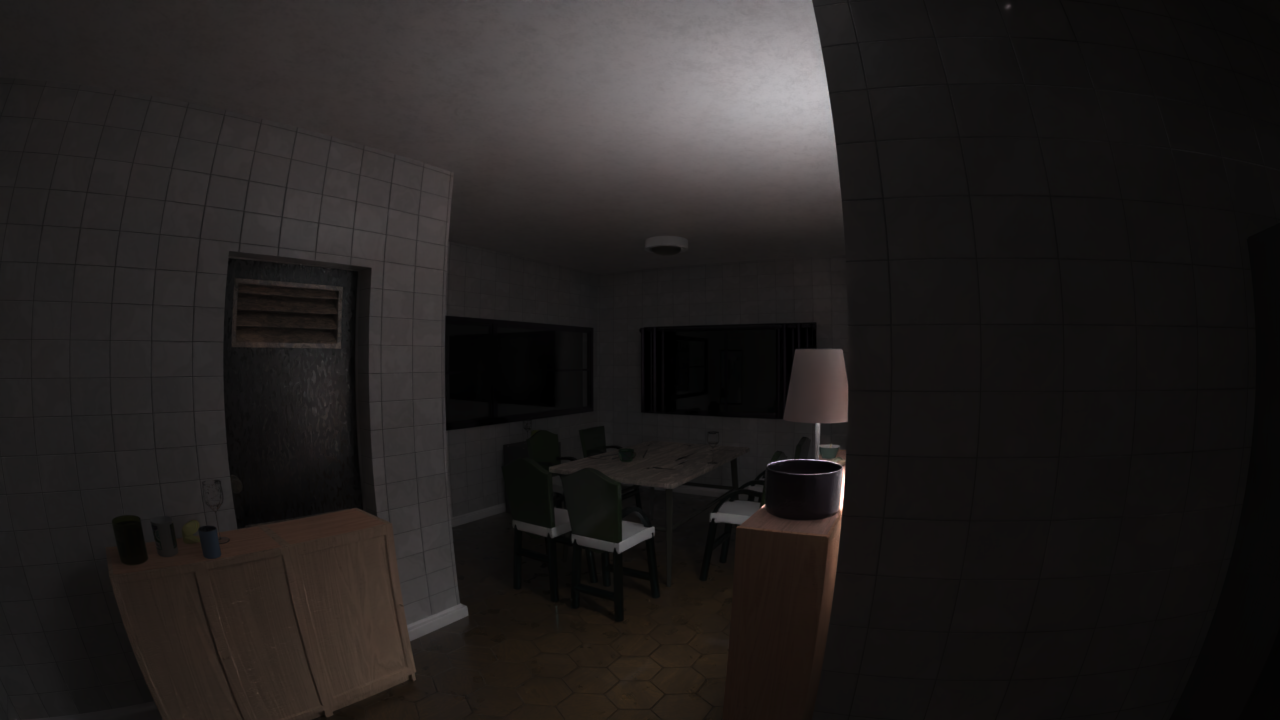} &
    \includegraphics[width=\cellW,height=\cellH,keepaspectratio]{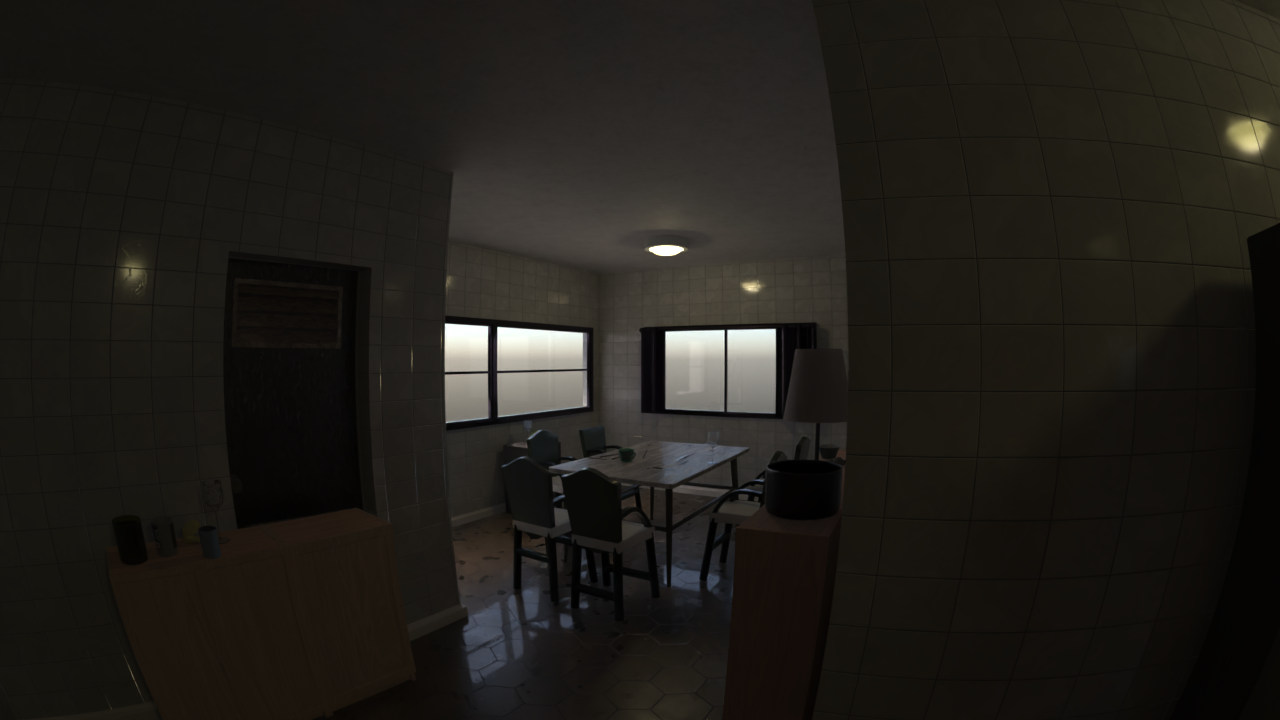} &
    \includegraphics[width=\cellW,height=\cellH,keepaspectratio]{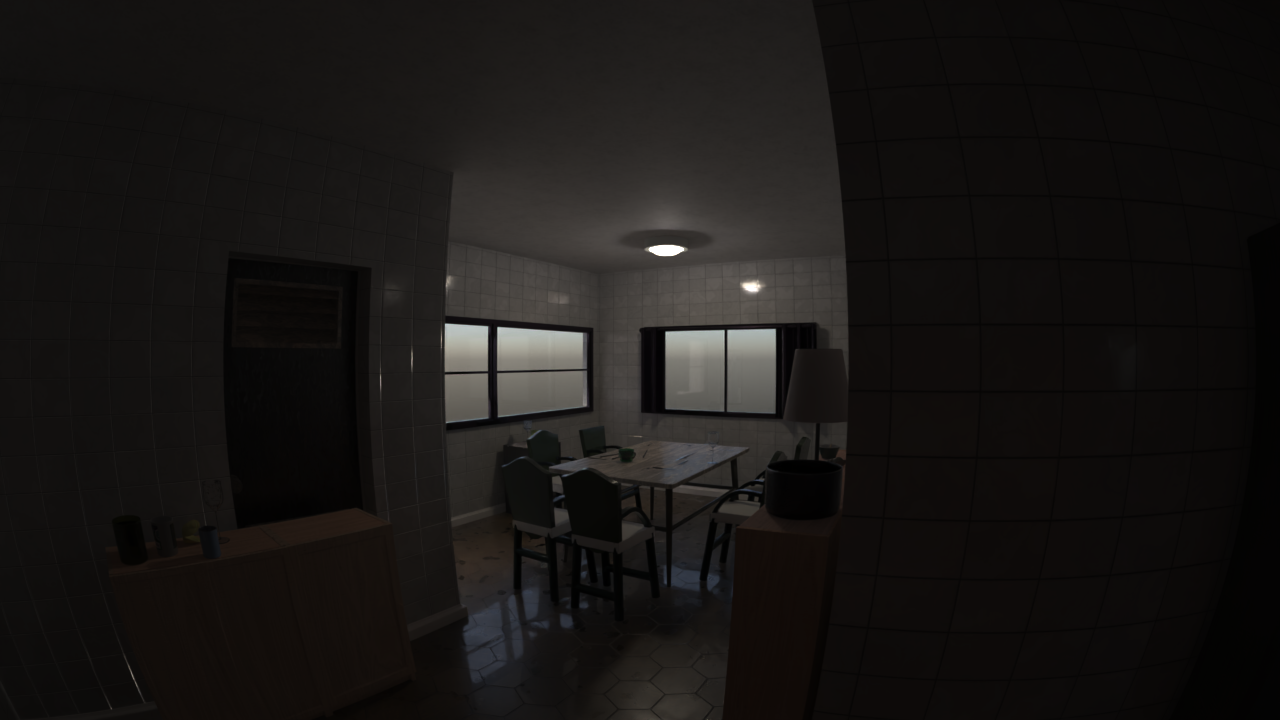} &
    \includegraphics[width=\cellW,height=\cellH,keepaspectratio]{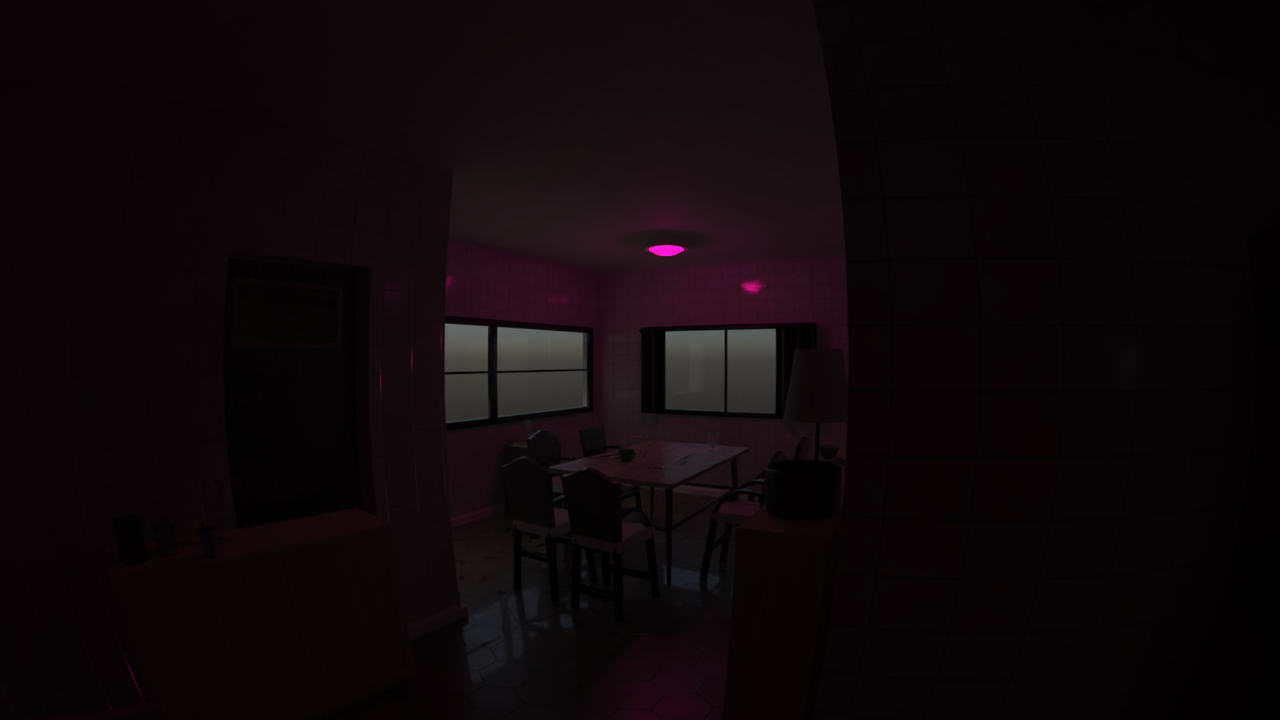}  \\[-2pt]
    \includegraphics[width=\cellW,height=\cellH,keepaspectratio]{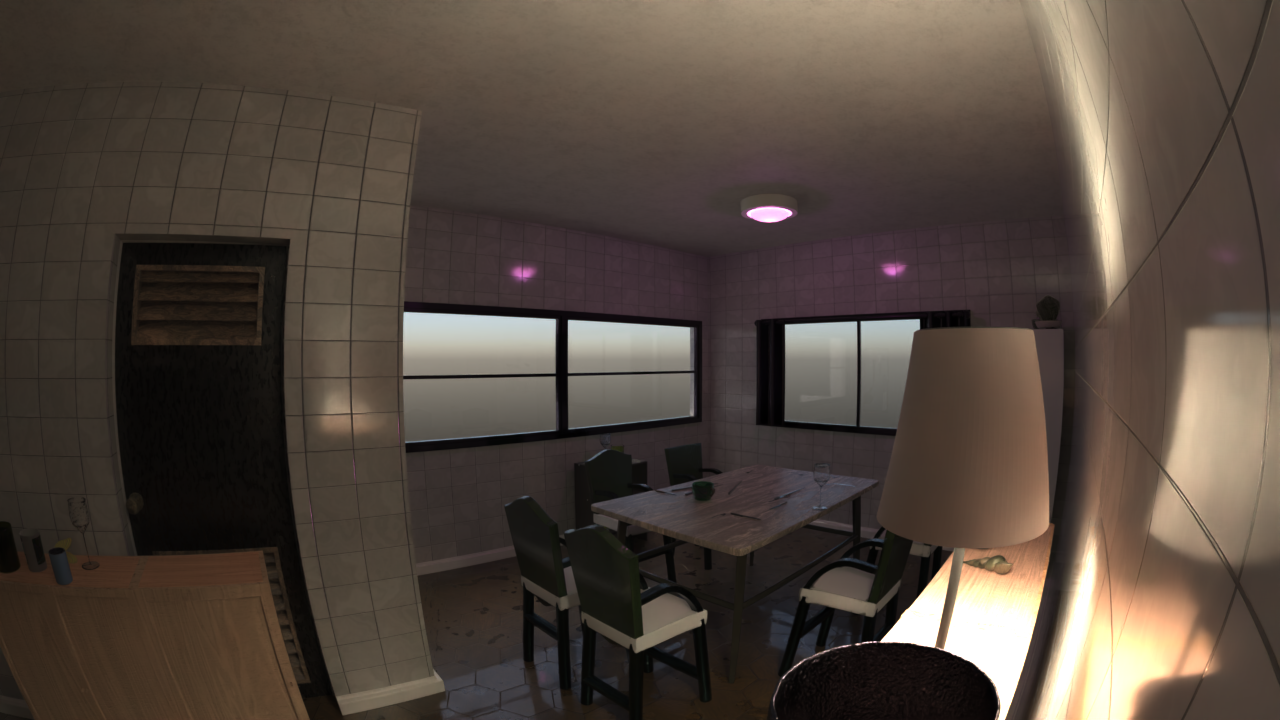} &
    \includegraphics[width=\cellW,height=\cellH,keepaspectratio]{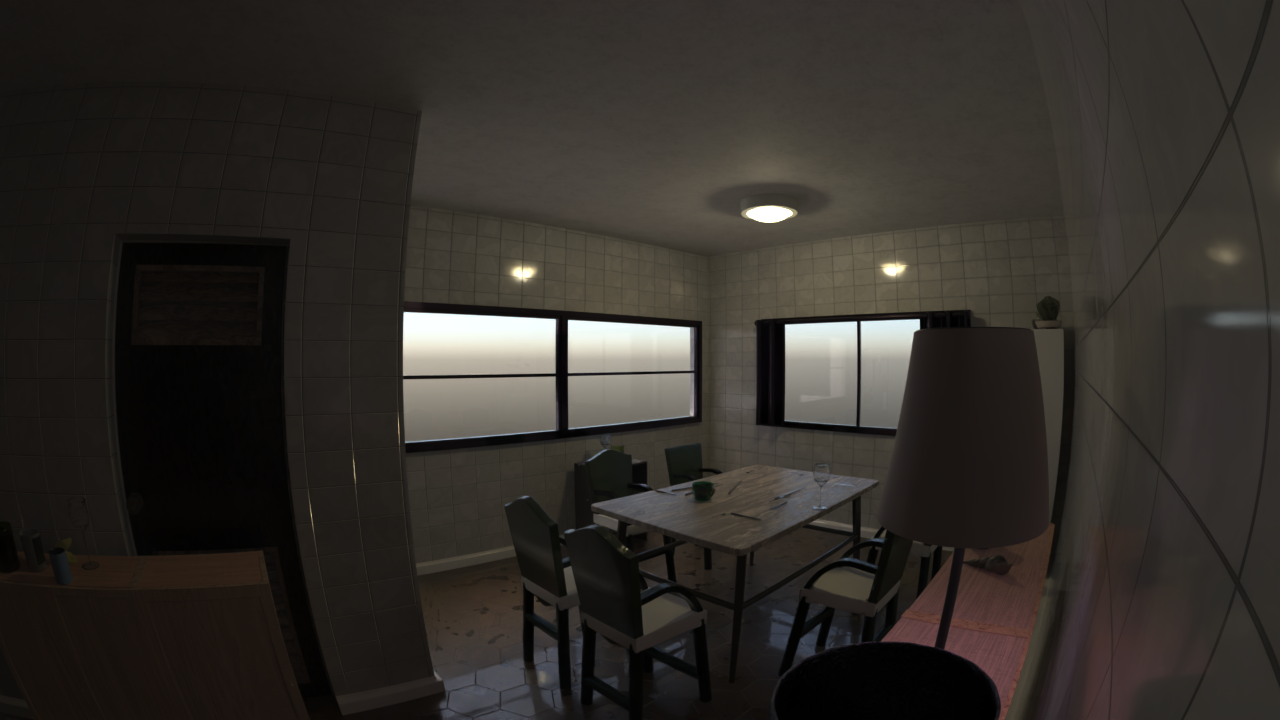} &
    \includegraphics[width=\cellW,height=\cellH,keepaspectratio]{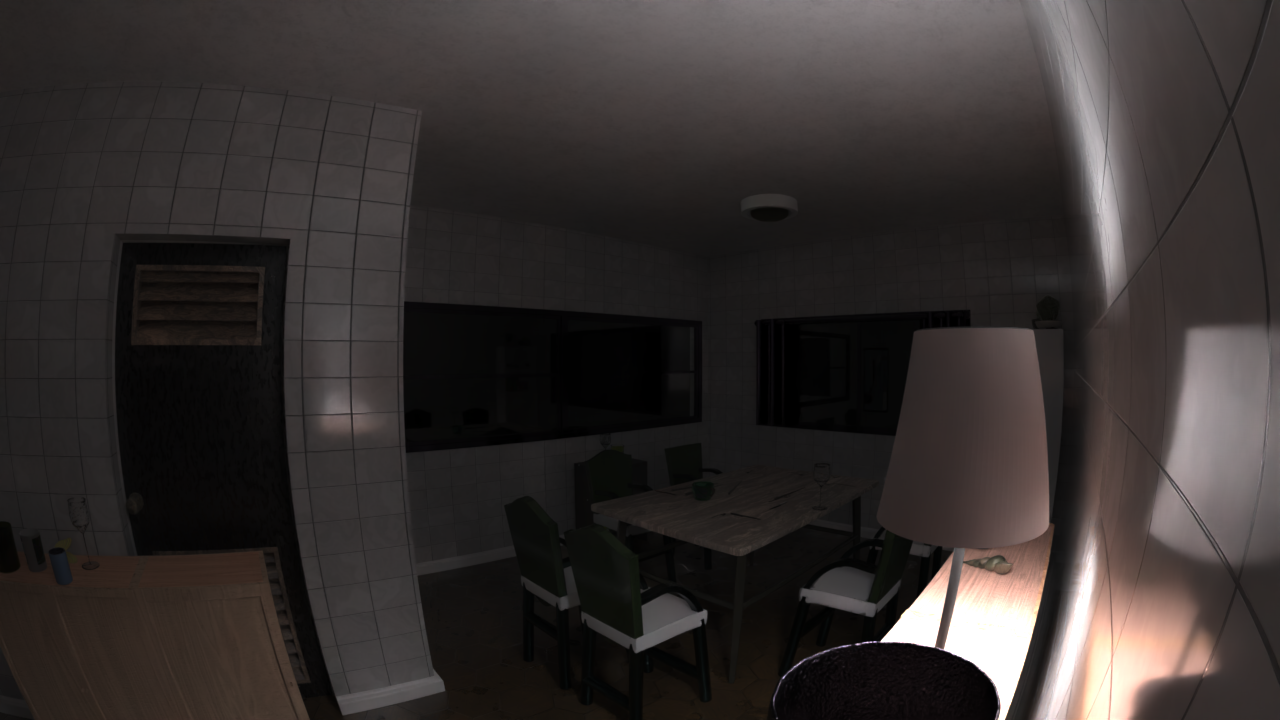} &
    \includegraphics[width=\cellW,height=\cellH,keepaspectratio]{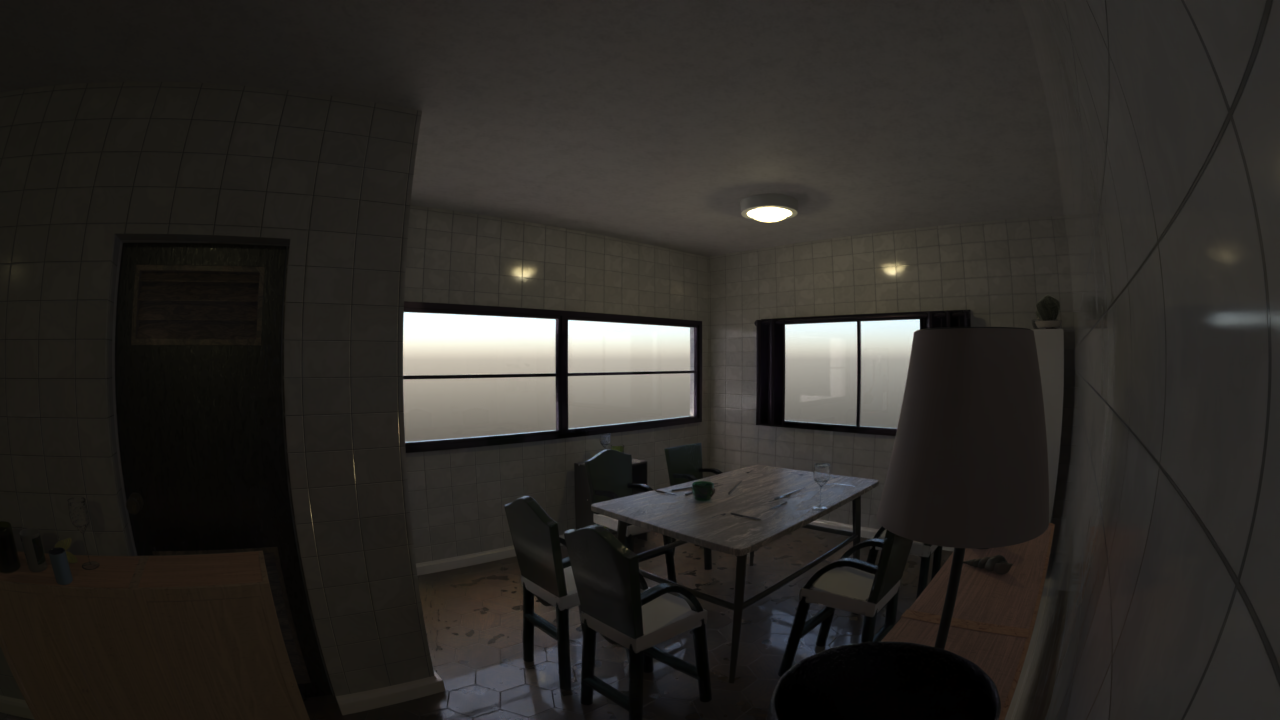} &
    \includegraphics[width=\cellW,height=\cellH,keepaspectratio]{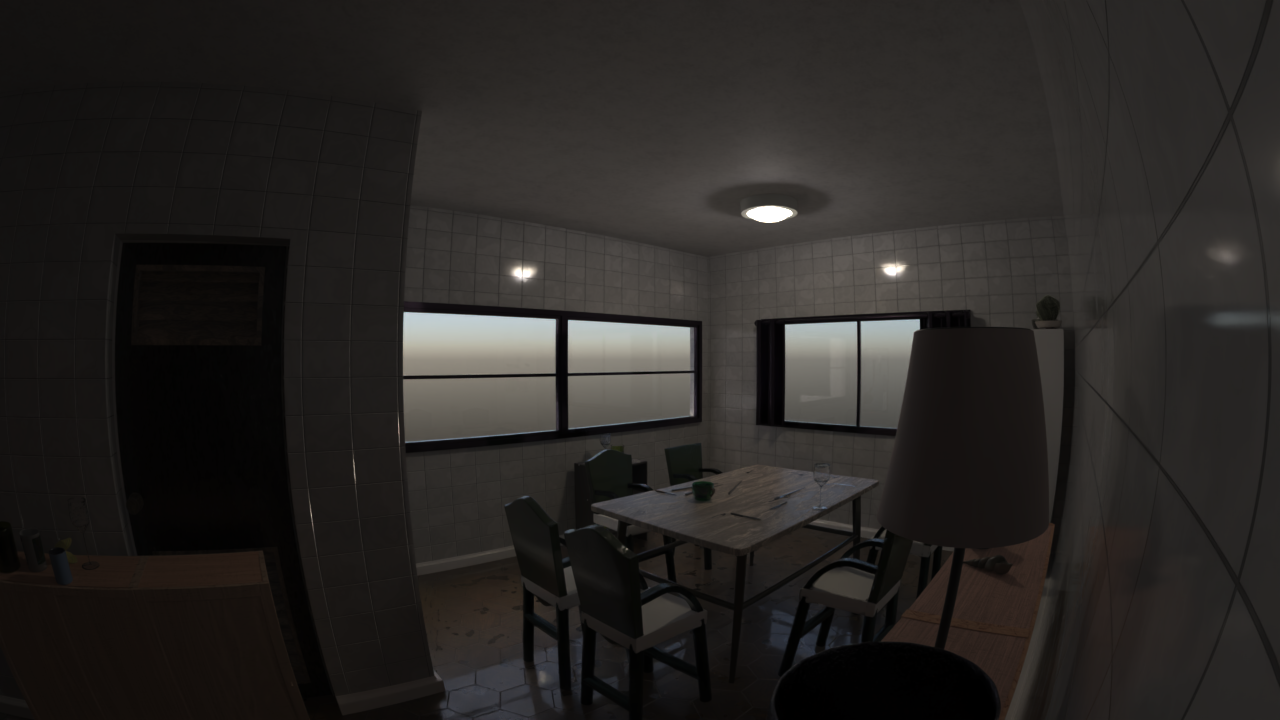} &
    \includegraphics[width=\cellW,height=\cellH,keepaspectratio]{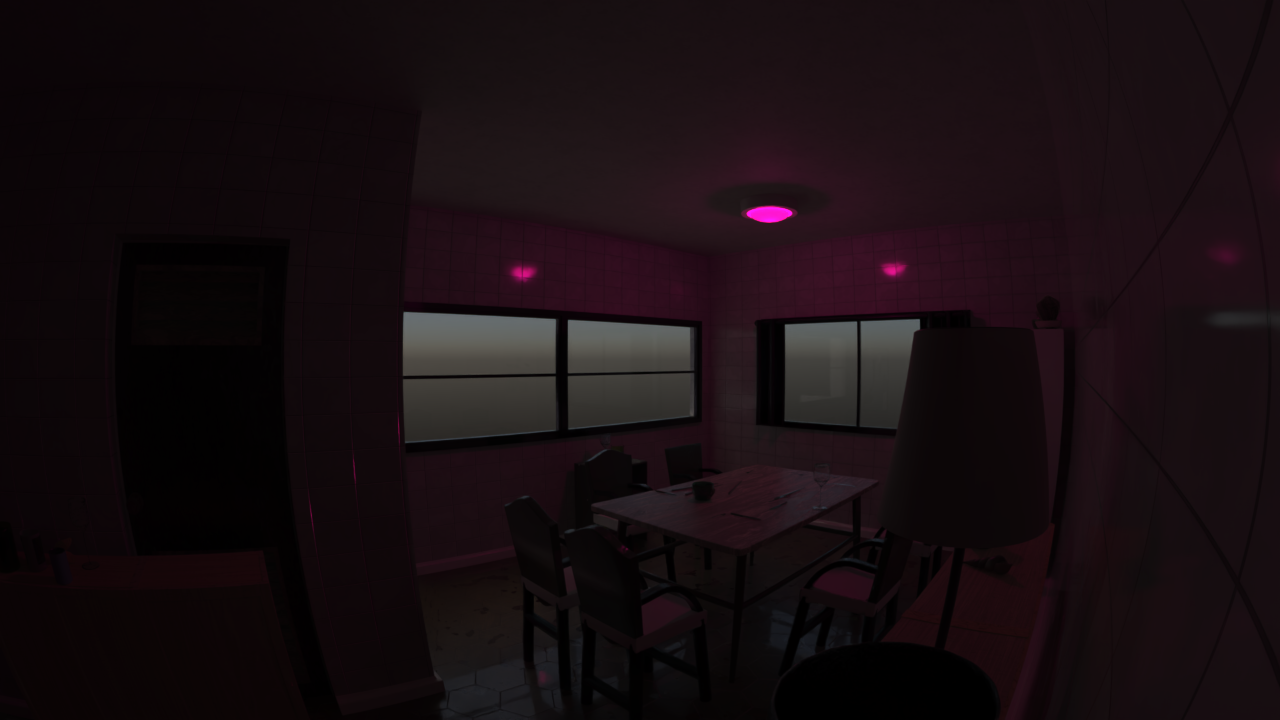}  \\
        \parbox[c][\baselineskip][c]{\cellW}{\centering\small Lighting 1} &
    \parbox[c][\baselineskip][c]{\cellW}{\centering\small Lighting 2} &
    \parbox[c][\baselineskip][c]{\cellW}{\centering\small Lighting 3} &
    \parbox[c][\baselineskip][c]{\cellW}{\centering\small Lighting 4} &
    \parbox[c][\baselineskip][c]{\cellW}{\centering\small Lighting 5} &
    \parbox[c][\baselineskip][c]{\cellW}{\centering\small Lighting 6} 
  \end{tabular}
  \caption{\textbf{An example of a scene used to train our method.} Each column corresponds to a different lighting condition, and each row to a different view. We train our relighting model by feeding into it a video of the scene under one lighting condition and predicting the same scene under a target lighting condition.}
  \label{fig:datagen1}
\end{figure*}

\setlength{\cellW}{0.16\textwidth}

\setlength{\cellH}{0.5625\cellW}

\setlength{\rowLabelW}{0.03\textwidth}

\begin{figure*}[p]
  \ContinuedFloat
  \centering
  \setlength{\tabcolsep}{2pt}          %
  \renewcommand{\arraystretch}{0.8}    %
  \begin{tabular}{@{}c*{5}{c}@{}}

    \includegraphics[width=\cellW,height=\cellH,keepaspectratio]{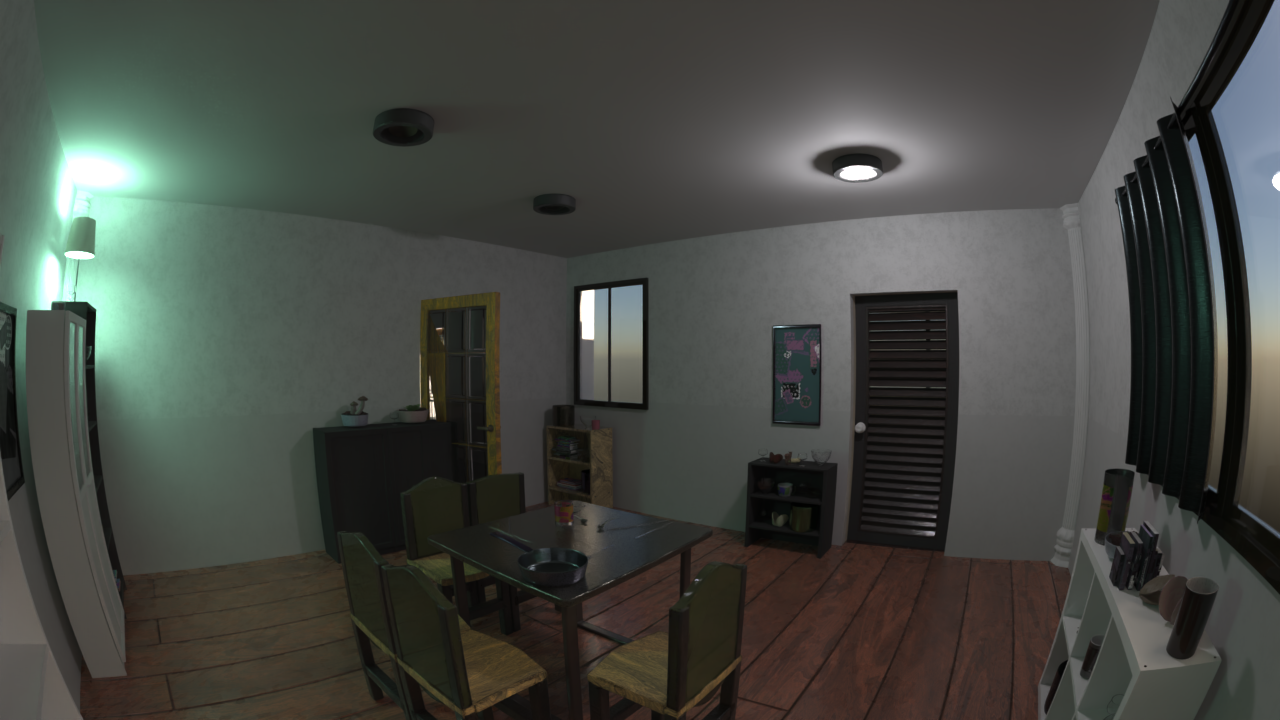} &
    \includegraphics[width=\cellW,height=\cellH,keepaspectratio]{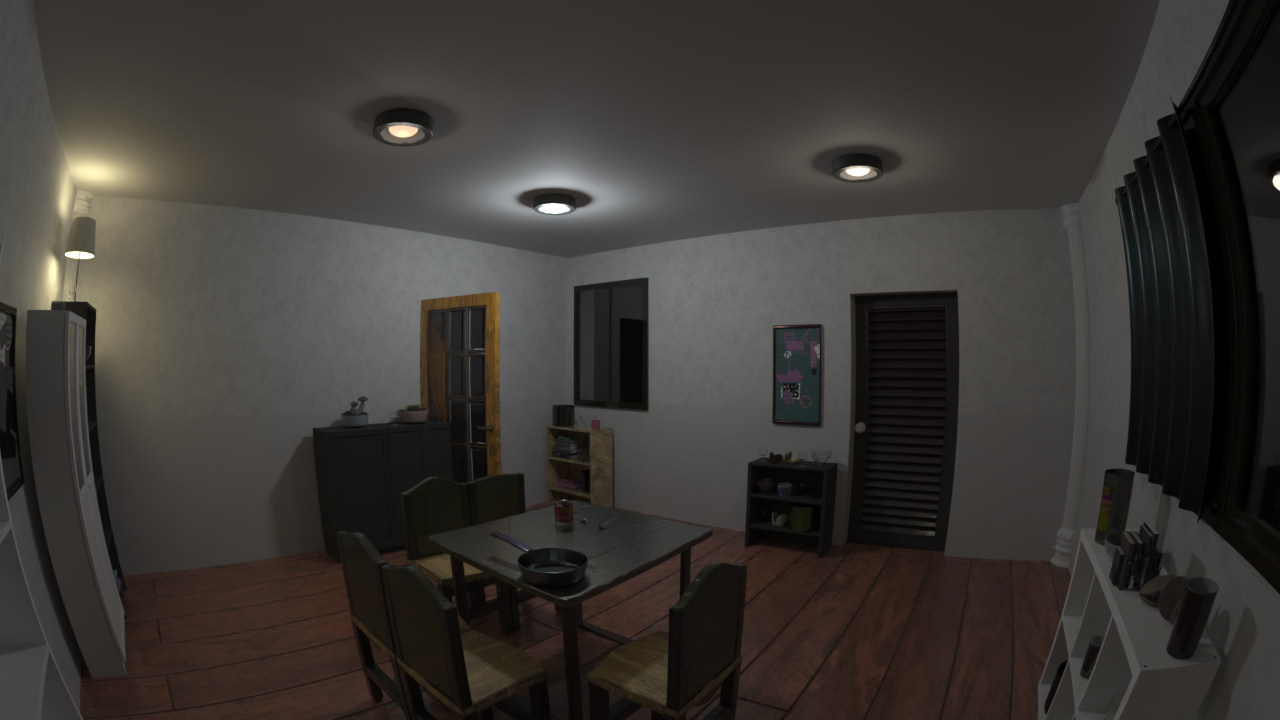} &
    \includegraphics[width=\cellW,height=\cellH,keepaspectratio]{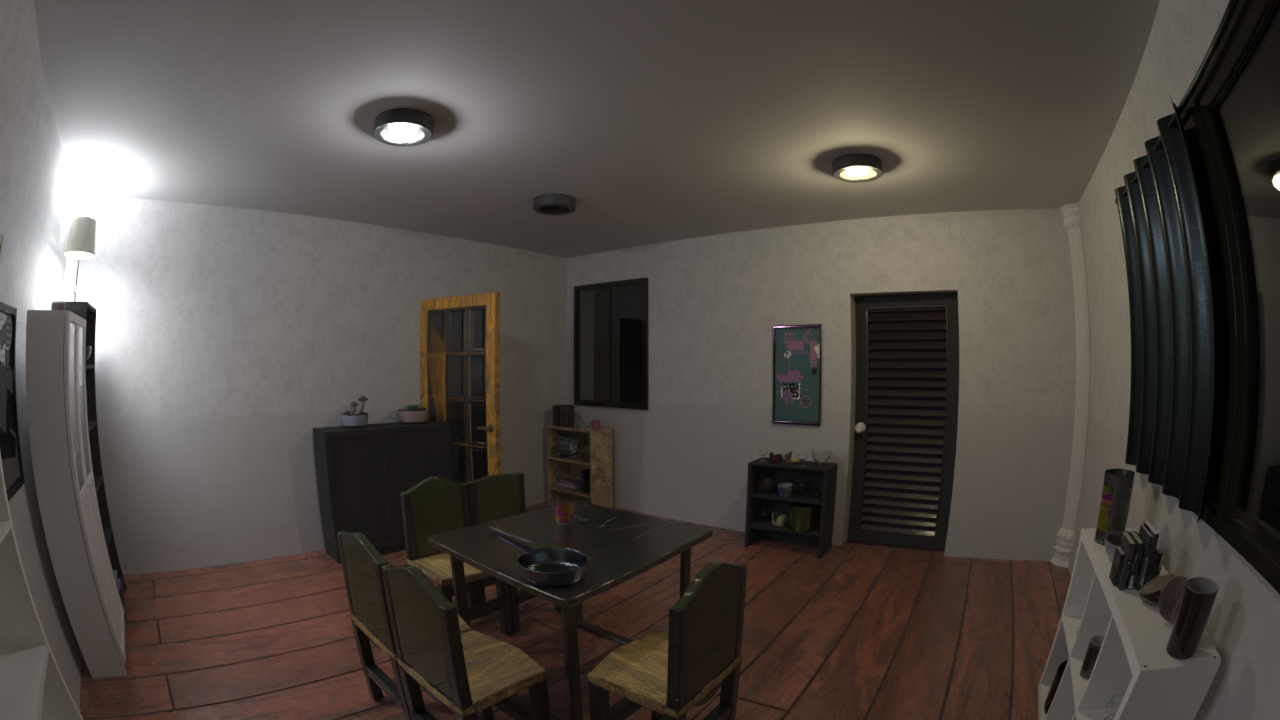} &
    \includegraphics[width=\cellW,height=\cellH,keepaspectratio]{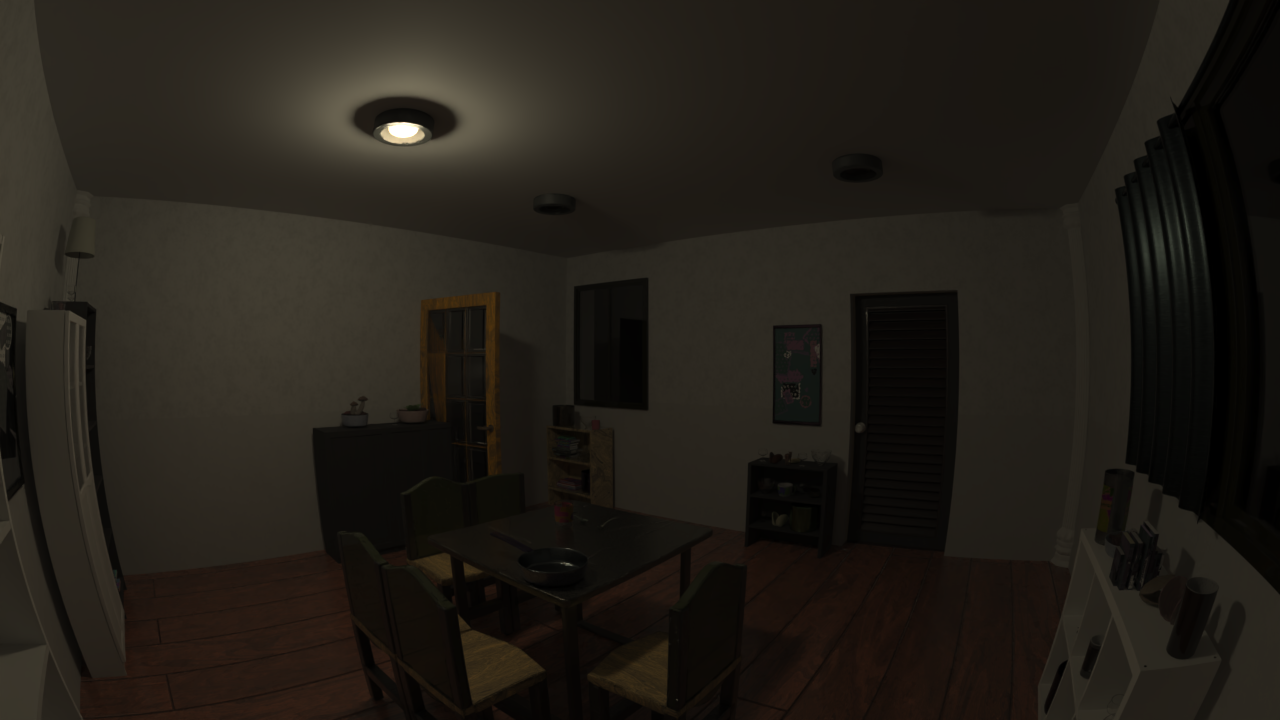} &
    \includegraphics[width=\cellW,height=\cellH,keepaspectratio]{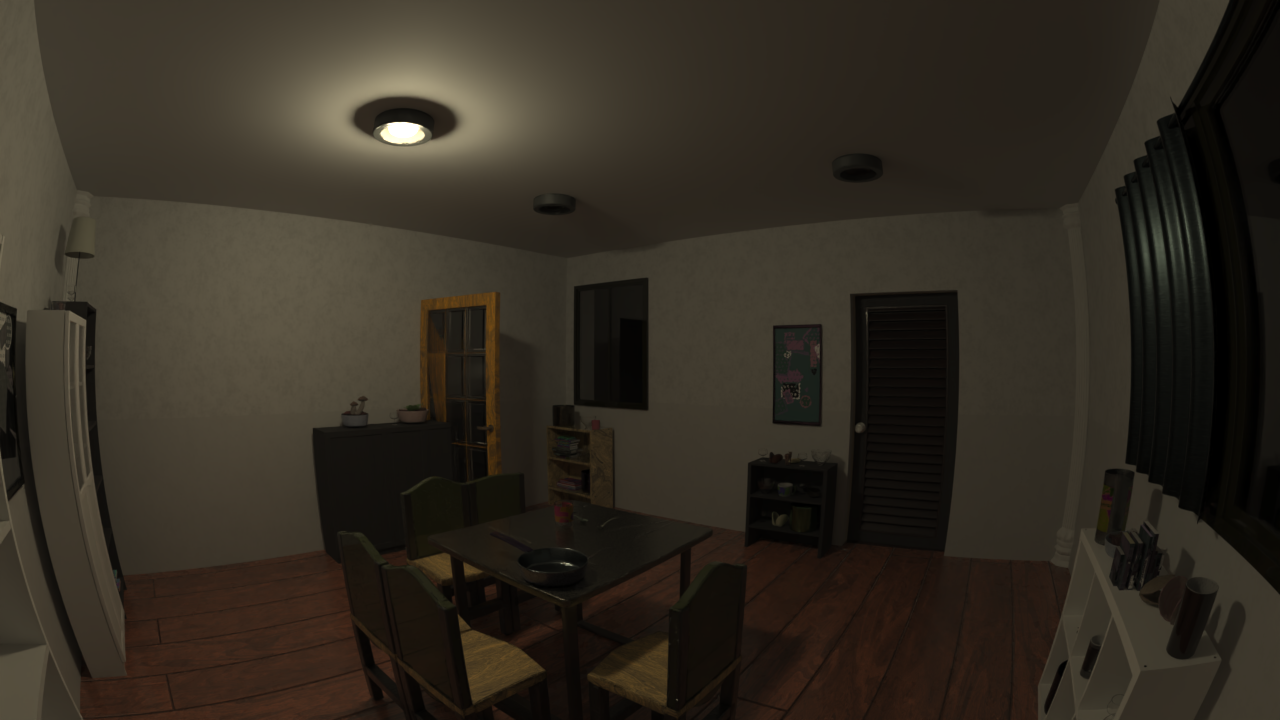} &
    \includegraphics[width=\cellW,height=\cellH,keepaspectratio]{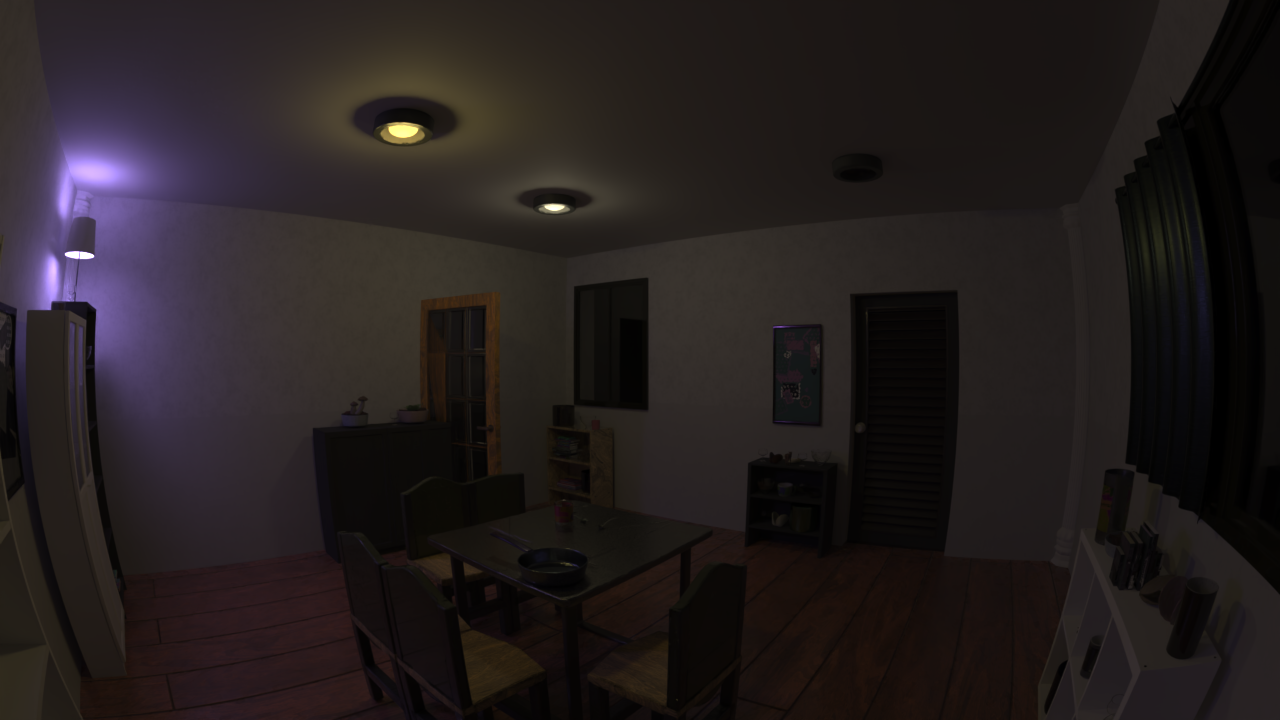}   \\[-2pt]
    \includegraphics[width=\cellW,height=\cellH,keepaspectratio]{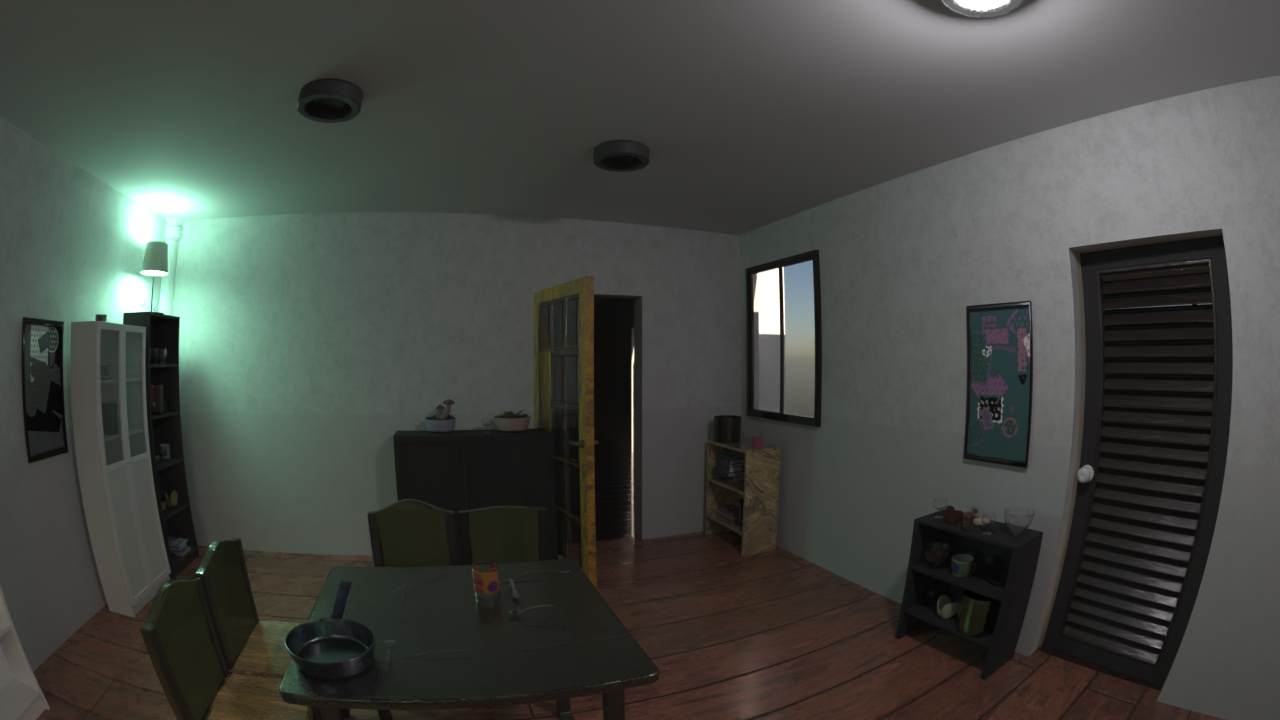} &
    \includegraphics[width=\cellW,height=\cellH,keepaspectratio]{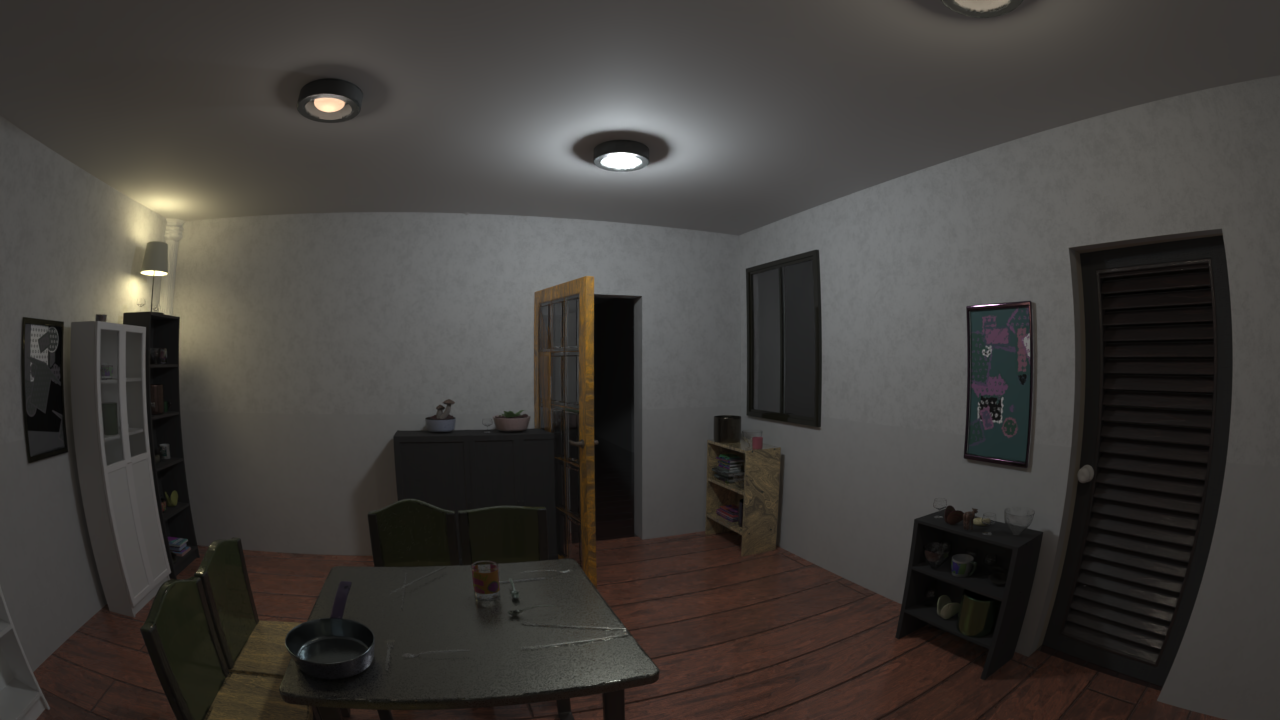} &
    \includegraphics[width=\cellW,height=\cellH,keepaspectratio]{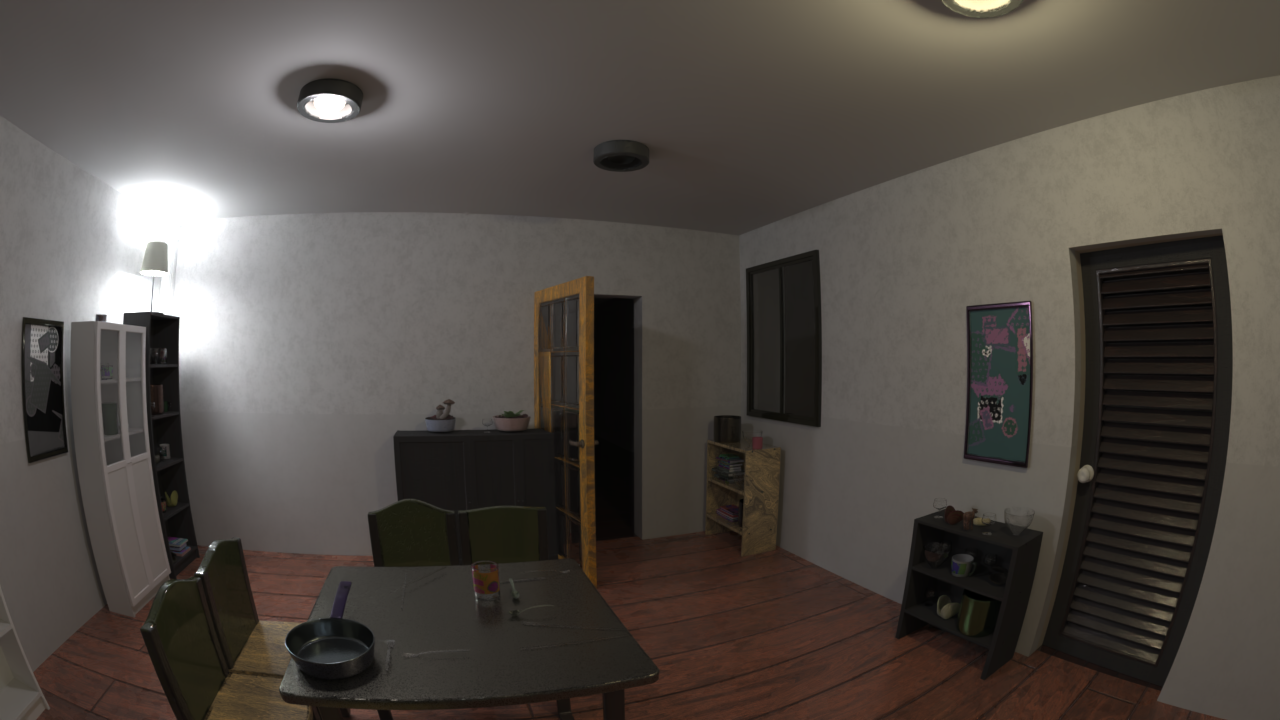} &
    \includegraphics[width=\cellW,height=\cellH,keepaspectratio]{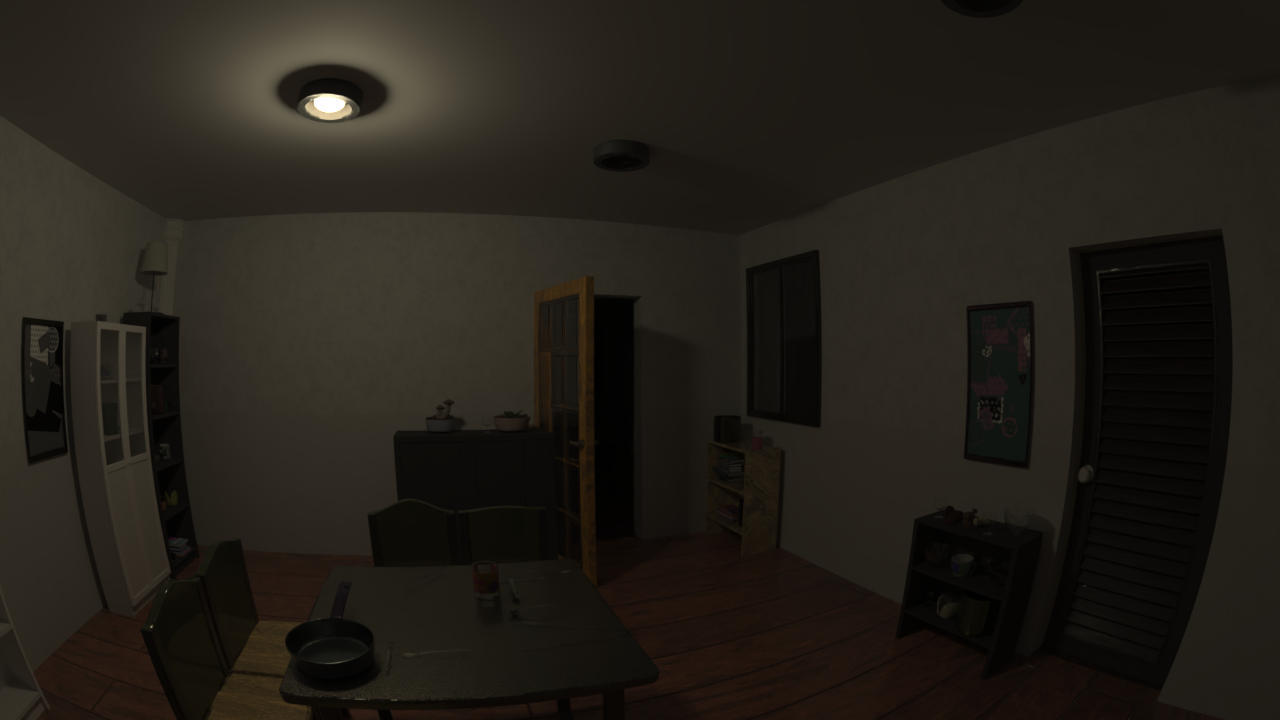} &
    \includegraphics[width=\cellW,height=\cellH,keepaspectratio]{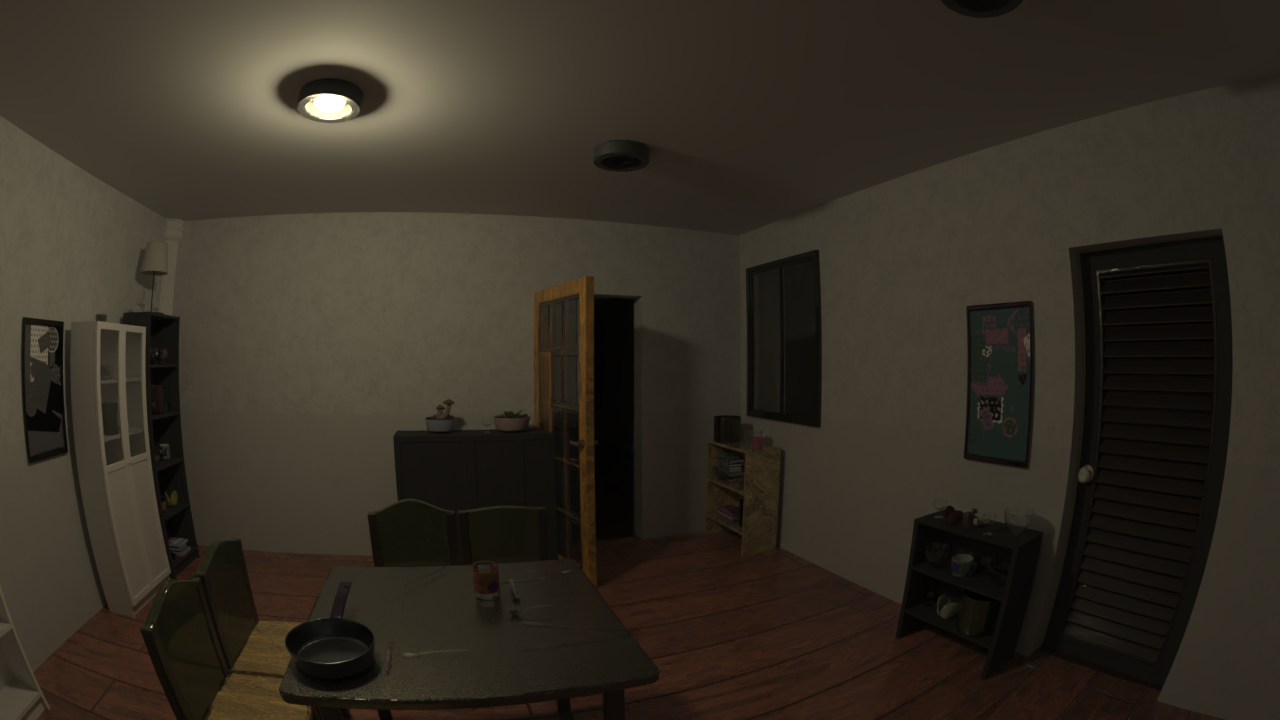} &
    \includegraphics[width=\cellW,height=\cellH,keepaspectratio]{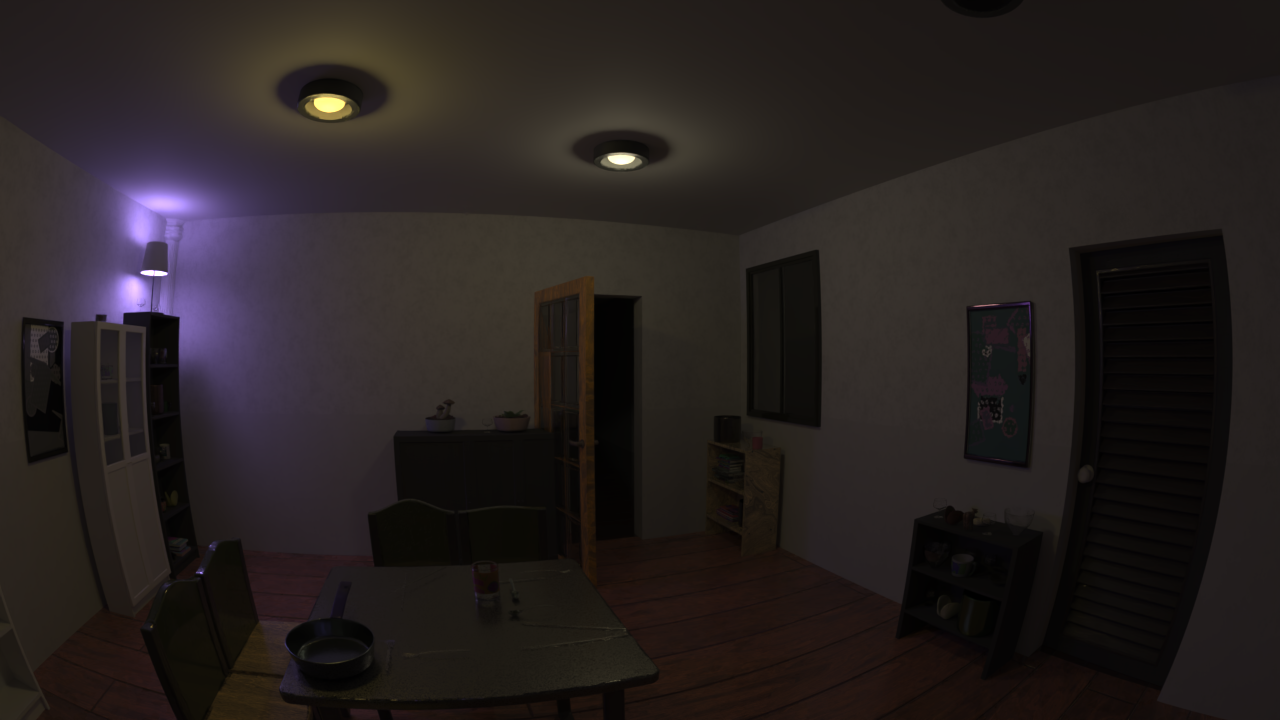}  \\[-2pt]
    \includegraphics[width=\cellW,height=\cellH,keepaspectratio]{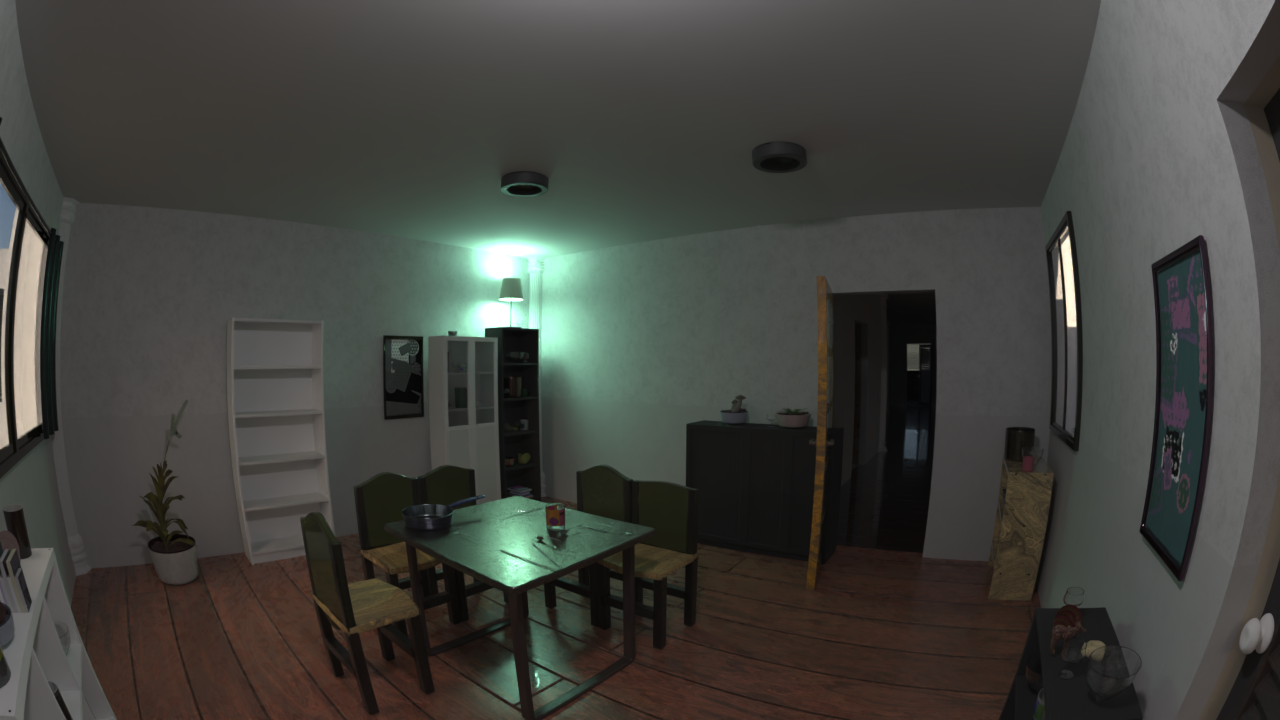} &
    \includegraphics[width=\cellW,height=\cellH,keepaspectratio]{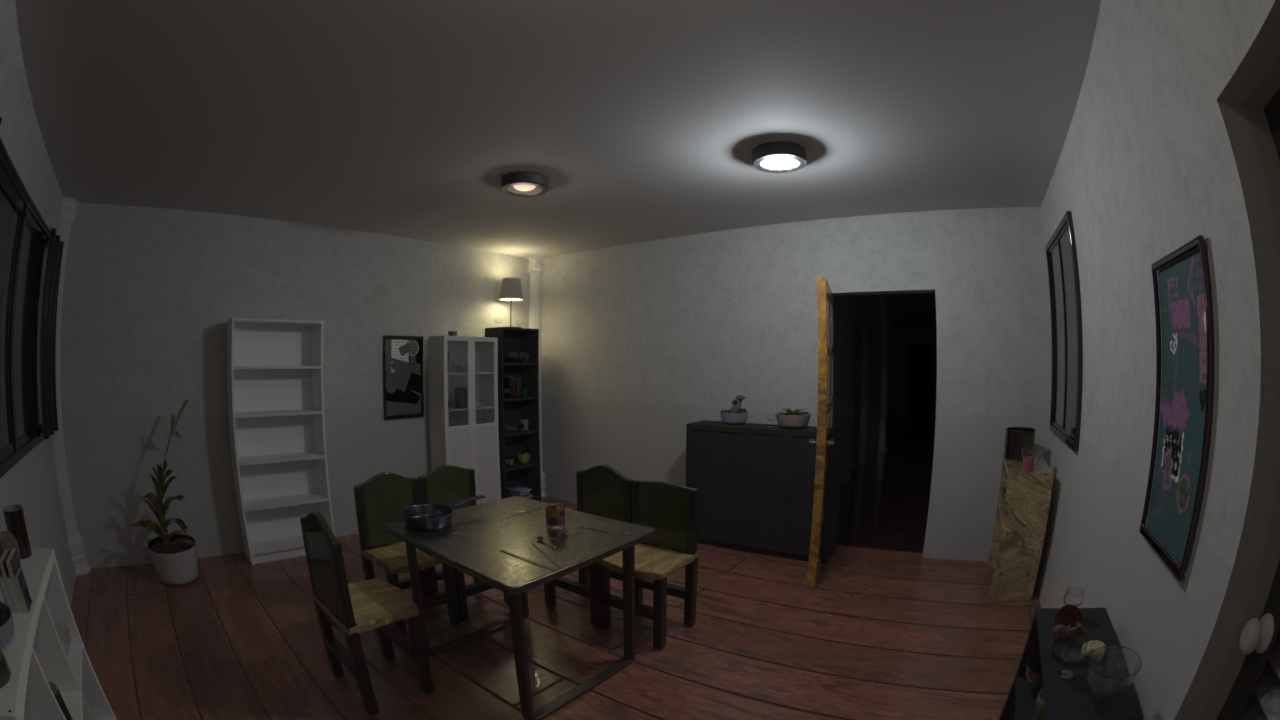} &
    \includegraphics[width=\cellW,height=\cellH,keepaspectratio]{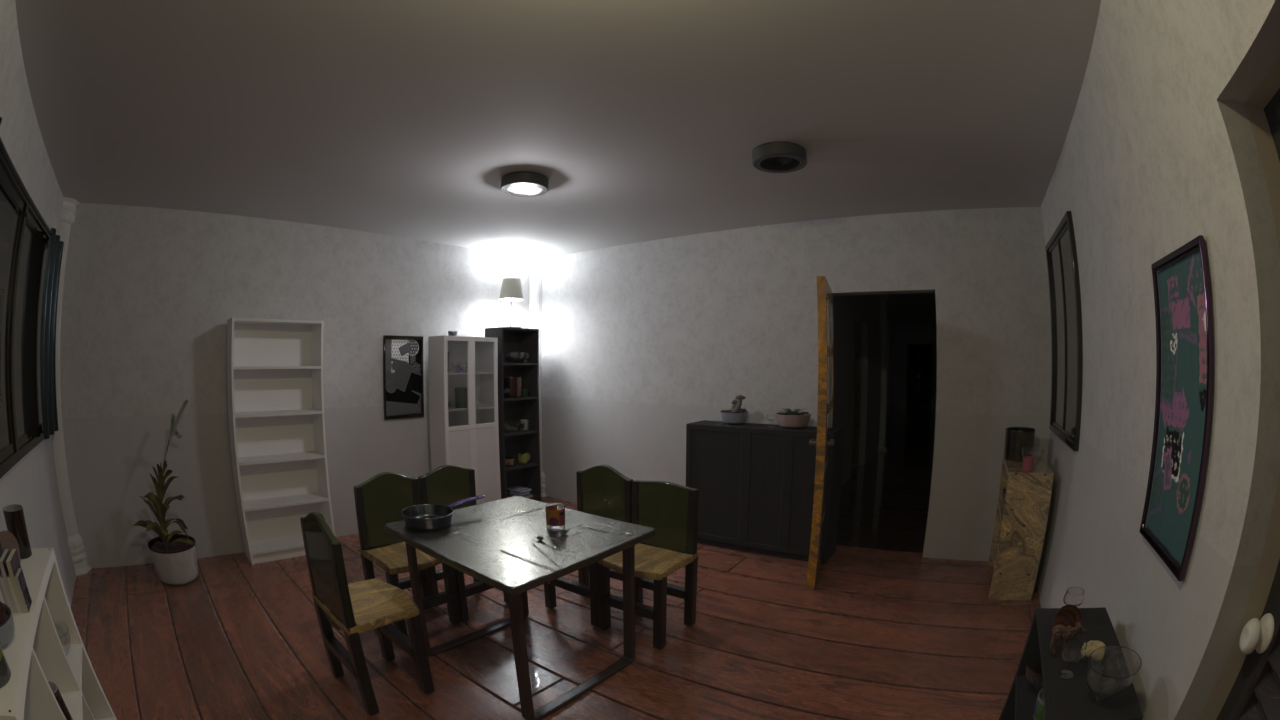} &
    \includegraphics[width=\cellW,height=\cellH,keepaspectratio]{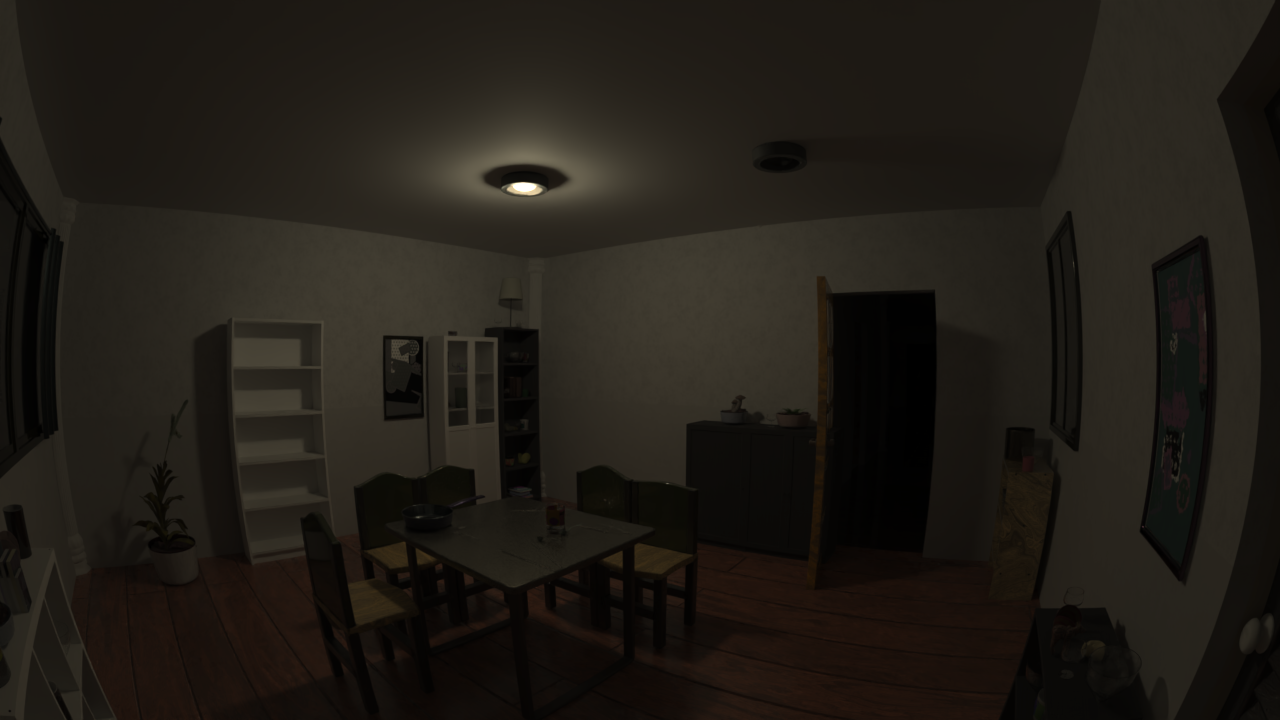} &
    \includegraphics[width=\cellW,height=\cellH,keepaspectratio]{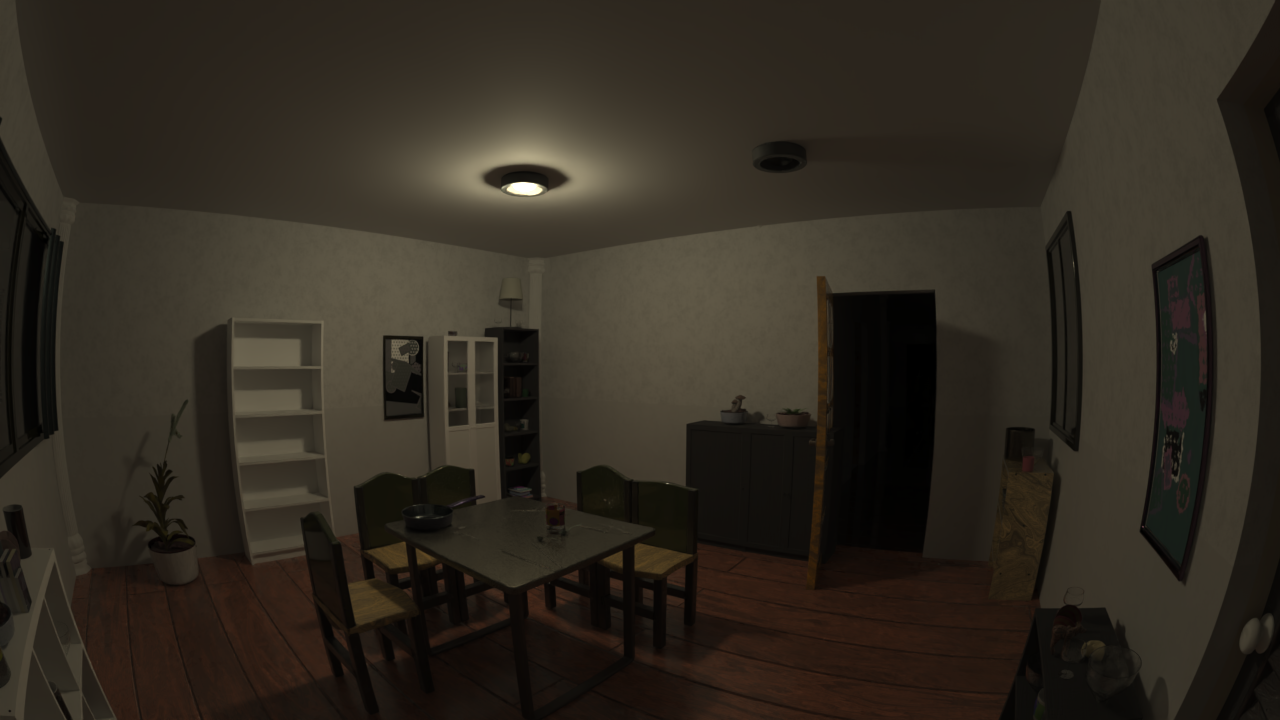} &
    \includegraphics[width=\cellW,height=\cellH,keepaspectratio]{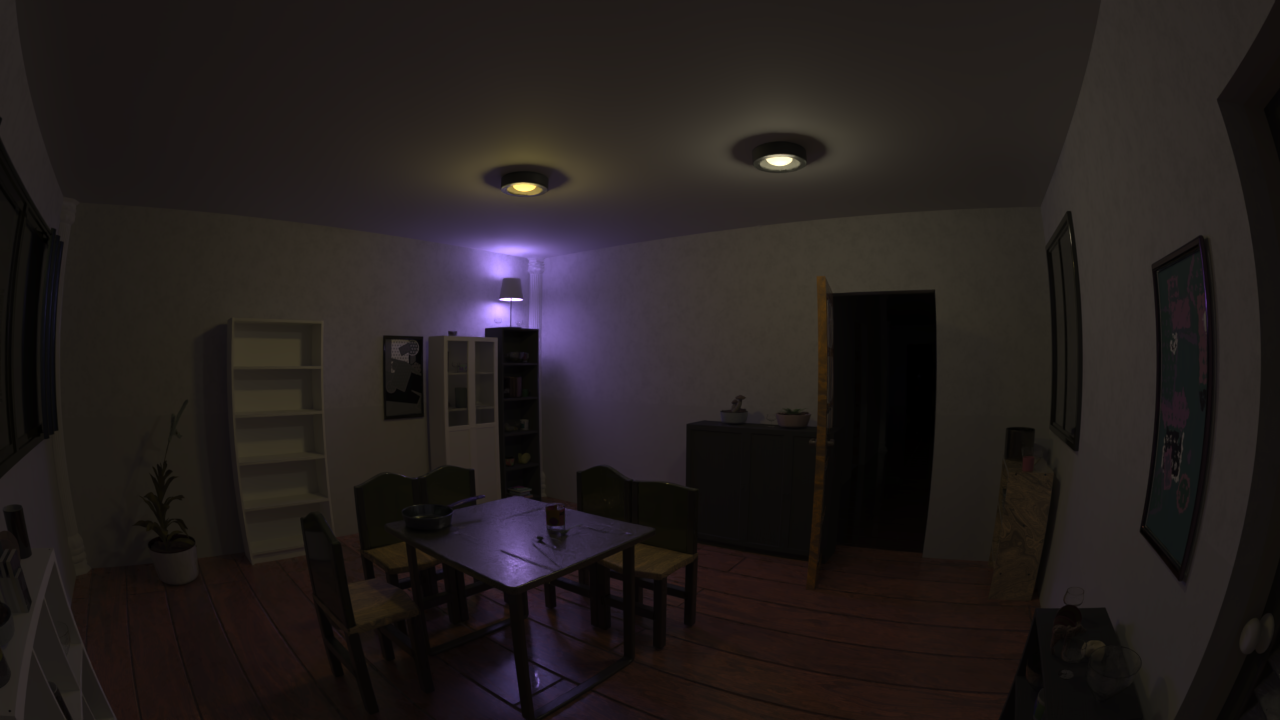}  \\[-2pt]
    \includegraphics[width=\cellW,height=\cellH,keepaspectratio]{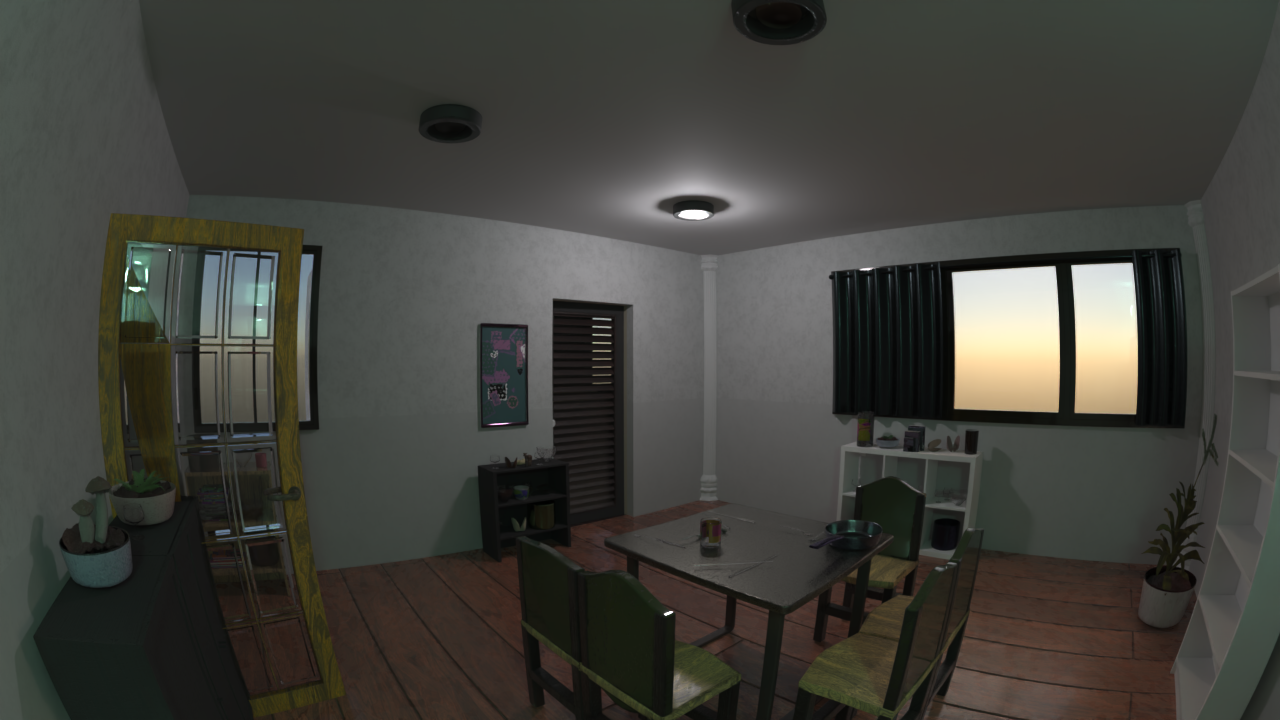} &
    \includegraphics[width=\cellW,height=\cellH,keepaspectratio]{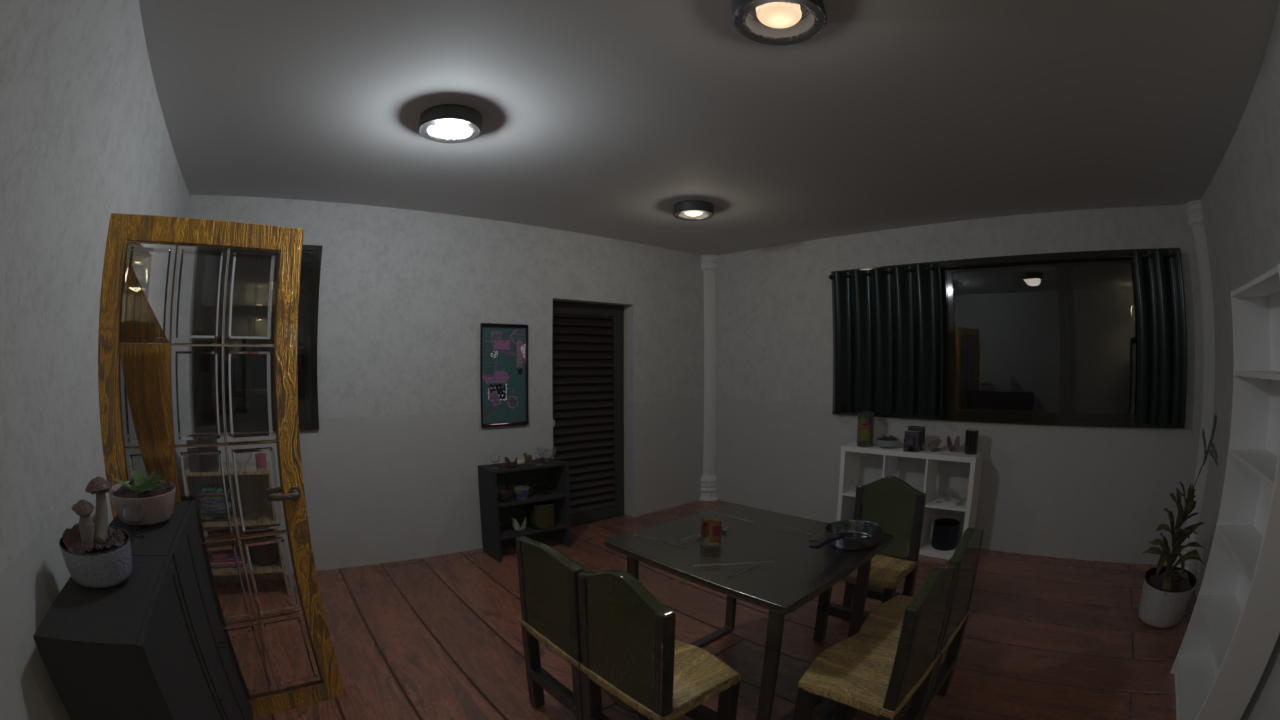} &
    \includegraphics[width=\cellW,height=\cellH,keepaspectratio]{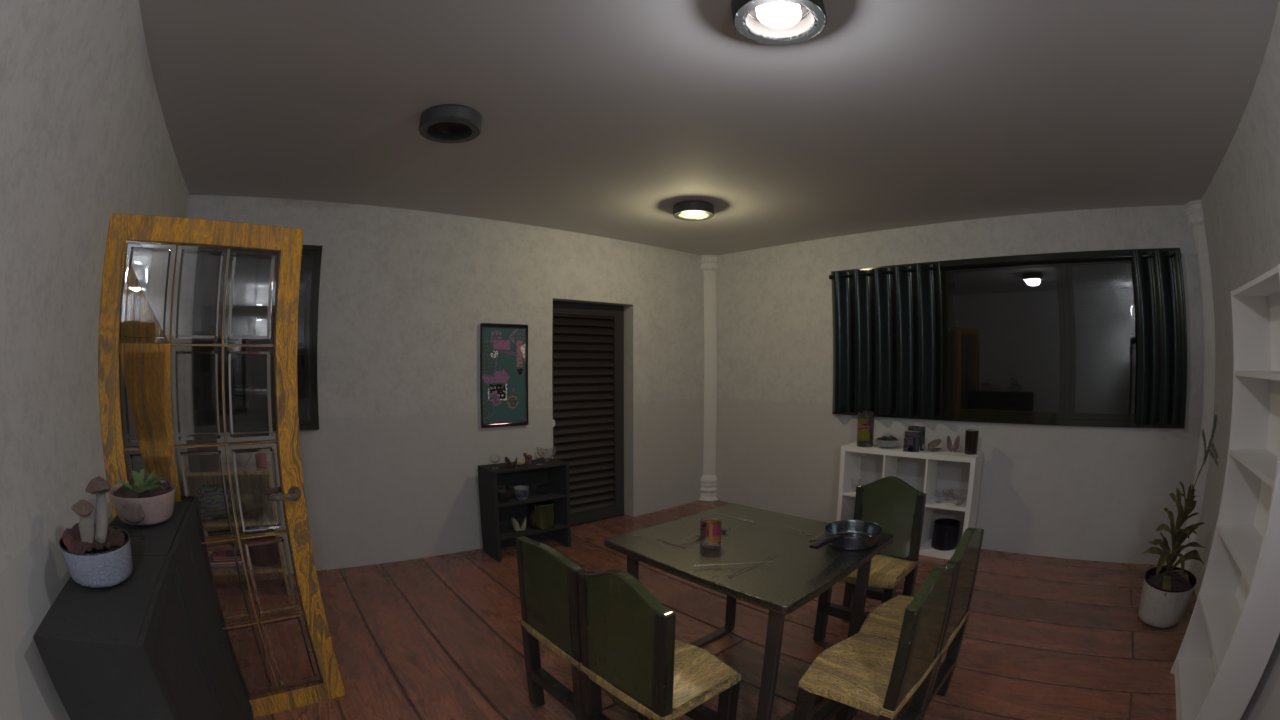} &
    \includegraphics[width=\cellW,height=\cellH,keepaspectratio]{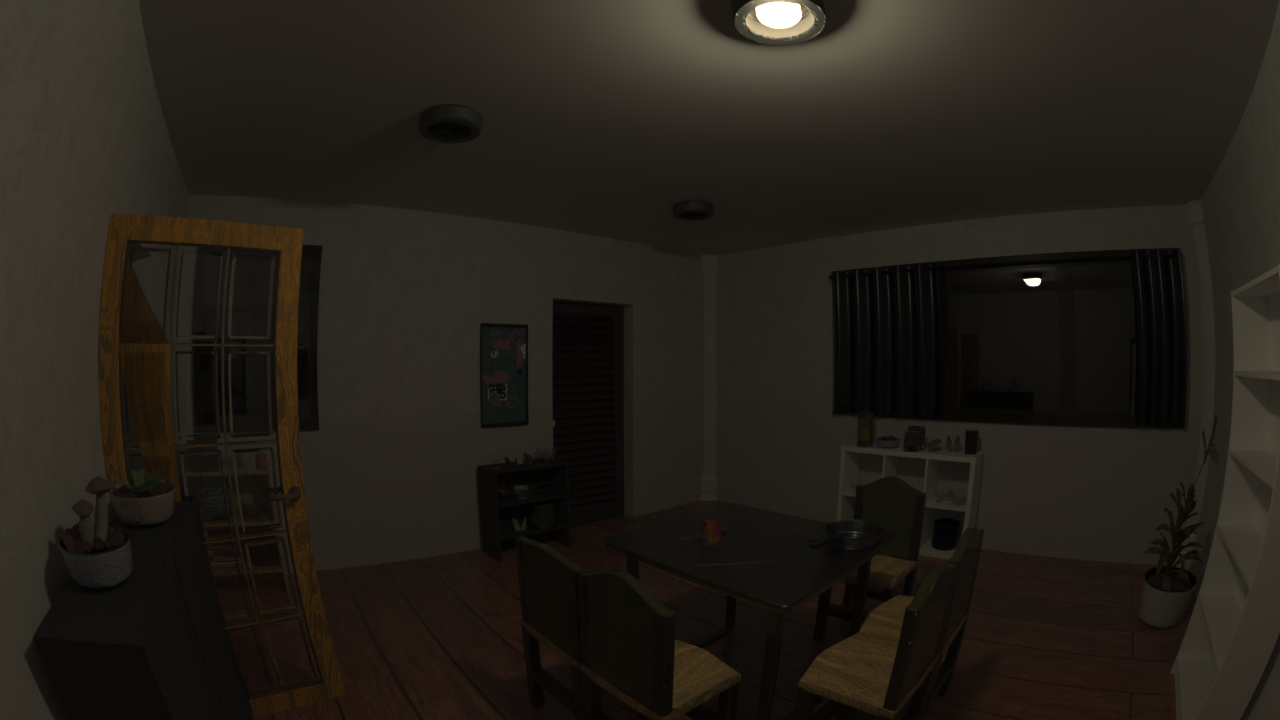} &
    \includegraphics[width=\cellW,height=\cellH,keepaspectratio]{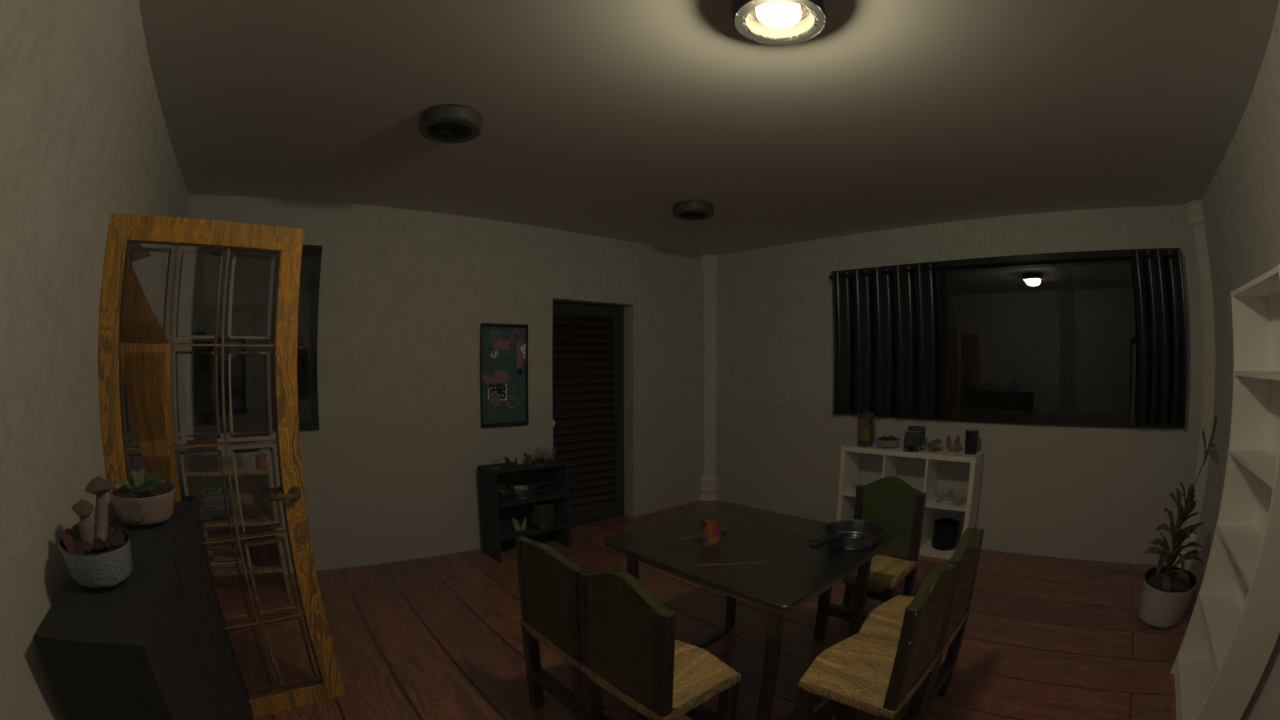} &
    \includegraphics[width=\cellW,height=\cellH,keepaspectratio]{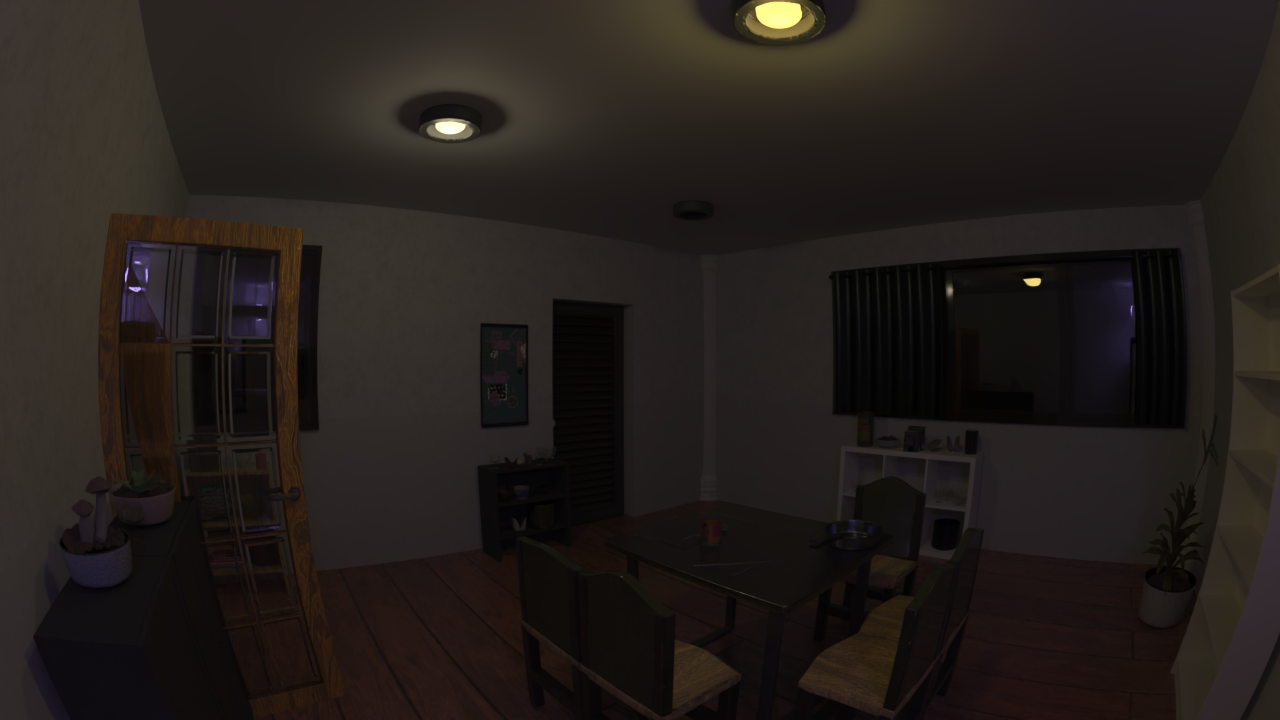}  \\
    \parbox[c][\baselineskip][c]{\cellW}{\centering\small Lighting 1} &
    \parbox[c][\baselineskip][c]{\cellW}{\centering\small Lighting 2} &
    \parbox[c][\baselineskip][c]{\cellW}{\centering\small Lighting 3} &
    \parbox[c][\baselineskip][c]{\cellW}{\centering\small Lighting 4} &
    \parbox[c][\baselineskip][c]{\cellW}{\centering\small Lighting 5} &
    \parbox[c][\baselineskip][c]{\cellW}{\centering\small Lighting 6} 
  \end{tabular}
  \caption[]{\textbf{(continued.)} Another scene from the dataset we use to train our method.}
  \label{fig:main_compare_novel_view}
\end{figure*}

\end{document}